%% file: main.tex
\documentclass{article}

\usepackage[export]{adjustbox}
\usepackage{amsmath}
\usepackage{amssymb}
\usepackage{amsthm}
\usepackage{booktabs}
\usepackage{bm}
\usepackage{color}
\usepackage{colortbl}
\usepackage{enumitem}
\usepackage{graphicx}
\usepackage[utf8]{inputenc}
\usepackage{listings}
\usepackage{makecell}
\usepackage{mathtools}
\usepackage{microtype}
\usepackage{multirow}
\usepackage[numbers,sort&compress]{natbib}
\usepackage[olditem,oldenum]{paralist}
\usepackage{tabularx}
\usepackage[dvipsnames]{xcolor}
\usepackage{xfrac}
\usepackage{xspace}

\usepackage[pagebackref,breaklinks,colorlinks,bookmarks=False]{hyperref}

\usepackage[capitalize,noabbrev]{cleveref}
\crefname{equation}{Eqn.}{Eqns.}
\creflabelformat{equation}{#2\textup{#1}#3}

\usepackage[accepted]{icml2023}

\usepackage{inconsolata}

\renewcommand{\paragraph}[1]{\noindent \textbf{#1}}

\renewcommand{\arraystretch}{1.2}
\def\eg{\emph{e.g.,}\xspace} 
\def\ie{\emph{i.e.,}\xspace} 

\newcommand{\vu}{\mathbf{u}}
\newcommand{\vv}{\mathbf{v}}
\newcommand{\vx}{\mathbf{x}}
\newcommand{\vy}{\mathbf{y}}
\newcommand{\vz}{\mathbf{z}}
\newcommand{\vO}{\mathbf{O}}
\newcommand{\bbR}{\mathbb{R}}

\def\cvl{\sqrt{c} \; \lVert \vv \rVert _\mathcal{L}}

\icmltitlerunning{Hyperbolic Image-Text Representations}

\begin{document}

\twocolumn[
\icmltitle{Hyperbolic Image-Text Representations}

\begin{icmlauthorlist}
\icmlauthor{Karan Desai}{umich}
\icmlauthor{Maximilian Nickel}{meta}
\icmlauthor{Tanmay Rajpurohit}{ind}
\icmlauthor{Justin Johnson}{umich,meta}
\icmlauthor{Ramakrishna Vedantam}{nyu}
\end{icmlauthorlist}

\icmlaffiliation{umich}{University of Michigan}
\icmlaffiliation{meta}{Meta AI}
\icmlaffiliation{nyu}{New York University}
\icmlaffiliation{ind}{Independent Researcher}

\icmlcorrespondingauthor{Karan Desai}{\texttt{kdexd@umich.edu}}
\icmlkeywords{vision and language, representation learning, riemannian geometry, transformers}

\vskip 0.3in
]

\printAffiliationsAndNotice{KD and Rama did this work while at Meta.}

\begin{abstract}
    Visual and linguistic concepts naturally organize themselves in a hierarchy, where a textual concept ``dog'' entails all images that contain dogs.
    Despite being intuitive, current large-scale vision and language models such as CLIP~\citep{radford2021clip} do not explicitly capture such hierarchy.
    We propose MERU, a contrastive model that yields hyperbolic representations of images and text.
    Hyperbolic spaces have suitable geometric properties to embed tree-like data,
    so MERU can better capture the underlying hierarchy in image-text datasets.
    Our results show that MERU learns a highly interpretable and structured representation space
    while being competitive with CLIP's performance on standard multi-modal tasks
    like image classification and image-text retrieval.
    Our code and models are available at:
    \url{https://github.com/facebookresearch/meru}
\end{abstract}

\input{sections/introduction.tex}

\input{sections/preliminaries.tex}

\input{sections/approach.tex}

\input{sections/experiments.tex}

\input{sections/qualitative.tex}

\input{sections/related.tex}

\input{sections/conclusion.tex}

\clearpage

\bibliography{references}
\bibliographystyle{icml2023}

\clearpage

\section*{Acknowledgments}

We thank our wonderful colleagues for helpful discussions and feedback on the paper
(\emph{in alphabetical order}, grouped by their affiliation during the undertaking of this project):
\begin{compactitem}[\hspace{1pt}--]
    \item \emph{Meta:}
    L\'eon Bottou, Kamalika Chaudhuri, Ricky Chen, Piotr Doll\'ar, Surya Ganguli, Rohit Girdhar, Naman Goyal, Wei-Ning Hsu, Mark Ibrahim, Ishan Misra, Ari Morcos, Devi Parikh, David Schwab, Mannat Singh, Shubham Toshniwal, and Karen Ullrich.
    \item \emph{University of Michigan:}
    Mohamed El Banani, Ang Cao, Daniel Geng, Richard Higgins, Gaurav Kaul, Nilesh Kulkarni, Andrew Lee, Tiange Luo, Andrew Owens, Jeongsoo Park, Chris Rockwell, Dandan Shan, Ayush Shrivastava, and Stella Yu.
    \item \emph{Meta (intern cohort):}
    Desi Ivanova, Lyle Kim, Andre Niyongabo Rubungo, Elizabeth Salesky, and Sagar Vaze.
\end{compactitem}

KD thanks Julius Berner and Steffen Schneider for fruitful discussions over fruity cocktails.
KD also thanks \emph{Cannelle} and many other caf\'es in Ann Arbor for their high-quality espresso shots and many hours of free Wi-Fi.

\appendix
\begin{center}
    \Large\textbf{Appendix}
\end{center}

\input{sections/appendix_approach.tex}

\input{sections/appendix_clip_baseline.tex}

\input{sections/appendix_evaluations.tex}

\clearpage

\input{sections/appendix_traversals.tex}

\clearpage

\end{document}

%% file: sections/introduction.tex
\vspace{-15pt}
\section{Introduction}
\label{sec:meru_intro}

\begin{figure}
    \centering
    \includegraphics[width=\linewidth]{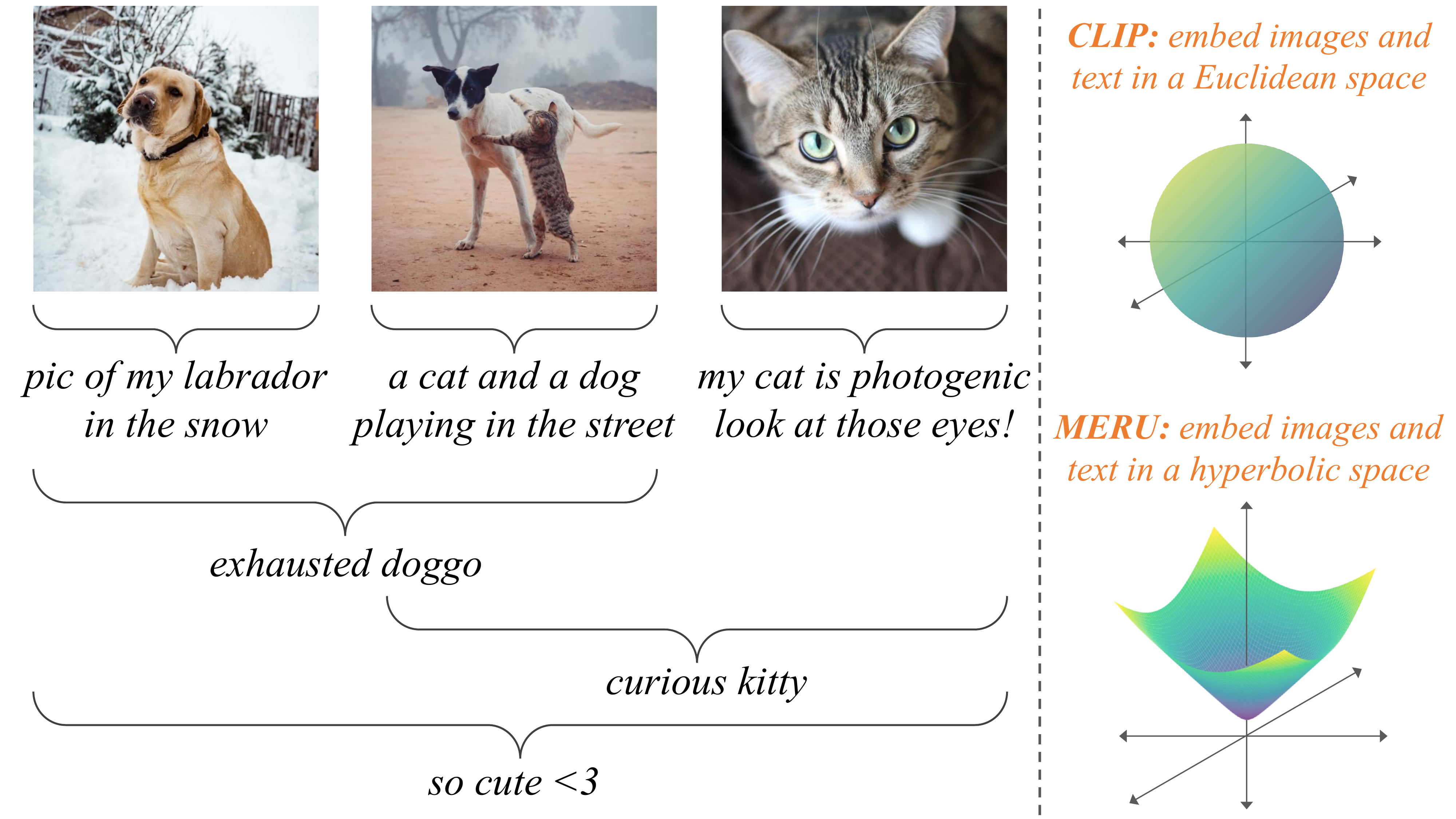}
    \vspace{-20pt}
    \caption{
        \textbf{Hyperbolic image-text representations.}
        \textbf{Left:} Images and text depict \emph{concepts} and can be jointly viewed in a \emph{visual-semantic hierarchy}, wherein text \emph{`exhausted doggo'} is more generic than an image
        (which might have more details like a cat or snow).
        Our method MERU embeds images and text in a hyperbolic space that is well-suited to embed tree-like data.
        \textbf{Right:} Representation manifolds of CLIP (\emph{hypersphere}) and MERU (\emph{hyperboloid}) illustrated in 3D.
        MERU assumes the origin to represent the \emph{most generic concept},
        and embeds text closer to the origin than images.
    }
    \label{fig:meru_intro}
\end{figure}

\paragraph{Visual-semantic hierarchy.}
It is commonly said that \emph{`an image is worth a thousand words'} --
consequently, images contain a lot more information than the sentences which typically describe them.
For example, given the middle image in~\Cref{fig:meru_intro} one might describe it as
\emph{`a cat and a dog playing in the street'}
or with a less specific sentence like \emph{`exhausted doggo'} or \emph{`so cute $<$3'}.
These are not merely diverse descriptions but contain varying levels of
detail about the underlying semantic contents of the image.

As humans, we can reason about the relative detail in each caption,
and can organize such concepts into a meaningful visual-semantic hierarchy~\citep{vendrov2015order}, namely,
\emph{`exhausted doggo'}
$\rightarrow$ \emph{`a cat and a dog playing in the street'}
$\rightarrow$ (\Cref{fig:meru_intro} middle image).
Providing multimodal models access to this inductive bias about vision and language has the potential to improve generalization~\citep{radford2021clip},
interpretability~\citep{selvaraju2017grad} and enable better exploratory data analysis of large-scale datasets~\citep{schuhmann2022laion5b,radford2021clip}.

\paragraph{Vision-language representation learning.}
Approaches such as CLIP~\citep{radford2021clip} and ALIGN~\citep{jia2021align} have catalyzed a lot of recent progress in computer vision
by showing that Transformer-based~\citep{vaswani2017attention} models trained using large amounts of image-text data from the internet can yield transferable representations,
and such models can perform \emph{zero-shot} recognition and retrieval using natural language queries.
All these models represent images and text as vectors in a high-dimensional Euclidean, affine space
and normalize the embeddings to unit $L^2$ norm.
However, such a choice of geometry can find it hard to capture the visual-semantic hierarchy.

An affine Euclidean space treats all embedded points in the same manner,
with the same distance metric being applied to all points~\citep{murphy2013machine}.
Conceptually, this can cause issues when modeling hierarchies --
a \emph{generic} concept (closer to the \emph{root node} of the hierarchy) is close to many other concepts
compared to a \emph{specific} concept (which is only close to its immediate neighbors).
Thus, a Euclidean space can find it hard to pack all the images that say a generic concept \emph{`curious kitty'}
should be close to while also respecting the embedding structure for \emph{`a cat and a dog playing on the street'}.
Such issues are handled naturally by hyperbolic spaces --
the volume increases exponentially as we move away from the origin~\citep{lee2019textbook}, making them a continuous relaxation of trees.
This allows a generic concept (\emph{`cat'}) to have many neighbors by placing it close to the origin~\citep{Nickel2017PoincareF}, and more specific concepts further away.
Thus, distinct specific concepts like images in \Cref{fig:meru_intro} can be far away from each other while being close to some generic concept (\emph{`animal'}).

\paragraph{Hyperbolic representations with MERU.}
In this work, we train the first large-scale contrastive image-text models that yield hyperbolic representations~\citep{Nickel2017PoincareF} -- MERU
\footnote{
Meru is a mountain that symbolizes the \emph{center of all physical, metaphysical, and spiritual universes} in Eastern religions like Hinduism and Buddhism.
Our method is named MERU because the origin of the hyperboloid entails everything and plays a more vital role than in Euclidean (or generally, affine) spaces.
See also: \emph{Mount Semeru, Indonesia}
(Sources -- \url{wikipedia.org/wiki/Mount_Meru} and \url{wikipedia.org/wiki/Semeru})
} that captures the visual-semantic hierarchy (\Cref{fig:meru_intro}).
Our method conceptually resembles current state-of-the-art contrastive methods~\citep{radford2021clip,jia2021align}.
Importantly the hierarchy \emph{emerges} in the representation space, given access only to image-text pairs during training such models.

Practically, MERU confers multiple benefits such as
(a) better performance on image retrieval and classification tasks,
(b) more efficient usage of the embedding space, making it suited for resource-constrained, on-device scenarios,
(c) an interpretable representation space that allows one to infer the relative semantic specificity of images and text.
Overall, we summarize our contributions as follows:

\begin{compactitem}[\hspace{1pt}--]
    \item We introduce MERU, the first implementation of deep hyperbolic representations
    we are aware of, training ViTs \citep{dosovitskiy2021vit} with 12M image-text pairs.
    \item We provide a strong CLIP baseline that outperforms previous
    re-implementations~\citep{mu2021slip} at comparable data scale,
    and systematically demonstrate the benefits of hyperbolic representations over this
    baseline on \emph{zero-shot} retrieval and classification,
    and effectiveness for small embedding dimensions~\citep{kusupati2022matryoshka}.
    \item We perform thorough qualitative analysis with MERU to demonstrate its potential
    for exploratory data analysis of large-scale multimodal datasets.
\end{compactitem}

%% file: sections/preliminaries.tex
\section{Preliminaries}
\label{sec:meru_prelim}

We briefly review Riemannian manifolds (\Cref{subsec:meru_prelim_riemannian})
and essential concepts of hyperbolic geometry (\Cref{subsec:meru_prelim_lorentz}).
For a more thorough treatment of the topic, we refer the reader to textbooks by \citet{ratcliffe2006textbook} and \citet{lee2019textbook}.

\subsection{Riemannian manifolds}
\label{subsec:meru_prelim_riemannian}

A \emph{smooth surface} is a two-dimensional sheet which is \emph{locally Euclidean} --
every point on the surface has a local neighborhood which can be mapped to $\bbR^2$ via a differentiable and invertible function.
\emph{Smooth manifolds} extend the notion of smooth surfaces to higher dimensions.

A \emph{Riemannian manifold} $(\mathcal{M}, g)$ is a smooth manifold $\mathcal{M}$ equipped with a \emph{Riemannian metric} $g$.
The metric $g$ is a collection of inner product functions $g_\vx$ for all points $\vx \in \mathcal{M}$, and varies smoothly over the manifold.
At any point $\vx$, the inner product $g_\vx$ is defined in the \emph{tangent space} $\mathcal{T}_\vx \mathcal{M}$, which is a Euclidean space that gives a linear approximation of $\mathcal{M}$ at $\vx$.
Euclidean space $\bbR^n$ is also a Riemannian manifold, where $g$ is the standard Euclidean inner product.

Our main topic of interest is hyperbolic spaces, which are Riemannian manifolds with \emph{constant negative curvature}.
They are fundamentally different from Euclidean spaces that are \emph{flat} (zero curvature).
A hyperbolic manifold of $n$ dimensions cannot be represented with $\bbR^n$ in a way that preserves both distances and angles.
There are five popular models of hyperbolic geometry that either represent $n$-dimensional hyperbolic spaces either in $\bbR^n$ while distorting distances and/or angles (e.g. Poincar\'e ball model),
or as a sub-manifold of $\bbR^{n+1}$ (e.g. the Lorentz model).

\subsection{Lorentz model of hyperbolic geometry}
\label{subsec:meru_prelim_lorentz}

We use the Lorentz model of hyperbolic geometry for developing MERU.
This model represents a hyperbolic space of $n$ dimensions on the upper half of a two-sheeted hyperboloid in $\bbR^{n+1}$.
See \Cref{fig:meru_intro} for an illustration of $\mathcal{L}^2$ in $\bbR^3$.
Hyperbolic geometry has a direct connection to the study of special relativity theory~\citep{einstein1905elektrodynamik,einstein2015principle}.
We borrow some of its terminology in our discussion --
we refer to the hyperboloid's axis of symmetry as \emph{time dimension}
and all other axes as \emph{space dimensions}~\citep{minkowski1908raum}.
Every vector $\vx \in \bbR^{n+1}$ can be written as $[\vx_{space}, x_{time}]$, where $\vx_{space} \in \bbR^n$ and $x_{time} \in \bbR$.

\paragraph{Definition.}
Let $\langle \cdot , \cdot \rangle$ is Euclidean inner product and $\langle \cdot , \cdot \rangle_\mathcal{L}$ denote the \emph{Lorentzian inner product} that is
induced by the Riemannian metric of the Lorentz model.
For two vectors $\vx, \vy \in \bbR^{n+1}$, it is computed as follows:
\begin{equation}\label{eqn:meru_lorentz_inner}
    \langle \vx,\vy \rangle_\mathcal{L} = \langle \vx_{space},\vy_{space} \rangle - x_{time} \; y_{time}
\end{equation}

\noindent The induced \emph{Lorentzian norm} is $\lVert \vx \rVert _\mathcal{L} = \sqrt{\lvert \langle \vx,\vx \rangle_\mathcal{L} \rvert}$.
The Lorentz model possessing a constant curvature $-c$ is defined as a following set of vectors:
\begin{equation}\label{eqn:meru_set_notation}
    \mathcal{L}^n = \{\vx \in \bbR^{n+1} : \langle \vx,\vx \rangle_\mathcal{L} = \sfrac{-1}{c} \} \; , \; c > 0
\end{equation}

\noindent All vectors in this set satisfy the following constraint:
\begin{equation}\label{eqn:meru_spacetime}
    x_{time} = \sqrt{\sfrac{1}{c} + \lVert \vx_{space} \rVert^2}
\end{equation}

\paragraph{Geodesics.}
A \emph{geodesic} is the shortest path between two points on the manifold.
Geodesics in the Lorentz model are curves traced by the intersection of the hyperboloid with hyperplanes passing through the origin of $\bbR^{n+1}$.
The \emph{Lorentzian distance} between two points $\vx,\vy \in \mathcal{L}^n$ is:
\begin{equation}\label{eqn:meru_distance}
    d_\mathcal{L} (\vx, \vy) = \sqrt{\sfrac{1}{c}} \cdot \cosh^{-1} ( -c \; \langle \vx,\vy \rangle_\mathcal{L} )
\end{equation}

\paragraph{Tangent space.}
The tangent space at some point $\vz \in \mathcal{L}^n$ is a Euclidean space of vectors that are orthogonal to $\vz$ according to the Lorentzian inner product:
\begin{equation}{\label{eqn:meru_tangent_space}}
    \mathcal{T}_\vz \mathcal{L}^n = \{\vv \in \bbR^{n+1} : \langle \vz,\vv \rangle_\mathcal{L} = 0\}
\end{equation}

Any vector in ambient space $\vu \in \bbR^{n+1}$ can be projected to the tangent space $\mathcal{T}_\vz \mathcal{L}^n$ via an orthogonal projection:
\begin{equation}\label{eqn:meru_projection}
    \vv = \text{proj}_\vz (\vu) = \vu + c \; \vz \; \langle \vz,\vu \rangle_\mathcal{L}
\end{equation}

\paragraph{Exponential and logarithmic maps.}
The \emph{exponential map} provides a way to map vectors from tangent spaces onto the manifold.
For a point $\vz$ on the hyperboloid, it is defined as
$\text{expm}_\vz : \mathcal{T}_\vz \mathcal{L}^n \rightarrow \mathcal{L}^n$ with the expression:
\begin{equation}\label{eqn:meru_expmap}
    \vx = \text{expm}_\vz(\vv) = \cosh(\cvl) \; \vz
                         + \frac{\sinh(\cvl)}{\cvl} \; \vv
\end{equation}

Intuitively the exponential map shows how $\mathcal{T}_x \mathcal{L}^n$ \emph{folds} on the manifold.
Its inverse is the \emph{logarithmic map} ($\text{logm}_\vz : \mathcal{L}^n \rightarrow \mathcal{T}_\vz \mathcal{L}^n$),
that maps $\vx$ from the hyperboloid back to $\vv$ in the tangent space:
\begin{equation}\label{eqn:meru_logmap}
    \def\czl{c \; \langle \vz,\vx \rangle _\mathcal{L}}
    \vv = \text{logm}_\vz(\vx) = \frac{\cosh^{-1} (-\czl)}{\sqrt{\left(\czl \right)^2 - 1}} \; \text{proj}_\vz (\vx)
\end{equation}

For our approach, we will only consider these maps where $\vz$ is the origin of the hyperboloid ($\vO = [\mathbf{0}, \sqrt{\sfrac{1}{c}}]$).

%% file: sections/approach.tex
\section{Approach}
\label{sec:meru_approach}

In this section, we discuss the modeling pipeline and learning objectives of MERU
to learn hyperbolic representations of images and text.
We use the tools of hyperbolic geometry introduced in \Cref{sec:meru_prelim} throughout our discussion.

\begin{figure}
    \centering
    \includegraphics[width=\linewidth]{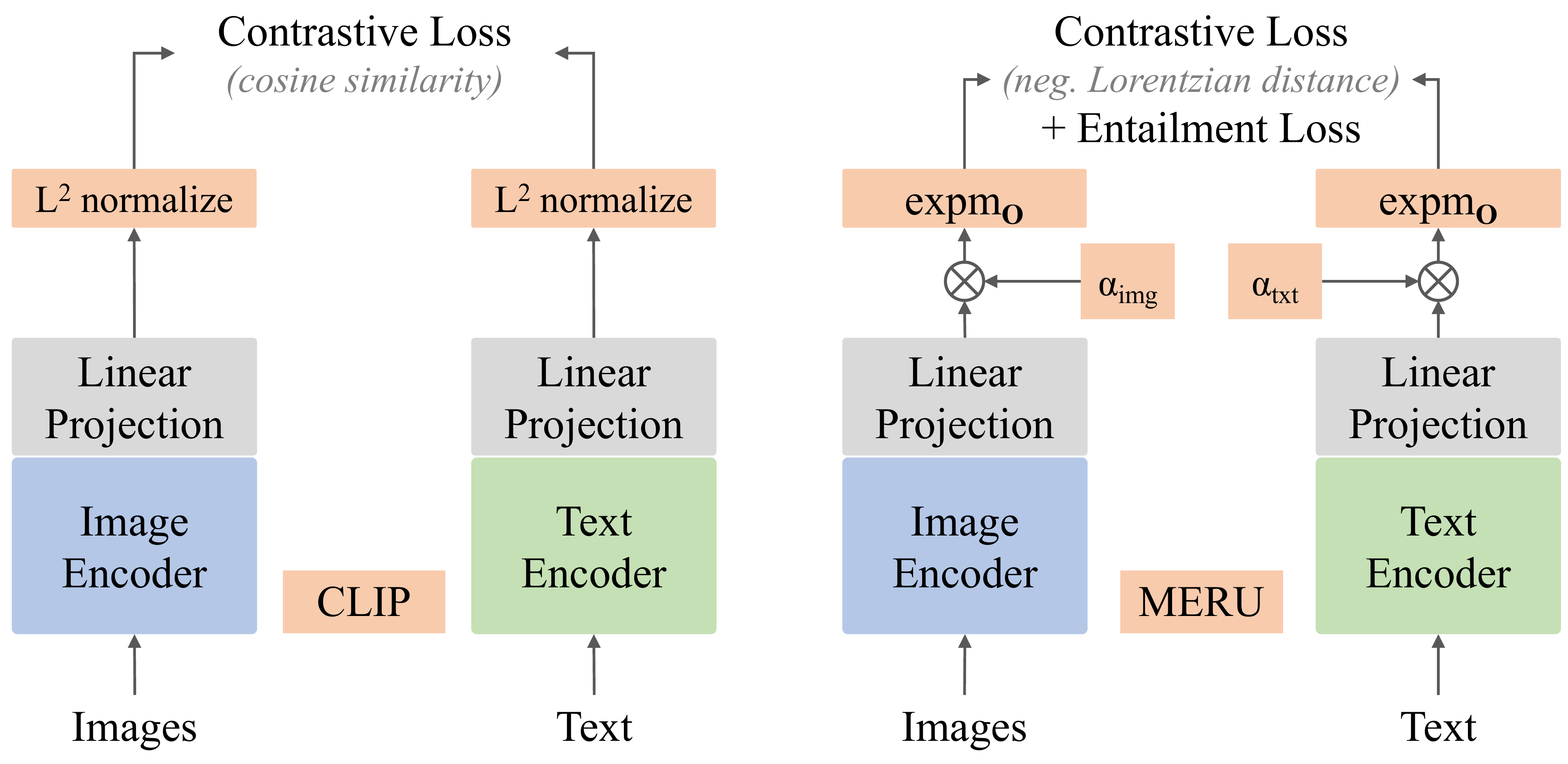}
    \vspace{-10pt}
    \caption{
        \textbf{MERU model design:}
        MERU comprises similar architectural components as standard image-text contrastive models like CLIP.
        While CLIP projects the embeddings to a unit hypersphere,
        MERU lifts them onto the Lorentz hyperboloid using the exponential map.
        The contrastive loss uses the negative of Lorentzian distance as a similarity metric,
        and a special entailment loss enforces \emph{`text entails image'} partial order in the representation space.
    }
    \label{fig:meru_model}
\end{figure}

Our model design is inspired by CLIP~\citep{radford2021clip} due to its simplicity and scalability.
As shown in \Cref{fig:meru_model}, we process images and text using two separate encoders,
and obtain embedding vectors of a fixed dimension $n$.
Beyond this, there are two crucial design choices:
(1) transferring embeddings from Euclidean space to the Lorentz hyperboloid, and
(2) designing suitable training objectives that induce semantics and structure in the representation space.

\paragraph{Lifting embeddings onto the hyperboloid.}
Let the embedding vector from the image encoder or text encoder, after linear projection be $\vv_{enc} \in \mathbb{R}^{n}$.
We need to apply a transformation such that the resulting vector $\vx$ lies on the Lorentz hyperboloid $\mathcal{L}^n$ in $\mathbb{R}^{n+1}$.
Let the vector $\vv = [\vv_{enc}, 0] \in \mathbb{R}^{n+1}$.
We observe that $\vv$ belongs to the tangent space at the hyperboloid origin $\vO$, as \cref{eqn:meru_tangent_space} is satisfied:
$\langle \vO, \vv \rangle _\mathcal{L} = 0$.
Thus, we parameterize \emph{only} the \emph{space} components of the Lorentz model ($\vv_{enc} = \vv_{space}$).
Due to such parameterization, we can simplify the exponential map from \cref{eqn:meru_expmap} by writing only \emph{space} components:
\begin{equation}
    \vx_{space}
        = \cosh (\sqrt{c} \; \lVert \vv \rVert _{\mathcal{L}} ) \mathbf{0}
        + \frac{ \sinh(\sqrt{c} \; \lVert \vv \rVert _{\mathcal{L}} )}{\sqrt{c} \; \lVert \vv \rVert _{\mathcal{L}}} \vv_{space} \nonumber
\end{equation}

The first term reduces to $\mathbf{0}$.
Moreover, the Lorentzian norm of $\vv$ simplifies to the Euclidean norm of \emph{space} components:
$\lVert \vv \rVert _{\mathcal{L}}^2 = \langle \vv,\vv \rangle_\mathcal{L} = \langle \vv_{space},\vv_{space} \rangle - 0 = \lVert \vv_{space} \rVert ^2$.
This substitution simplifies the above equation as follows:
\begin{equation}\label{eqn:meru_simple_expmap}
    \vx_{space} = \frac{\sinh(\sqrt{c} \; \lVert \vv_{space} \rVert)}{\sqrt{c} \; \lVert \vv_{space} \rVert} \vv_{space}
\end{equation}

The corresponding \emph{time} component $x_{time}$ can be computed from $\vx_{space}$ using \cref{eqn:meru_spacetime},
the resulting $\vx$ \emph{always} lies on the hyperboloid.
This eliminates the need for an orthogonal projection (\cref{eqn:meru_projection}) and simplifies the exponential map.
Our parameterization is simpler than previous work which parameterizes vectors in full ambient space $\mathbb{R}^{n+1}$
\citep{law2019lorentzian,nickel2018lorentz,le2019inferringch}.

\paragraph{Preventing numerical overflow.}
The exponential map scales $\vv_{space}$ using an exponential operator.
According to CLIP-style weight initialization, $\vv_{space} \in \bbR^n$ would have an expected norm $=\sqrt{n}$.
After exponential map, it becomes $e^{\sqrt{n}}$, which can be numerically large (\eg $n=512$ and $c=1$ gives $||\vx_{space}|| \approx 6.7 \times 10^{10}$).

To fix this issue, we \emph{scale} all vectors $\vv_{space}$ in a batch before applying $\text{expm}_{\vO}$ using two learnable scalars $\alpha_{img}$ and $\alpha_{txt}$.
These are initialized to $\sqrt{\sfrac{1}{n}}$ so that the Euclidean embeddings have an expected unit norm at initialization.
We learn these scalars in logarithmic space to avoid collapsing all embeddings to zero.
After training, they can be absorbed into the preceding projection layers.

\paragraph{Learning structured embeddings.}
Having lifted standard Euclidean embeddings onto the hyperboloid,
we next discuss the losses used to enforce structure and semantics in representations learned by MERU.
Recall that our motivation is to capture the visual-semantic hierarchy (\Cref{fig:meru_intro}) to better inform the generalization capabilities of vision-language models.
For this, an important desideratum is a meaningful notion of distance between semantically similar text and image pairs.
We also want to induce a partial order between text and images as per the visual-semantic hierarchy to have better interpretability.
We do this with a modified version of an entailment loss proposed by \citet{le2019inferringch},
that works for arbitrary hyperboloid curvatures $-c$.

\subsection{Contrastive learning formulation}
\label{subsec:meru_contrastive}

Given a batch of size $B$ of image-text pairs and any $j^{th}$ instance in batch,
its image embedding $\vy_j$ and text embedding $\vx_j$ form a \emph{positive} pair,
whereas the remaining $B-1$ text embeddings in the batch $\vx_i (i \neq j)$ form \emph{negative} pairs.

In contrastive learning, we compute the negative Lorentzian distance as a similarity measure (\cref{eqn:meru_distance})
for all $B$ pairs in the batch.
These logits are divided by a temperature $\tau$ and apply a softmax operator.
Similarly, we also consider a contrastive loss for text, that treats images as negatives.
The total loss $\mathcal{L}_{cont}$ is the average of these two losses computed for every image-text pair in the batch.
Our implementation of the contrastive loss is the same as the multi-class N-pair loss from~\citep{Sohn2016-vp} used in CLIP~\citep{radford2021clip}
with the crucial difference being that we compute distances on the hyperboloid instead of cosine similarity.

\subsection{Entailment loss}
\label{subsec:meru_entailment}

\begin{figure}
    \centering
    \includegraphics[width=0.8\linewidth]{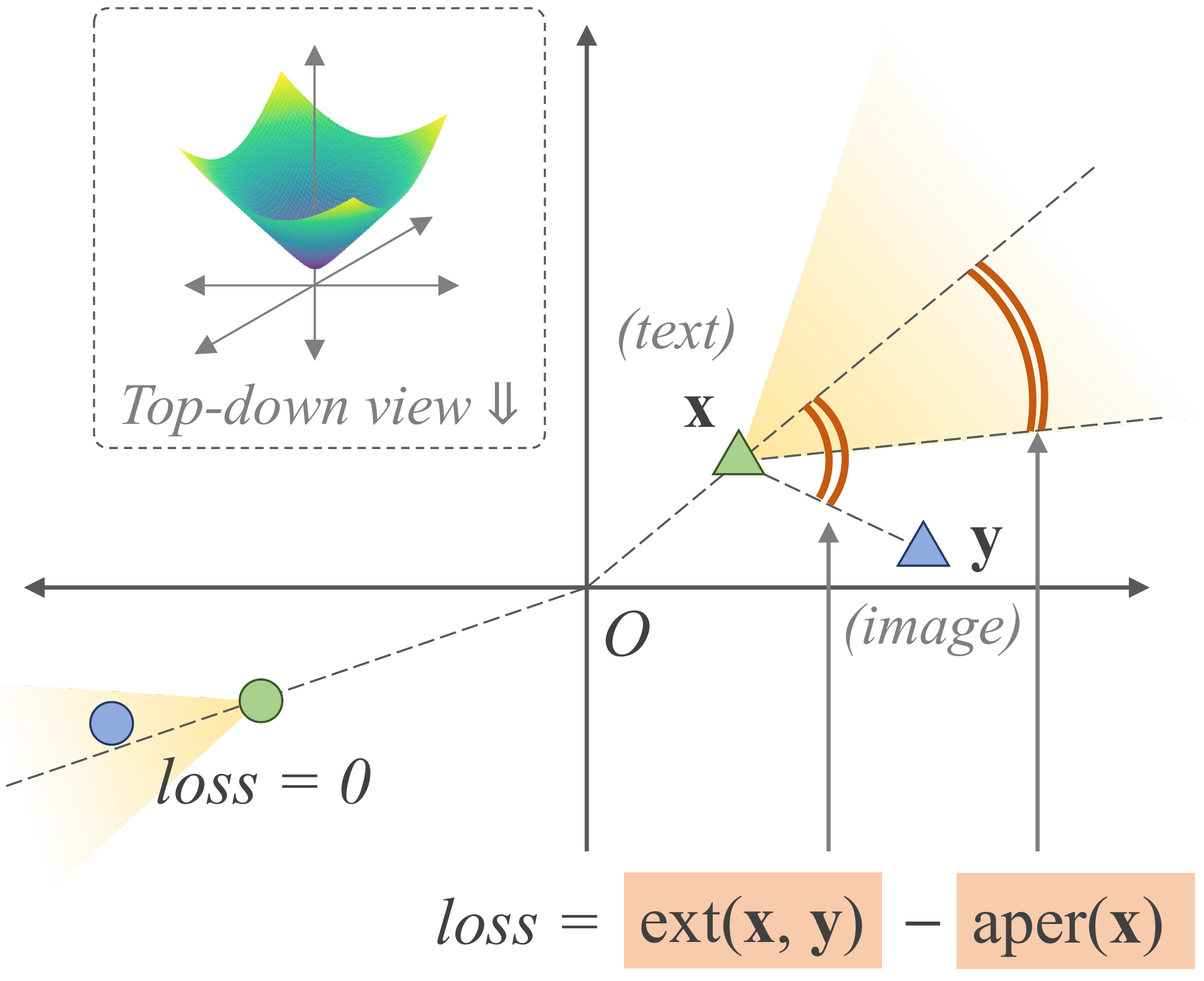}
    \vspace{-10pt}
    \caption{
        \textbf{Entailment loss (illustrated for $\mathcal{L}^2$):}
        This loss pushes image embedding $\vy$ inside an imaginary cone projected by the paired text embedding $\vx$,
        and is implemented as the difference of exterior angle $\angle O \vx \vy$ and half aperture of the cone.
        Loss is zero if the image embedding is already inside the cone \emph{(left quadrant)}.
    }
    \label{fig:meru_entail_loss}
\end{figure}

In addition to the contrastive loss, we adapt an entailment loss~\citep{le2019inferringch,ganea2018entailment} to enforce partial order relationships
between paired text and images.
\citet{ganea2018entailment} is more different from ours since they parameterize their representations according to the Poincar\'e ball model.
\citet{le2019inferringch} use this loss with a fixed $c=1$, which we extend to
handle arbitrary, learned curvatures.

Refer \Cref{fig:meru_entail_loss} for an illustration in two dimensions.
Let $\vx$ and $\vy$ denote the text and image embeddings of a single image-text pair.
Note that the encoders only give $\vx_{space}$ and $\vy_{space}$ according to our parameterization.
Corresponding $x_{time}$ and $y_{time}$ are calculated using \cref{eqn:meru_spacetime}.
We define an \emph{entailment cone} for each $\vx$, which narrows as we go farther from the origin.
This cone is defined by the half-aperture:
\begin{equation}
    \text{aper}(\vx) = \sin^{-1} \left( \frac{2K}{\sqrt{c} \; \lVert \vx_{space} \rVert} \right)
\end{equation}

where a constant $K=0.1$ is used for setting boundary conditions near the origin.
We now aim to identify and penalize when the paired image embedding $\vy$ lies outside the entailment cone.
For this, we measure the exterior angle $\text{ext}(\vx, \vy) = \pi - \angle \vO \vx \vy$
as shown in \Cref{fig:meru_entail_loss}:
\begin{equation}
    \def\cvl{c \; \langle \vx,\vy \rangle _\mathcal{L}}
    \text{ext}(\vx, \vy) = \cos^{-1} \left( \frac{y_{time} + x_{time} \; \cvl{}}{\lVert \vx_{space} \rVert \sqrt{\left( \cvl \right)^2 - 1}} \right)
\end{equation}

If the exterior angle is smaller than the aperture,
then the partial order relation between $\vx$ and $\vy$
is already satisfied and we need not penalize anything,
while if the angle is greater, we need to reduce it.
This is captured by the following loss function
(written below for a single $\vx$, $\vy$ pair):
\begin{equation}\label{eqn:meru_entailment}
    \mathcal{L}_{entail}(\vx, \vy) = \max(0, \; \text{ext}(\vx, \vy) - \text{aper}(\vx))
\end{equation}

We provide exact derivations of the above equations for half-aperture and exterior angle in \Cref{sec:meru_appendix_approach}.
Overall, our total loss is $\mathcal{L}_{cont} + \lambda \mathcal{L}_{entail}$ averaged over each minibatch.

%% file: sections/experiments.tex
\section{Experiments}
\label{sec:meru_experiments}

Our main objective in the experiments is to establish the competitiveness of hyperbolic representations of MERU as compared to Euclidean representations obtained from CLIP-style models.
To this end, we train models using large amounts of image-text pairs and transfer them to a variety of image classification and retrieval tasks.

\subsection{Training details}
\label{subsec:meru_training_details}

\paragraph{Baselines.}
We primarily compare with CLIP~\citep{radford2021clip}, that embeds images and text on a unit hypersphere in a Euclidean space.
CLIP was trained using a private dataset of 400M image-text pairs.
Several follow-up works re-implement CLIP and use publicly accessible datasets like YFCC~\citep{yfcc100m},
Conceptual Captions~\citep{sharma2018conceptual,changpinyo2021conceptual}, and LAION~\citep{schuhmann2021laion400m,schuhmann2022laion5b};
notable examples are OpenCLIP~\citep{ilharco2021openclip}, SLIP~\citep{mu2021slip}, DeCLIP~\citep{li2021declip}, and FILIP~\citep{yao2022filip}.
We develop our CLIP baseline and train it using a \emph{single} public dataset -- RedCaps~\citep{desai2021redcaps} -- for easier reproducibility.
Our smallest model trains using $8\times$ V100 GPUs in \emph{less than one day} and significantly outperforms recent CLIP re-implementations that use YFCC~\citep{mu2021slip}.

Refer \Cref{sec:meru_appendix_clip_baseline} for details about our CLIP baseline.
Our implementation is based on PyTorch~\citep{pytorch} and \texttt{timm}~\cite{rw2019timm} libraries.

\paragraph{Models.}
We use the Vision Transformer~\citep{dosovitskiy2021vit} as image encoder,
considering three models of varying capacity -- ViT-S~\citep{touvron2021training,chen2021mocov3}, ViT-B, and ViT-L.
All use a patch size of 16.
The text encoder is same as CLIP -- a 12-layer, 512 dimensions wide Transformer~\citep{vaswani2017attention} language model.
We use the same byte-pair encoding tokenizer~\citep{sennrich2016bpe} as CLIP, and truncate input text at maximum 77 tokens.

\paragraph{Data augmentation.}
We randomly crop 50--100\% area of images and resize them to $224 \times 224$, following \citep{mu2021slip}.
For text augmentation, we randomly \emph{prefix} the subreddit names to captions as \texttt{`\{subreddit\} : \{caption\}'}.

\paragraph{Initialization.}
We initialize image/text encoders in the same style as CLIP, except for one change:
we use a \emph{sine-cosine} position embedding in ViT, like \citep{chen2021mocov3,he2022mae},
and keep it frozen while training.
We initialize the softmax temperature as $\tau=0.07$ and clamp it to a minimum value of $0.01$.
For MERU, we initialize the learnable projection scalars $\alpha_{img} = \alpha_{txt} = \sfrac{1}{\sqrt{512}}$,
the curvature parameter $c=1.0$ and clamp it in $[0.1, 10.0]$ to prevent training instability.
All scalars are learned in logarithmic space as $log(1/\tau)$, $log(c)$, and $log(\alpha)$.

\paragraph{Optimization.}
We use AdamW~\citep{loshchilov2019decoupled} with weight decay $0.2$ and $(\beta_1,\beta_2)=(0.9,0.98)$.
We disable weight decay for all gains, biases, and learnable scalars.
All models are trained for $120K$ iterations with batch size $2048$ ($\approx20$ epochs).
The maximum learning rate is $5\times10^{-4}$, increased linearly for the first $4K$ iterations,
followed by cosine decay to zero~\citep{loshchilov2016sgdr}.
We use mixed precision~\citep{micikevicius2018mixed} to accelerate training,
except computing exponential map and losses for MERU in FP32 precision for numerical stability.

\paragraph{Loss multiplier ($\lambda$) for MERU.}
We set $\lambda=0.2$ by running a hyperparameter sweep with ViT-B/16 models for one epoch.
Some $\lambda>0$ is necessary to induce partial order structure,
however, quantitative performance is less sensitive to the choice of $\lambda \in [0.01, 0.3]$;
Higher values of $\lambda$ strongly regularize against the contrastive loss and hurt performance.

\subsection{Image and text retrieval}
\label{subsec:meru_retrieval_eval}

\input{figtabs/retrieval.tex}

CLIP-style contrastive models perform image and text retrieval within batch during training,
making them ideal for retrieval-related downstream applications.
We evaluate the retrieval capabilities of MERU as compared to CLIP on two established benchmarks:
COCO and Flickr30K \citep{chen2015microsoft,young2014flickr}, that comprise 5000 and 1000 images respectively and five captions per image.
COCO evaluation uses the \texttt{val2017} split while Flickr30K uses the \texttt{test} split defined by \citet{karpathy2015splits}.
We perform \emph{zero-shot transfer}, without any additional training using these datasets.
We \emph{squeeze} images to 224$\times$224 pixels before processing them through the image encoder.

\paragraph{Inference with MERU.}
We rank a pool of candidate image/text embeddings for retrieval
in decreasing order of their Lorentzian inner product (\cref{eqn:meru_lorentz_inner}) with a text/image query embedding.
Some transfer tasks like \emph{open-vocabulary detection} \citep{zareian2021ovrcnn,gu2022vild} may require calibrated scores,
for them we recommend using the training procedure -- compute the negative of distance (\cref{eqn:meru_distance}), divide by temperature and apply a softmax classifier.

\paragraph{Results.}
\Cref{tab:meru_retrieval} reports recall@\{5,10\} of MERU and the reproduced CLIP baselines on these benchmarks.
Hyperbolic representations of MERU mostly perform best for all tasks and models (except Flickr30K text retrieval with ViT-B/16).
This is encouraging evidence that hyperbolic spaces have suitable geometric properties to learn strong representations for retrieval applications.
Surprisingly, increasing model size (ViT-B/16 $\rightarrow$ ViT-L/16) does not improve image retrieval for both, MERU and CLIP.
We believe that better quality of text queries is important for image retrieval -- increasing the size of text encoder can alleviate this issue.

\input{figtabs/zero_shot_cls.tex}

\subsection{Image classification}
\label{subsec:meru_classification_eval}

Learning from language supervision allows CLIP to perform \emph{zero-shot} image classification,
wherein one may specify label sets as text queries~\citep{write_a_classifier} instead of using pre-defined ontologies~\citep{miller1992wordnet,deng2009imagenet}.
Classifier weights are obtained by embedding label-based queries (also called \emph{prompts}) using the text encoder.

In this section, we evaluate MERU on 20 image classification benchmarks covering a wide variety of visual concepts.
These are used by \citet{radford2021clip} and several follow-up works~\citep{mu2021slip,yao2022filip,li2021declip},
and available with open-source libraries like \texttt{tensorflow-datasets} and \texttt{torchvision}
\footnote{\url{tensorflow.org/datasets} and \url{pytorch.org/vision}}.
We report top-1 mean per-class accuracy for all datasets to account for any label imbalance.
We use multiple prompts per dataset, most of which follow \citet{radford2021clip}.
We \emph{ensemble} these multiple prompts by averaging their embeddings before lifting them onto the hyperboloid (\cref{eqn:meru_simple_expmap}).
See \Cref{tab:meru_appendix_datasets,tab:meru_appendix_prompts} in Appendix for details about datasets and prompts.

\paragraph{Results.}
\Cref{tab:meru_zeroshot_cls} shows strong transfer performance of MERU, matching or outperforming CLIP on 13 out of 16 standard datasets.
While MERU is effective on recall-based measures (\Cref{tab:meru_retrieval}),
it does not come at the expense of precision~\citep{murphy2013machine}.
Overall, hyperbolic representations from MERU are competitive with their Euclidean counterparts across varying model architectures (ViT-S/B/L).

All models have \emph{near-random} performance on four benchmarks.
Concepts in these datasets have low coverage in RedCaps,
like PCAM~\citep{veeling2018pcam} containing medical scans,
or SST2~\citep{socher2013sst} containing movie reviews rendered as images.
Performance on these benchmarks does not indicate the efficacy of our RedCaps-trained models;
using larger training datasets like LAION~\citep{schuhmann2022laion5b} may yield meaningful trends.

\subsection{Resource-constrained deployment}
\label{subsec:meru_embedding_width}

We hypothesize that embeddings that capture a rich visual-semantic hierarchy can use the volume in the representation space more efficiently.
This is useful for on-device deployments with runtime or memory constraints that necessitate low-dimensional embeddings~\citep{kusupati2022matryoshka}.

\begin{table}[t]
    \footnotesize
    \setlength{\tabcolsep}{5.5pt}
    \newcommand{\best}[0]{\cellcolor{ForestGreen!30}\bf}

    \caption{
        \textbf{MERU and CLIP with different embedding widths.}
        We report \emph{zero-shot} COCO recall@5 and ImageNet top-1 accuracy.
        MERU outperforms CLIP at lower embedding widths.
    }
    \label{tab:meru_embedding_width}
    \begin{tabularx}{\linewidth}{cX ccccc}
        \toprule
        ~ & ~ & \multicolumn{5}{c}{Embedding width} \\
        \cmidrule{3-7}
        ~ & ~ & 512 & 256 & 128 & 96 & 64 \\
        \midrule
        \cellcolor{Apricot!50} & CLIP & 31.7 & 31.8 & 31.4 & 29.6 & 25.7 \\
        \cellcolor{Apricot!50} \multirow{-2}{*}{\shortstack{COCO \\ \emph{text$\rightarrow$image}}}
        & MERU & \best 32.6 & \best 32.7 & \best 32.7 & \best 31.0 & \best 26.5 \\
        \midrule
        \cellcolor{Apricot!50} & CLIP & 40.6 & 41.0 & 40.4 & 37.9 & 33.3 \\
        \cellcolor{Apricot!50} \multirow{-2}{*}{\shortstack{COCO \\ \emph{image$\rightarrow$text}}}
        & MERU & \best 41.9 & \best 42.5 & \best 42.6 & \best 40.5 & \best 34.2 \\
        \midrule
        \cellcolor{Apricot!50} & CLIP & 38.4 & 38.3 & 37.9 & 35.2 & 30.2 \\
        \cellcolor{Apricot!50} \multirow{-2}{*}{\shortstack{ImageNet}}
        & MERU & \best 38.8 & \best 38.8 & \best 38.8 & \best 37.3 & \best 32.3 \\
        \bottomrule
    \end{tabularx}
\end{table}

To verify this hypothesis, we train MERU and CLIP models that output 64--512 dimensions wide embeddings.
We initialize the encoders from ViT-L/16 models (\Cref{tab:meru_zeroshot_cls}, last two rows) to reduce compute requirements,
keep them frozen, and re-initialize projection layers and learnable scalars.
We train for $30K$ iterations and evaluate on \emph{zero-shot} COCO retrieval and ImageNet~\citep{Russakovsky2014ImageNetLS} classification.
Results in \Cref{tab:meru_embedding_width} show that MERU consistently performs better at low embedding widths.
This indicates that hyperbolic embeddings may be an appealing solution for resource-constrained on-device applications.

\subsection{Ablations}
\label{subsec:meru_ablations}

In this section, we ablate our MERU models to observe the impact of our design choices.
We experiment with two image encoders, ViT-B/16 and ViT-L/16,
and evaluate for zero-shot COCO retrieval and ImageNet classification.

\begin{table}
    \footnotesize
    \setlength{\tabcolsep}{3pt}
    \caption{
        \textbf{MERU ablations.}
        We ablate three design choices of MERU and report \emph{zero-shot} COCO recall@5 and ImageNet top-1 accuracy.
        Our design choices are crucial for training stability when using a larger model (ViT-L/16) with MERU.
    }
    \label{tab:meru_ablations}
    \begin{tabularx}{\linewidth}{X ccc}
        \toprule
        & \shortstack{\\COCO\\\emph{text$\rightarrow$image}} & \shortstack{\\COCO\\\emph{image$\rightarrow$text}} & ImageNet \\
        \midrule
        \rowcolor{Apricot!50} MERU ViT-B/16 & 33.2 & 41.8 & 37.5 \\
        \textbf{1.} \hspace{2pt} \emph{no entailment loss} & 33.7 & 43.5 & 36.2 \\
        \textbf{2.} \hspace{2pt} \emph{fixed} $c=1$ & 33.2 & 42.1 & 37.9 \\
        \textbf{3.} \hspace{2pt} $\langle \cdot,\cdot \rangle _\mathcal{L}$ \emph{in contrastive} & 32.6 & 42.3 & 37.3 \\
        \midrule
        \rowcolor{Apricot!50} MERU ViT-L/16 & 32.6 & 41.9 & 38.8 \\
        \textbf{1.} \hspace{2pt} \emph{no entailment loss} & 32.7 & 42.2 & 33.8 \\
        \textbf{2.} \hspace{2pt} \emph{fixed} $c=1$ & 0.9 & 0.9 & 0.7 \\
        \textbf{3.} \hspace{2pt} $\langle \cdot,\cdot \rangle _\mathcal{L}$ \emph{in contrastive} & \multicolumn{3}{c}{-- \quad \emph{did not converge} \quad --} \\
        \bottomrule
    \end{tabularx}
\end{table}

Specifically, we train three ablations with the default hyperparameters (\Cref{subsec:meru_training_details}),
except having one difference each.
Results are shown in \Cref{tab:meru_ablations} above.

\paragraph{No entailment loss:}
We only use the contrastive loss for training this ablation.
This effectively means setting $\lambda=0$.
Note that this ablation is mathematically impossible for
CLIP as there is no obvious notion of entailment that can be defined
when all the embeddings have a unit norm.
Disabling the entailment loss is mostly inconsequential to MERU's performance.
This shows that choosing a hyperbolic space is sufficient to improve \emph{quantitative} performance over CLIP.
Entailment loss is crucial for better structure and interpretability, as will be discussed in \Cref{sec:meru_qualitative}.

\paragraph{Fixed curvature parameter:}
Recall that our models treat the hyperboloid curvature as a learnable parameter during training.
Here we train an ablation using a fixed curvature $c=1$.
This has negligible impact on MERU ViT-B/16, but learning curvature is crucial when scaling model size --
MERU ViT-L/16 model with fixed $c=1$ is difficult to optimize and performs poorly on convergence.
As far as we are aware, no prior work learns the curvature~\citep{atigh2022hyperbolicsegm,Khrulkov2020HyperbolicIE,nickel2018lorentz}.

\paragraph{Lorentzian inner product in contrastive loss:}
CLIP-style contrastive loss uses the inner product defined on the hypersphere (cosine similarity).
Similarly, we consider the \emph{Lorentzian inner product} (\cref{eqn:meru_lorentz_inner}) in the contrastive loss instead of negative Lorentzian distance.
With this, MERU ViT-L/16 is difficult to train.
Loss diverges due to numerical overflow, as
Lorentzian inner product is numerically large and unbounded in $(-\infty, \sfrac{-1}{c}]$,
unlike cosine similarity $\in [-1, 1]$.
Lorentzian distance applies a logarithmic operator ($cosh^{-1}$) on the Lorentzian inner product,
slowing down its growth and hence improving numerical stability.

We hope these ablations serve as guidelines for work in other domains that study hyperbolic representation learning.

%% file: figtabs/retrieval.tex
\begin{table}[t]
    \centering
    \footnotesize
    \newcolumntype{Z}{>{\centering\arraybackslash}X}
    \setlength\tabcolsep{3pt}
    \newcommand{\best}[0]{\cellcolor{ForestGreen!30}\bf}

    \caption{\textbf{Zero-shot image and text retrieval.}
        Best performance in every column is highlighted in \textcolor{ForestGreen}{green}.
        MERU performs better than CLIP for both datasets and across all model sizes.
    }
    \label{tab:meru_retrieval}
    \begin{tabularx}{\linewidth}{c X cc cc cc cc}
    \toprule
    && \multicolumn{4}{c}{\emph{text $\rightarrow$ image}} & \multicolumn{4}{c}{\emph{image $\rightarrow$ text}} \\
    \cmidrule(lr){3-6} \cmidrule(lr){7-10}
    && \multicolumn{2}{c}{COCO} & \multicolumn{2}{c}{Flickr} & \multicolumn{2}{c}{COCO} & \multicolumn{2}{c}{Flickr} \\
    \cmidrule(lr){3-4} \cmidrule(lr){5-6} \cmidrule(lr){7-8} \cmidrule(lr){9-10}
    && R5 & R10 & R5 & R10 & R5 & R10 & R5 & R10 \\
    \midrule
    \cellcolor{Apricot!50}
    & CLIP & 29.9 & 40.1 & 35.3 & 46.1 & 37.5 & 48.1 & 42.1 & 54.7 \\
    \cellcolor{Apricot!50} \multirow{-2}{*}{\shortstack{ViT \\ S/16}}
    & MERU & \bf 30.5 & \bf 40.9 & \bf 37.1 & \bf 47.4 & \bf 39.0 & \bf 50.5 & \bf 43.5 & \bf 55.2 \\
    \midrule
    \cellcolor{Apricot!50}
    & CLIP & 32.9 & 43.3 & 40.3 & 51.0 & 41.4 & 52.7 & \bf 50.2 & \bf 60.2 \\
    \cellcolor{Apricot!50} \multirow{-2}{*}{\shortstack{ViT \\ B/16}}
    & MERU & \best 33.2 & \best 44.0 & \best 41.1 & \best 51.6 & \bf 41.8 & \bf 52.9 & 48.1 & 58.9 \\
    \midrule
    \cellcolor{Apricot!50}
    & CLIP & 31.7 & 42.2 & 39.0 & 49.3 & 40.6 & 51.3 & 47.8 & 58.5 \\
    \cellcolor{Apricot!50} \multirow{-2}{*}{\shortstack{ViT \\ L/16}}
    & MERU & \bf 32.6 & \bf 43.0 & \bf 39.6 & \bf 50.3 & \best 41.9 & \best 53.3 & \best 50.3 & \best 60.6 \\
    \bottomrule
    \end{tabularx}
    \vspace{-10pt}
\end{table}

%% file: figtabs/zero_shot_cls.tex
\begin{table*}[t]
    \footnotesize
    \setlength{\tabcolsep}{3pt}
    \newcommand{\xx}[0]{\cellcolor{Gray!30}}
    \newcommand{\BF}[0]{\cellcolor{ForestGreen!30}\bf}

    \caption{
        \textbf{Zero-shot image classification.}
        We train MERU and CLIP models with varying parameter counts and transfer them \emph{zero-shot} to 20 image classification datasets.
        Best performance in every column is highlighted in \textcolor{ForestGreen}{green}.
        Hyperbolic representations from MERU match or outperform CLIP on 13 out of the first 16 datasets.
        On the last four datasets (\textcolor{Gray}{gray columns}),
        both MERU and CLIP have \emph{near-random} performance, as concepts in these datasets are not adequately covered in the training data.
    }
    \label{tab:meru_zeroshot_cls}
    \begin{tabularx}{\linewidth}{c X c cccccccccccccccccccc}
        \toprule
            &&
            \rotatebox[origin=lb]{90}{\smash{ImageNet}} &
            \rotatebox[origin=lb]{90}{\smash{Food-101}} &
            \rotatebox[origin=lb]{90}{\smash{CIFAR-10}} &
            \rotatebox[origin=lb]{90}{\smash{CIFAR-100}} &
            \rotatebox[origin=lb]{90}{\smash{CUB}} &
            \rotatebox[origin=lb]{90}{\smash{SUN397}} &
            \rotatebox[origin=lb]{90}{\smash{Cars}} &
            \rotatebox[origin=lb]{90}{\smash{Aircraft}} &
            \rotatebox[origin=lb]{90}{\smash{DTD}} &
            \rotatebox[origin=lb]{90}{\smash{Pets}} &
            \rotatebox[origin=lb]{90}{\smash{Caltech-101}} &
            \rotatebox[origin=lb]{90}{\smash{Flowers}} &
            \rotatebox[origin=lb]{90}{\smash{STL-10}} &
            \rotatebox[origin=lb]{90}{\smash{EuroSAT}} &
            \rotatebox[origin=lb]{90}{\smash{RESISC45}} &
            \rotatebox[origin=lb]{90}{\smash{Country211}} &
            \xx \rotatebox[origin=lb]{90}{\smash{MNIST}} &
            \xx \rotatebox[origin=lb]{90}{\smash{CLEVR}} &
            \xx \rotatebox[origin=lb]{90}{\smash{PCAM}} &
            \xx \rotatebox[origin=lb]{90}{\smash{SST2}} \\
        \midrule
        \cellcolor{Apricot!50}
        & CLIP &     34.3 &     74.5 & \bf 60.1 &     24.4 & \bf 33.8 &     27.5 & \bf 11.3 & \bf 1.4 &     15.0 & \bf 73.7 &     63.9 &     47.0 &     88.2 &     18.6 &     31.4 & \bf 5.2 & \xx 10.0 & \xx 19.4 & \xx 50.2 & \xx 50.1 \\
        \cellcolor{Apricot!50} \multirow{-2}{*}{\shortstack{ViT \\ S/16}}
        & MERU & \bf 34.4 & \bf 75.6 &     52.0 & \bf 24.7 &     33.7 & \bf 28.0 &     11.1 &     1.3 & \bf 16.2 &     72.3 & \bf 64.1 & \bf 49.2 & \bf 91.1 & \BF 30.4 & \bf 32.0 &     4.8 & \xx  7.5 & \xx 14.5 & \xx 51.0 & \xx 50.0 \\
        \midrule
        \cellcolor{Apricot!50}
        & CLIP & \bf 37.9 & \bf 78.9 &     65.5 & \bf 33.4 &     33.3 &     29.8 & \bf 14.4 &     1.4 &     17.0 &     77.9 & \BF 68.5 &     50.9 &     92.2 &     25.6 &     31.0 & \bf 5.8 & \xx 10.4 & \xx 14.3 & \xx 54.1 & \xx 51.5 \\
        \cellcolor{Apricot!50} \multirow{-2}{*}{\shortstack{ViT \\ B/16}}
        & MERU &     37.5 &     78.8 & \bf 67.7 &     32.7 & \bf 34.8 & \bf 30.9 &     14.0 & \bf 1.7 & \BF 17.2 & \bf 79.3 & \BF 68.5 & \BF 52.1 & \bf 92.5 & \bf 30.2 & \bf 34.5 &     5.6 & \xx 13.0 & \xx 13.5 & \xx 49.8 & \xx 49.9 \\
        \midrule
        \cellcolor{Apricot!50}
        & CLIP &     38.4 &     80.3 & \BF 72.0 & \BF 36.4 &     36.3 &     32.0 & \BF 18.0 &     1.1 &     16.5 &     78.8 & \bf 68.3 &     48.6 & \BF 93.7 &     26.7 &     35.4 &     6.1 & \xx 14.8 & \xx 13.6 & \xx 51.2 & \xx 51.1 \\
        \cellcolor{Apricot!50} \multirow{-2}{*}{\shortstack{ViT \\ L/16}}
        & MERU & \BF 38.8 & \BF 80.6 &     68.7 &     35.5 & \BF 37.2 & \BF 33.0 &     16.6 & \BF 2.2 & \BF 17.2 & \BF 80.0 &     67.5 & \BF 52.1 & \BF 93.7 & \bf 28.1 & \BF 36.5 & \BF 6.2 & \xx 11.8 & \xx 13.1 & \xx 52.7 & \xx 49.3 \\
        \bottomrule
    \end{tabularx}

\end{table*}

%% file: sections/qualitative.tex
\section{Qualitative analysis}
\label{sec:meru_qualitative}

\begin{figure}[t]
    \centering
    \footnotesize
    \newcolumntype{Z}{>{\centering\arraybackslash}X}
    \setlength{\tabcolsep}{1pt}

    \begin{tabularx}{\linewidth}{cZZ}
        ~~~~ & MERU (ViT-L/16) & CLIP (ViT-L/16) \\
    \end{tabularx}
    \includegraphics[width=\linewidth]{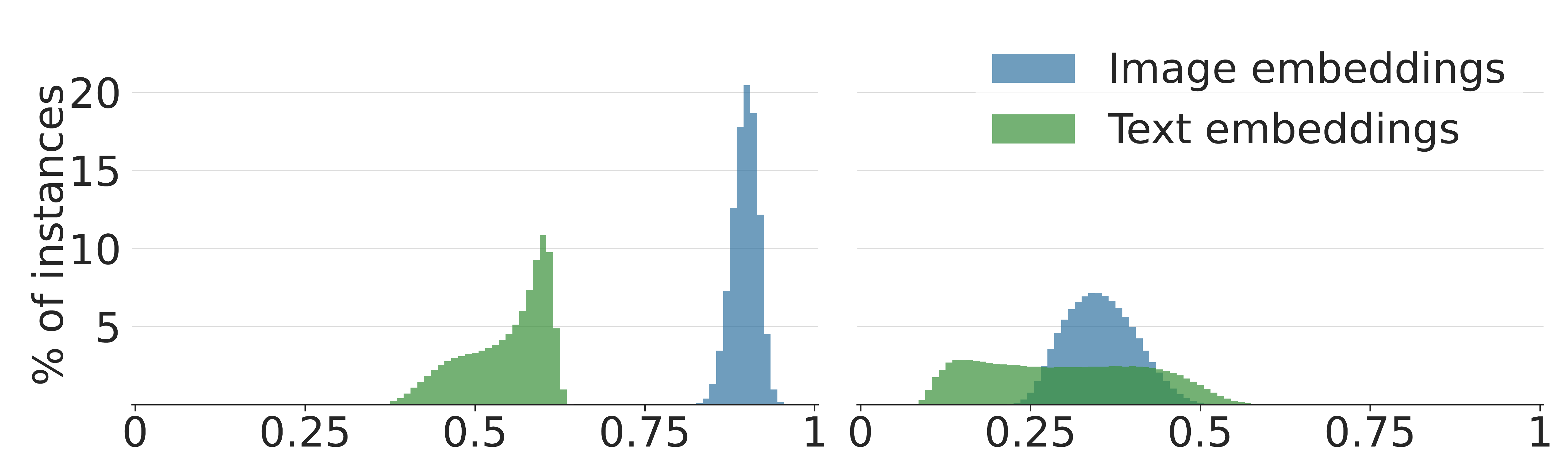} \\
    \begin{tabularx}{\linewidth}{cZZ}
        \rowcolor{Apricot!50}
        ~~~~ &
        $d(\vz) = \lVert \vz_{space} \rVert$ &
        $d(\vz) = 0.5 (1 - \langle \vz, \texttt{[ROOT]} \rangle)$ \\
    \end{tabularx}
    \vspace{-10pt}
    \caption{
        \textbf{Distribution of embedding distances from \texttt{[ROOT]}:}
        We embed all 12M training images and text using trained MERU and CLIP.
        Note that precise distance is not necessary for this analysis,
        so we compute simple monotonic transformations of distances, $d(\vz)$.
        MERU embeds text closer to \texttt{[ROOT]} than images.
    }
    \label{fig:meru_embed_dists}
\end{figure}

In this section, we probe our trained models to infer the visual-semantic hierarchy captured by MERU and CLIP.
Apriori we hypothesize that MERU is better equipped to capture this hierarchy due to the geometric properties of hyperbolic spaces
and an entailment loss that enforces the partial-order relationship \emph{`text entails image'}.
All our analysis in this section uses ViT-L/16 models.

\paragraph{Preliminary: \texttt{[ROOT]} embedding.}
Recall \Cref{fig:meru_intro} -- if we think of the visual-semantic hierarchy as a tree,
then its \emph{leaf nodes} are images and the \emph{intermediate nodes} are text descriptions with varying \emph{semantic specificity}.
Naturally, the \emph{root node} should represent the \emph{most generic concept}.
We denote its embedding in the representation space as \texttt{[ROOT]}.

For MERU, \texttt{[ROOT]} is the origin of the Lorentz hyperboloid as it entails the entire representation space.
The location of \texttt{[ROOT]} for CLIP is not as intuitive --
the notion of entailment is mathematically not defined, and the origin does not lie on the hypersphere.
We empirically estimate CLIP's \texttt{[ROOT]} as an embedding vector that has the least distance from all embeddings of the training dataset.
Hence, we average all 2$\times$12M embeddings of images and text in RedCaps, followed by $L^2$ normalization.
\texttt{[ROOT]} will be different for different CLIP models, whereas it is fixed for MERU.

\input{figtabs/main_traversals.tex}

\paragraph{Embedding distances from \texttt{[ROOT]}.}
In a representation space that effectively captures the visual-semantic hierarchy,
text embeddings should lie closer to \texttt{[ROOT]} than image embeddings,
since text is more \emph{generic} than images (\Cref{fig:meru_intro}).
\Cref{fig:meru_embed_dists} shows the distribution of embedding distances from \texttt{[ROOT]} --
these distributions overlap for CLIP but are separated for MERU.
The range of distributions in \Cref{fig:meru_embed_dists} (left) hints that MERU embeds text and images in two \emph{concentric, high-dimensional rings} around \texttt{[ROOT]}.
The \emph{ring} of text is more \emph{spread out}, whereas the ring of images is relatively \emph{thin}.
This resembles the structure of the visual-semantic hierarchy -- images only occupy \emph{leaf nodes} whereas text occupies many intermediate nodes.

\paragraph{Image traversals.}
In a discrete tree, one can discover the \emph{ancestors} of any node by performing shortest-path traversal to the \emph{root node} \cite{dijkstra1959algorithm}.
We perform such traversals for images with MERU and CLIP.
If the representation space has captured the visual-semantic hierarchy,
then a shortest-path traversal from an image to \texttt{[ROOT]} should let us infer textual concepts that describe the image with varying levels of abstraction.
We briefly describe this analysis here,
refer \Cref{sec:meru_appendix_traversals} for more details.

We traverse from an image and \texttt{[ROOT]} by interpolating 50 equally spaced steps along the geodesic connecting their embedding vectors.
We use every interpolated step embedding as a query to perform retrieve the nearest neighbor from a set of text embeddings $\mathcal{X}$, that also include \texttt{[ROOT]}.

We display results with 60 randomly selected images collected from \href{https://www.pexels.com}{\texttt{pexels.com}}, a website that offers freely usable stock photos.
We use two different sets $\mathcal{X}$ having text sourced from:
(1) 750 captions obtained using the image metadata from \href{https://www.pexels.com}{\texttt{pexels.com}},
and
(2) 8.7M captions from the YFCC dataset~\citep{yfcc100m}.

\Cref{fig:meru_main_image_traversals} shows results with 8 selected images and captions from \href{https://www.pexels.com}{\texttt{pexels.com}}.
\Cref{sec:meru_appendix_traversals} includes results with 52 other images and with YFCC captions \emph{without cherry-picking}.
CLIP seems to capture hierarchy to some extent, often retrieving very few (or zero) captions between image and \texttt{[ROOT]}.
MERU captures it with much finer granularity,
retrieving concepts that gradually become more \emph{generic} as we move closer to \texttt{[ROOT]}.

%% file: figtabs/main_traversals.tex
\begin{figure*}[t!]
    \footnotesize
    \newcolumntype{Z}{>{\centering\arraybackslash}X}
    \setlength{\tabcolsep}{1pt}

    \hfill
    \frame{\includegraphics[width=0.18\linewidth]{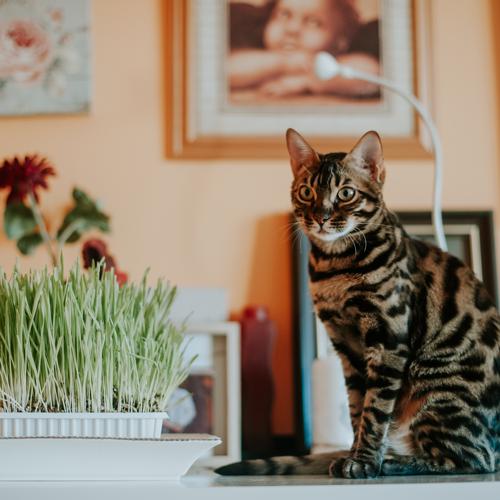}} \hfill \hfill
    \frame{\includegraphics[width=0.18\linewidth]{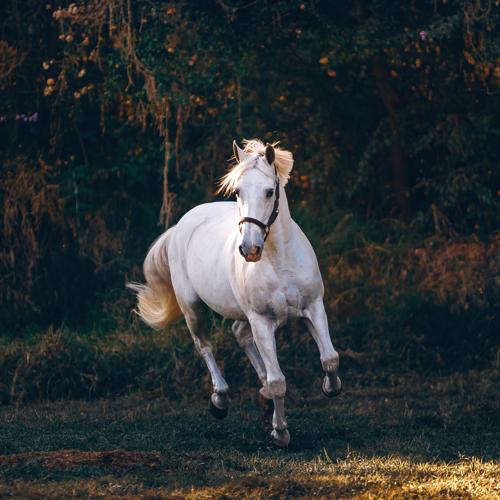}} \hfill \hfill
    \frame{\includegraphics[width=0.18\linewidth]{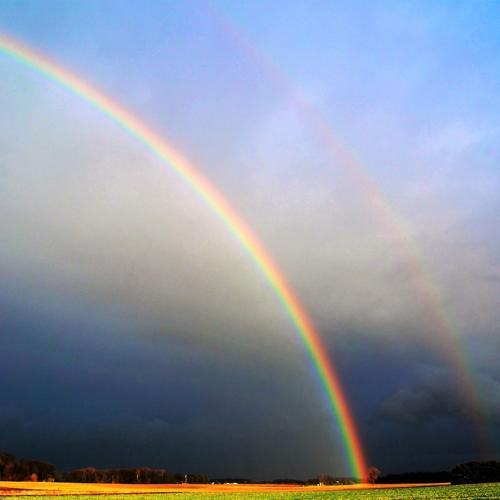}} \hfill \hfill
    \frame{\includegraphics[width=0.18\linewidth]{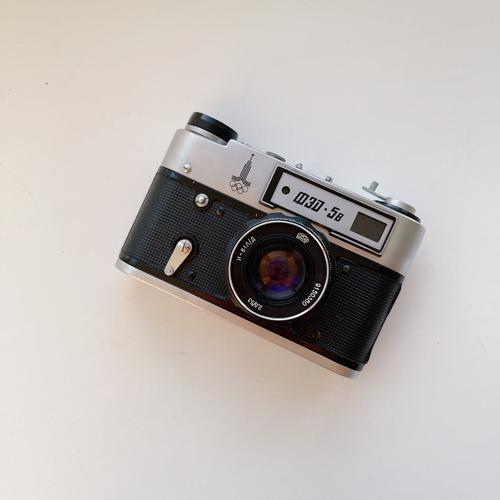}} \hfill \hfill

    \begin{tabularx}{0.24\linewidth}[t]{ZZ}
        \rowcolor{Gray!40} \textbf{MERU} & \textbf{CLIP} \\
        \midrule
        \rowcolor{Apricot!100} \emph{a bengal cat sitting beside wheatgrass on a white surface} & \emph{a bengal cat sitting beside wheatgrass on a white surface} \\
        \midrule
        \rowcolor{Apricot!73} \emph{bengal} & $\downarrow$ \\
        \midrule
        \rowcolor{Apricot!46} \emph{cat} & $\downarrow$ \\
        \midrule
        \rowcolor{Apricot!20} \emph{domestic} & $\downarrow$ \\
        \midrule 
        \texttt{\textbf{[ROOT]}} & \texttt{\textbf{[ROOT]}} \\
    \end{tabularx}
    \hfill
    \begin{tabularx}{0.24\linewidth}[t]{ZZ}
        \rowcolor{Gray!40} \textbf{MERU} & \textbf{CLIP} \\
        \midrule
        \rowcolor{Apricot!100} \emph{white horse} & \emph{white horse} \\
        \midrule
        \rowcolor{Apricot!84} \emph{equine} & $\downarrow$ \\
        \midrule
        \rowcolor{Apricot!68} \emph{equestrian} & $\downarrow$ \\
        \midrule
        \rowcolor{Apricot!52} \emph{beauty} & $\downarrow$ \\
        \midrule
        \rowcolor{Apricot!36} \emph{female} & $\downarrow$ \\
        \midrule
        \rowcolor{Apricot!20} \emph{fluffy} & $\downarrow$ \\
        \midrule 
        \texttt{\textbf{[ROOT]}} & \texttt{\textbf{[ROOT]}} \\
    \end{tabularx}
    \hfill
    \begin{tabularx}{0.24\linewidth}[t]{ZZ}
        \rowcolor{Gray!40} \textbf{MERU} & \textbf{CLIP} \\
        \midrule
        \rowcolor{Apricot!100} \emph{photography of rainbow during cloudy sky} & \emph{phenomenon} \\
        \midrule
        \rowcolor{Apricot!73} \emph{rainbow} & $\downarrow$ \\
        \midrule
        \rowcolor{Apricot!46} \emph{phenomenon} & $\downarrow$ \\
        \midrule
        \rowcolor{Apricot!20} \emph{rural} & $\downarrow$ \\
        \midrule 
        \texttt{\textbf{[ROOT]}} & \texttt{\textbf{[ROOT]}} \\
    \end{tabularx}
    \hfill
    \begin{tabularx}{0.24\linewidth}[t]{ZZ}
        \rowcolor{Gray!40} \textbf{MERU} & \textbf{CLIP} \\
        \midrule
        \rowcolor{Apricot!100} \emph{retro photo camera on table} & $\downarrow$ \\
        \midrule
        \rowcolor{Apricot!73} \emph{fujinomiya} & $\downarrow$ \\
        \midrule
        \rowcolor{Apricot!46} \emph{vintage} & $\downarrow$ \\
        \midrule
        \rowcolor{Apricot!20} \emph{style} & $\downarrow$ \\
        \midrule 
        \texttt{\textbf{[ROOT]}} & \texttt{\textbf{[ROOT]}} \\
    \end{tabularx}

    \vspace{5pt}
    \hfill
    \frame{\includegraphics[width=0.18\linewidth]{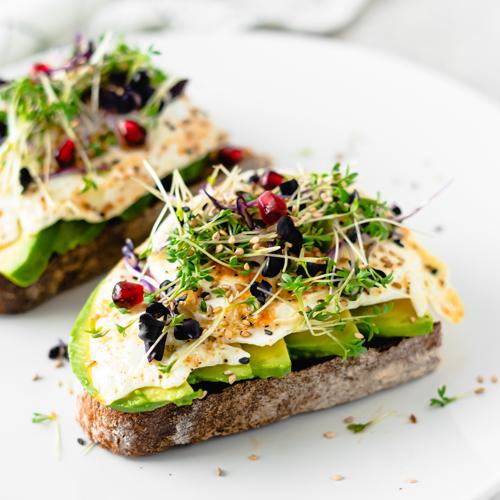}} \hfill \hfill
    \frame{\includegraphics[width=0.18\linewidth]{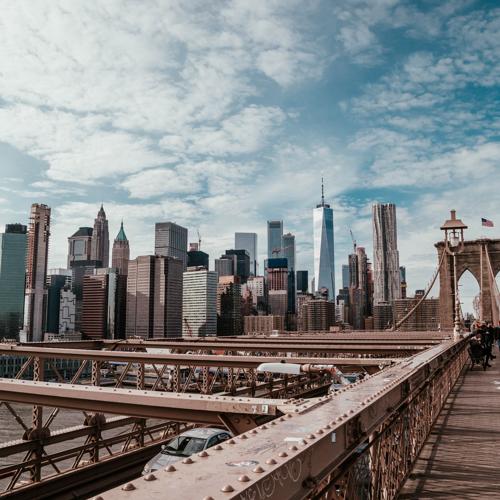}} \hfill \hfill
    \frame{\includegraphics[width=0.18\linewidth]{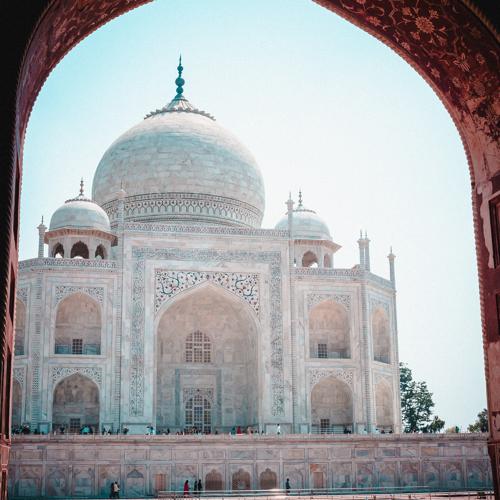}} \hfill \hfill
    \frame{\includegraphics[width=0.18\linewidth]{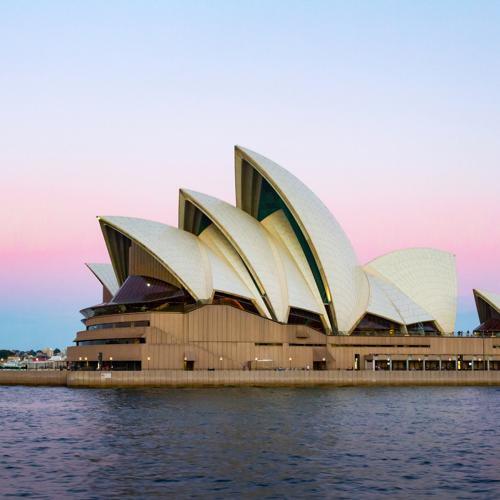}} \hfill \hfill

    \begin{tabularx}{0.24\linewidth}[t]{ZZ}
        \rowcolor{Gray!40} \textbf{MERU} & \textbf{CLIP} \\
        \midrule
        \rowcolor{Apricot!100} \emph{avocado toast} & \emph{avocado toast} \\
        \midrule
        \rowcolor{Apricot!80} \emph{healthy breakfast} & \emph{delicious} \\
        \midrule
        \rowcolor{Apricot!60} \emph{delicious} & $\downarrow$ \\
        \midrule
        \rowcolor{Apricot!40} \emph{homemade} & $\downarrow$ \\
        \midrule
        \rowcolor{Apricot!20} \emph{fresh} & $\downarrow$ \\
        \midrule 
        \texttt{\textbf{[ROOT]}} & \texttt{\textbf{[ROOT]}} \\
    \end{tabularx}
    \hfill
    \begin{tabularx}{0.24\linewidth}[t]{ZZ}
        \rowcolor{Gray!40} \textbf{MERU} & \textbf{CLIP} \\
        \midrule
        \rowcolor{Apricot!100} \emph{brooklyn bridge} & \emph{photo of brooklyn bridge, new york} \\
        \midrule
        \rowcolor{Apricot!80} \emph{new york city} & \emph{new york city} \\
        \midrule
        \rowcolor{Apricot!60} \emph{city} & \emph{new york} \\
        \midrule
        \rowcolor{Apricot!40} \emph{outdoors} & $\downarrow$ \\
        \midrule
        \rowcolor{Apricot!20} \emph{day} & $\downarrow$ \\
        \midrule 
        \texttt{\textbf{[ROOT]}} & \texttt{\textbf{[ROOT]}} \\
    \end{tabularx}
    \hfill
    \begin{tabularx}{0.24\linewidth}[t]{ZZ}
        \rowcolor{Gray!40} \textbf{MERU} & \textbf{CLIP} \\
        \midrule
        \rowcolor{Apricot!100} \emph{taj mahal} & \emph{taj mahal through an arch } \\
        \midrule
        \rowcolor{Apricot!80} \emph{monument} & \emph{travel} \\
        \midrule
        \rowcolor{Apricot!60} \emph{architecture} & \emph{inspiration} \\
        \midrule
        \rowcolor{Apricot!40} \emph{travel} & $\downarrow$ \\
        \midrule
        \rowcolor{Apricot!20} \emph{day} & $\downarrow$ \\
        \midrule 
        \texttt{\textbf{[ROOT]}} & \texttt{\textbf{[ROOT]}} \\
    \end{tabularx}
    \hfill
    \begin{tabularx}{0.24\linewidth}[t]{ZZ}
        \rowcolor{Gray!40} \textbf{MERU} & \textbf{CLIP} \\
        \midrule
        \rowcolor{Apricot!100} \emph{sydney opera house } & \emph{sydney opera house } \\
        \midrule
        \rowcolor{Apricot!73} \emph{opera house} & \emph{opera house} \\
        \midrule
        \rowcolor{Apricot!46} \emph{holiday} & \emph{gift} \\
        \midrule
        \rowcolor{Apricot!20} \emph{day} & \emph{beauty} \\
        \midrule 
        \texttt{\textbf{[ROOT]}} & \texttt{\textbf{[ROOT]}} \\
    \end{tabularx}
    \caption{
        \textbf{Image traversals with MERU and CLIP.}
        We perform text retrieval at multiple steps while traversing from an image embedding to \texttt{[ROOT]}.
        Overall, CLIP retrieves fewer textual concepts (top row), but in some cases it reveals a coarse hierarchy (bottom row).
        MERU captures hierarchy with significantly greater detail, we observe that:
        (1) Text becomes more \emph{generic} we move towards \texttt{[ROOT]}, \eg{} \textcolor{RedOrange}{\emph{white horse $\rightarrow$ equestrian}} and \textcolor{RedOrange}{\emph{retro photo camera $\rightarrow$ vintage}}.
        (2) MERU has higher recall of concepts than CLIP, like words in bottom row: \textcolor{RedOrange}{\emph{homemade, city, monument}}.
        (3) MERU also shows systematic text$\rightarrow$image entailment, \eg{} \textcolor{RedOrange}{\emph{day}} entails many images captured in daylight.
    }
    \vspace{-5pt}
    \label{fig:meru_main_image_traversals}
\end{figure*}

%% file: sections/related.tex
\section{Related work}
\label{sec:meru_related}

\paragraph{Visual-language representation learning.}
Soon after the initial success of deep learning on ImageNet~\citep{krizhevsky2012imagenet},
deep metric learning~\citep{Sohn2016-vp,Song2015-uq} was used to learn vision-language representations in a shared semantic space~\citep{Frome2013-rl,karpathy2015splits}.
The motivations at the time included the possibility of improving vision models~\citep{Frome2013-rl},
enabling zero-shot learning by expressing novel categories as sentences~\citep{Frome2013-rl,write_a_classifier}, and
better image-text retrieval~\citep{karpathy2015splits,young2014flickr}.
Another line of work proposed learning visual models from language supervision via objectives like textual n-gram prediction~\citep{li2017ngrams},
or \emph{generative} objectives like masked language modeling~\citep{bulent2020icmlm} or image captioning~\citep{desai2021virtex}.

More recent approaches like CLIP~\citep{radford2021clip} and ALIGN~\citep{jia2021align} use contrastive metric learning
to pre-train Vision Transformers~\citep{dosovitskiy2021vit} and have helped to better realize the motivations of the earlier works in practice.
While all prior works learn Euclidean embeddings,
MERU explicitly works in the hyperbolic space that is conceptually better for embedding the visual-semantic hierarchy (\Cref{fig:meru_intro}) underlying images and text.
Our results (\Cref{sec:meru_experiments}) demonstrate that MERU yields strong performance as prior works,
and also offers better interpretability to the representation space.

\paragraph{Entailment embeddings.}
In a vision and language context, Order Embeddings~\citep{vendrov2015order} propose capturing the partial order between language and vision
by enforcing that text embeddings $\vx$ and image embeddings $\vy$,
should satisfy $\vy \le \vx$ for all dimensions $i$.
While enforcing order is useful for retrieval,
in our initial experiments, we found that distance-based contrastive
learning to be crucial for better performance on classification and retrieval.
Thus, we focus on adapting the currently successful contrastive learning
and add our entailment objective in conjunction,
to obtain the desired structure in the representation space.

For NLP and knowledge graph embedding applications,
several approaches embed partially ordered data
\citep{ganea2018entailment,Nguyen2017HierarchicalEF,Bai2021ModelingHH,Dasgupta2020ImprovingLI,Vilnis2018ProbabilisticEO}
or discover ordering from pairwise similarities~\citep{tifrea2018poincare,Nickel2017PoincareF,le2019inferringch}.
Our work has a flavor of both these lines of work,
since we impose structure \emph{across} modalities,
but order also emerges \emph{within} modality (\Cref{fig:meru_main_image_traversals}).

\paragraph{Hyperbolic representations in computer vision.}
\citet{Khrulkov2020HyperbolicIE} learn hyperbolic image embeddings using image-label pairs,
while \citet{atigh2022hyperbolicsegm} study image segmentation by utilizing hyperbolic geometry.
More recently, \citet{ermolov2022hyperbolic} and \citet{ge2022hyperbolic} extend standard contrastive self-supervised learning framework~\citep{wu2018npid,he2020moco} in vision to learn hyperbolic representations.
In contrast to all these works, MERU learns multimodal representations with an order of magnitude more data
and shows strong \emph{zero-shot} transfer abilities across generic artificial intelligence tasks~\citep{radford2021clip}.

%% file: sections/conclusion.tex
\section{Conclusion}
\label{sec:conclusion}

In this paper, we learn large-scale image-text representations (MERU) to capture the visual-semantic hierarchy underlying images and text.
Our key innovation is to bring advances in learning hyperbolic representations to practical, large-scale deep learning applications.
MERU is competitive or more performant than approaches that learn Euclidean representations (like CLIP).
It does so along with capturing hierarchical knowledge which allows one to make powerful inferences
such as reasoning about images at different levels of abstraction.
Beyond this, our model also provides clear performance gains for small embedding dimensions
(which are useful in resource-constrained settings).
We hope this work catalyzes progress in learning useful representations from large amounts of unstructured data.

\paragraph{Future work.}
In this \emph{scaling} era, we are seeing rapid progress with large multi-modal models trained using millions (or even billions) of image-text pairs.
The quality and concept distribution of training data plays a vital role in the efficacy of these models.
Such training data is becoming increasingly opaque and black-box due to its unprecedented scale.
We believe that the time is ripe to revisit the unreasonable effectiveness of data in deep learning \citep{halevy2009unreasonable,sun2017revisiting}.
Modeling hierarchies can help uncover higher-order relationships beyond basic data statistics.
As a concrete example, \Cref{fig:meru_intro} ``so cute $<$3''
is an extremely generic caption and does not the \emph{precise} details in images.
Such captions add noisy supervision in contrastive loss by making false negative pairs with many images in the batch.
Image traversals with MERU \Cref{fig:meru_main_image_traversals} can discover such noisy captions.
ML practitioners can filter or re-caption such training images to improve dataset quality and train subsequent models for improved performance.

\paragraph{Limitations.}
Our work is not without limitations.
While MERU yields hyperbolic representations that excel at zero-shot retrieval and image classification tasks,
the linear probe evaluations in \Cref{tab:meru_appendix_linprobe} show that the underlying Euclidean representations from the image encoder of MERU underperform CLIP.
Exploring MERU's transferability to other tasks that involve
few-shot learning or full-model fine-tuning is also beyond scope of this paper.
Finally, while we provide ample qualitative analysis of image traversals,
future work can propose more systematic ways to evaluate the hierarchical knowledge captured by vision-language models.

%% file: sections/appendix_approach.tex
\section{Entailment loss derivations}
\label{sec:meru_appendix_approach}

We derive the entailment loss components (\cref{eqn:meru_entailment}) used in our approach.
Note that for $c > 0$, the curvature of the hyperboloid is $-c$.

\paragraph{Half-aperture.}
To derive the entailment loss for arbitrary curvatures $c > 0$,
we start with the expression of half-aperture for the Poincar\'e ball, introduced by \citet{ganea2018entailment}.
Let $\vx_b$ be a point on the Poincar\'e ball, the cone half-aperture is defined as follows:
\begin{equation}\label{eqn:meru_appendix_aperture_poincare}
\text{aper}_b (\vx_b) = \sin^{-1} \left(K \frac{1 - c \; \lVert \vx_b \rVert^2}{\sqrt{c} \; \lVert \vx_b \rVert}\right)
\end{equation}
The Poincar\'e ball model and Lorentz hyperboloid model are isometric to each other
-- one can map any point $\vx_b$ from the Poincar\'e ball
to another point $\vx_h$ on the hyperboloid using the following differentiable transformation:
\begin{equation}\label{eqn:meru_appendix_poincare_to_lorentz}
    \vx_h = \frac{2 \vx_b}{1 - c \; \lVert \vx_b \rVert^2}
\end{equation}
The half-aperture of a cone should be invariant to the exact hyperbolic model we use, hence $\text{aper}_h (\vx_h) = \text{aper}_b (\vx_b)$.
Substituting \cref{eqn:meru_appendix_poincare_to_lorentz} in \cref{eqn:meru_appendix_aperture_poincare}, we get the expression:
\begin{equation*}
    \text{aper}_{h}(\vx_h) = \sin^{-1} \left(\frac{2K}{\sqrt{c} \; \lVert \vx_h \rVert}\right)
\end{equation*}


\paragraph{Exterior angle.}
Consider three points $\vO$ (the origin), $\vx$ (text embedding) and $\vy$ (image embedding).
Then, a hyperbolic triangle is a closed shape formed by the geodesics connecting each pair of points.
Similar to the Euclidean plane, the hyperbolic plane also has its law of cosines
that allows us to talk about the angles in the triangle~\citep{lee2019textbook}.
Let the Lorentzian distances (\cref{eqn:meru_distance}) be
$x = d(\vO, \vy)$, $y = d(\vO, \vx)$, and $z = d(\vx, \vy)$.
We can write the expression of \emph{exterior angle} as follows:

\begin{align*}
    &\text{ext} (\vx, \vy) = \pi - \angle \vO \vx \vy \\
    &  = \pi - \cos^{-1} \left[ \frac{\cosh(z \sqrt{c}) \cosh(y \sqrt{c}) - \cosh(x \sqrt{c})}{\sinh(z \sqrt{c}) \sinh(y \sqrt{c})}\right]
\end{align*}

We use the relation $\pi - \cos^{-1}(t) = \cos^{-1}(-t)$ in the above equation.
Then, let us define a function $g(t) = \cosh(t \sqrt{c})$ for brevity, and substitute in the above equation.
We also substitute $\sinh(t) = \sqrt{\cosh^2(t) - 1}$ as per the hyperbolic trigonometric identity.
Putting it all together, we get:

\begin{equation}\label{eqn:meru_appendix_ext_with_g}
    \text{ext} (\vx, \vy) = \cos^{-1} \left[ \frac{g(x) - g(z) g(y)}{\sqrt{g(z)^2 -1} \sqrt{g(y)^2 -1}}\right]
\end{equation}

Now all we need is to compute $g(x)$, $g(y)$, and $g(z)$.
We substitute the $z = d(\vx, \vy)$ in $g(z)$ below:
\begin{align*}
    g(z) &= \cosh \left( d(\vx, \vy) \sqrt{c} \right) \\
         &= \cosh \left( \frac{1}{\sqrt{c}} \cosh^{-1} (-c \; \langle \vx, \vy \rangle_{\mathcal{L}}) \cdot \sqrt{c} \right) \\
         &= -c \; \langle \vx, \vy \rangle_{\mathcal{L}}
\end{align*}

Similarly, $g(x) = -c \langle \vO, \vy \rangle_{\mathcal{L}}$ and $g(y) = -c \langle \vO, \vx \rangle_{\mathcal{L}}$.
The Lorentzian inner product (\cref{eqn:meru_lorentz_inner}) with origin $\vO$ simplifies:
\begin{equation*}
    \langle \vO, \vx \rangle_{\mathcal{L}} = - \frac{x_{time}}{\sqrt{c}}
    \quad
    \text{and}
    \quad
    \langle \vO, \vy \rangle_{\mathcal{L}} = - \frac{y_{time}}{\sqrt{c}}
\end{equation*}

Through this, we get $g(x) = x_{time} \sqrt{c}$ and $g(y) = y_{time} \sqrt{c}$.
Finally, we can substitute $g(x)$, $g(y)$, and $g(z)$ to re-write \cref{eqn:meru_appendix_ext_with_g} to give the final expression as follows:
\begin{equation}
    \def\cvl{c \; \langle \vx,\vy \rangle _\mathcal{L}}
    \text{ext}(\vx, \vy) = \cos^{-1} \left( \frac{y_{time} + x_{time} \; \cvl{}}{\sqrt{x_{time}^2 c - 1} \sqrt{\left( \cvl \right)^2 - 1}} \right) \nonumber
\end{equation}

Finally, we use the relation between $x_{time}$ and $\vx_{space}$ (\cref{eqn:meru_spacetime}) to simplify the denominator,
giving the final expression of exterior angle as follows:
\begin{equation}
    \def\cvl{c \; \langle \vx,\vy \rangle _\mathcal{L}}
    \text{ext}(\vx, \vy) = \cos^{-1} \left( \frac{y_{time} + x_{time} \; \cvl{}}{\lVert \vx_{space} \rVert \sqrt{\left( \cvl \right)^2 - 1}} \right) \nonumber
\end{equation}

%% file: sections/appendix_clip_baseline.tex
\begin{table*}[t]
  \setlength{\tabcolsep}{2.5pt}
  \footnotesize
  \caption{
      \textbf{CLIP baseline.}
      We develop a strong CLIP baseline that trains on an 8-GPU machine in less than one day (ViT-S image encoder),
      starting with SLIP~\citep{mu2021slip} as a reference.
      We benchmark improvements on zero-shot image classification across 16 datasets.
      Our RedCaps-trained CLIP baseline (last row) is a significantly stronger baseline than its YFCC-trained counterparts.
  }
  \label{tab:meru_appendix_baseline}
  \begin{tabularx}{\linewidth}{X c ccccccccccccccccc}
      \toprule
          &
          \rotatebox[origin=lb]{90}{\shortstack{Images\\Seen}} &
          \rotatebox[origin=lb]{90}{\smash{ImageNet}} &
          \rotatebox[origin=lb]{90}{\smash{Food-101}} &
          \rotatebox[origin=lb]{90}{\smash{CIFAR-10}} &
          \rotatebox[origin=lb]{90}{\smash{CIFAR-100}} &
          \rotatebox[origin=lb]{90}{\smash{CUB}} &
          \rotatebox[origin=lb]{90}{\smash{SUN397}} &
          \rotatebox[origin=lb]{90}{\smash{Cars}} &
          \rotatebox[origin=lb]{90}{\smash{Aircraft}} &
          \rotatebox[origin=lb]{90}{\smash{DTD}} &
          \rotatebox[origin=lb]{90}{\smash{Pets}} &
          \rotatebox[origin=lb]{90}{\smash{Caltech-101}} &
          \rotatebox[origin=lb]{90}{\smash{Flowers}} &
          \rotatebox[origin=lb]{90}{\smash{STL-10}} &
          \rotatebox[origin=lb]{90}{\smash{EuroSAT}} &
          \rotatebox[origin=lb]{90}{\smash{RESISC45}} &
          \rotatebox[origin=lb]{90}{\smash{Country211}} &
          \rotatebox[origin=lb]{90}{\smash{Average}} \\
      \midrule
      \rowcolor{Apricot!50} \multicolumn{19}{l}{\textbf{YFCC15M-trained models}} \\
      SLIP's CLIP~\citep{mu2021slip} & 368M &
      32.0 & 43.7 & 61.9 & 30.2 & 30.9 & 41.3 & 3.5 & 3.9 & 18.1 & 26.1 & 51.4 & 48.7 & 87.3 & 17.5 & 16.8 & 8.7 & 32.6 \\
      Our implementation & 368M &
      33.1 & 42.3 & 64.9 & 34.4 & 33.7 & 43.8 & 2.9 & 5.1 & 19.1 & 25.0 & 49.8 & 47.2 & 87.4 & 26.8 & 21.6 & 9.0 & 34.1 \\
      \quad + BS 4096$\rightarrow$2048 & 184M &
      28.2 & 34.2 & 58.7 & 29.4 & 27.4 & 39.4 & 2.9 & 4.3 & 16.5 & 20.1 & 43.8 & 42.2 & 85.4 & 20.2 & 19.0 & 8.5 & 30.0 \\
      \quad + \emph{sin-cos pos embed} & 184M &
      28.7 & 34.2 & 67.3 & 33.6 & 25.4 & 41.1 & 3.1 & 4.2 & 17.8 & 21.0 & 44.3 & 43.6 & 86.4 & 18.6 & 19.6 & 8.3 & 31.1 \\
      \midrule
      \rowcolor{Apricot!50} \multicolumn{19}{l}{\textbf{RedCaps-trained models}} \\
      \quad + YFCC$\rightarrow$RedCaps & 184M &
      32.6 & 71.5 & 61.4 & 25.6 & 29.9 & 27.5 & 10.1 & 1.5 & 14.3 & 72.7 & 62.8 & 42.2 & 88.0 & 18.1 & 30.5 & 4.9 & 37.1 \\
      \quad + 90K$\rightarrow$120K iters. & 246M &
      33.9 & 72.5 & 60.1 & 24.4 & 30.0 & 27.5 & 11.3 & 1.4 & 13.1 & 73.7 & 63.9 & 44.4 & 88.2 & 18.6 & 31.4 & 5.2 & 37.5 \\
      \rowcolor{ForestGreen!30} \quad + our zero-shot prompts & 246M &
      34.3 & 74.5 & 60.1 & 24.4 & 33.8 & 27.5 & 11.3 & 1.4 & 15.0 & 73.7 & 63.9 & 47.0 & 88.2 & 18.6 & 31.4 & 5.2 & 38.1 \\
      \bottomrule
  \end{tabularx}
\end{table*}


\section{Developing a strong CLIP baseline}
\label{sec:meru_appendix_clip_baseline}

One of our contributions is to establish a lightweight, yet strong CLIP baseline.
The original CLIP models \citep{radford2021clip} are trained using a private dataset of 400M image-text pairs across 128 GPUs for more than 10 days.
We aim to maximize accessibility for future works, hence we decide our hyperparameters such that our smallest model can train on a single 8-GPU machine in less than one day.

We start with a reference CLIP ViT-S/16 baseline from SLIP~\citep{mu2021slip} and carefully introduce one modification at a time.
We benchmark improvements on zero-shot image classification across 16 datasets used in our main experiments,
using text prompts used by \citep{radford2021clip}.
Results are shown in \Cref{tab:meru_appendix_baseline}.

\paragraph{CLIP baseline by SLIP.}
This re-implemented baseline was trained using a 15M subset of the YFCC dataset~\citep{yfcc100m}.
We re-evaluate the publicly released ViT-S/16 checkpoint~\footnote{\url{github.com/facebookresearch/slip}} using our evaluation code;
it obtains $32.6\%$ average accuracy across all datasets.

\paragraph{Our re-implementation.}
We attempt a faithful replication of CLIP by following hyperparameters in SLIP.
Our implementation obtains slightly higher average performance ($34.1\%$) with three minor changes:
\begin{compactitem}[\hspace{1pt}--]
  \item We use an \emph{undetached} gather operation to collect all image/text features across all GPUs for contrastive loss.
  This ensures proper gradient flow across devices.
  \item The above change allows using weight decay = $=0.2$ like OpenAI's CLIP, unlike $0.5$ used by SLIP's CLIP.
  \item During training and inference, we resize input images using \emph{bicubic} interpolation like original CLIP, instead of bilinear interpolation in SLIP's CLIP.
\end{compactitem}

\paragraph{Fitting the model on 8-GPUs.}
This CLIP model requires 16$\times$ V100 32GB GPUs with a batch size of 4096 and automatic mixed precision~\citep{micikevicius2018mixed}.
Techniques like gradient checkpointing~\citep{chen2016training} can reduce memory requirements, but it comes at a cost of reduced training speed.
Hence we avoid making it a requirement and simply reduce the batch size to 2048.
This incurs a performance drop as the effective images seen by the model are halved.
We offset the effective shortening of the training schedule by using fixed \emph{sine-cosine} position embeddings in ViT,
so learning position-related inductive biases is not required.
This change slightly improves average accuracy ($30.0\% \rightarrow 31.1\%$ average accuracy).

\paragraph{Training with RedCaps dataset.}
RedCaps dataset~\citep{desai2021redcaps} comprises 12M image-text pairs from Reddit, sourced from Pushshift~\citep{baumgartner2020pushshift}.
Training with RedCaps significantly improves performance over YFCC-trained models ($31.1\% \rightarrow 37.1\%$ average accuracy),
especially on datasets whose concepts have high coverage in RedCaps, \eg{} Food-101~\citep{bossard2014food} and Pets~\citep{parkhi2012pets}.

To account for the smaller size of RedCaps, we increase the training iterations from 90K up to 120K.
Finally, we modify zero-shot prompts for some datasets to match the linguistic style of RedCaps.
For example, many captions in \texttt{r/food} simply mention the name of the dish in the corresponding image,
hence we use the prompt `\texttt{food : \{\}}'.
See \Cref{tab:meru_appendix_prompts} for the list of prompts for all datasets.
We did not extensively tune these prompts, but we checked performance on the held-out validation sets to avoid cheating on the test splits.

Finally, our CLIP ViT-S/16 baseline trains on 8$\times$ V100 32 GB GPUs within $\approx$14 hours and achieves 38.1\% average performance across 16 datasets.
We use these hyperparameters for all MERU and CLIP models in our experiments.

%% file: sections/appendix_evaluations.tex
\input{figtabs/appendix_datasets.tex}
\input{figtabs/appendix_linprobe.tex}

\section{Linear probe evaluation}
\label{sec:meru_appendix_linprobe}

Our experimental evaluations (\Cref{sec:meru_experiments}) focus on \emph{zero-shot} transfer~\citep{write_a_classifier,radford2021clip}.
Another established protocol to evaluate visual representations is \emph{linear probe evaluation}, which involves training linear models on \emph{frozen} image embeddings.
This protocol is popular in self-supervised representation learning literature,
with \citet{doersch2015context}, \citet{zhang2016colorful}, and \citet{noroozi2016jigsaw}
being notable early works.
We follow the implementation of \citet{kornblith2019better} as it is simple and less sensitive to choice of evaluation hyperparameters.
This setup is also followed by CLIP~\citep{radford2021clip} and many recent works on representation learning~\citep{li2021declip,furst2022cloob,elbanani2023languageguided}.

We evaluate using datasets listed in \Cref{tab:meru_appendix_datasets}.
We train a logistic regression classifier on embeddings extracted from the image encoder (before projection layer) of MERU and CLIP.
For MERU, these underlying representations belong to a Euclidean space.
We use the implementation from \texttt{scikit-learn}~\citep{scikit-learn} library, with L-BFGS~\citep{liu1989limited} optimizer and search the regularization cost per dataset, $C \in [10^{-6}, 10^{6}]$, performing two-step search on \texttt{val} split like \citet{radford2021clip}.
Then we train a final classifier on combined \texttt{train} and \texttt{val} splits for a maximum of 1000 iterations, then report top-1 mean per-class accuracy on the \texttt{test} split.

Results in \Cref{tab:meru_appendix_linprobe} show that MERU mostly matches or underperforms CLIP.
Our main focus is not on improving the underlying Euclidean representations from the encoders, but to demonstrate strong \emph{zero-shot} transfer and interpretability benefits.
Future work can focus on improving MERU's capabilities on other transfer applications.

\clearpage
\input{figtabs/appendix_prompts.tex}
\clearpage

%% file: figtabs/appendix_datasets.tex
\begin{table*}[t!]
    \footnotesize
    \setlength{\tabcolsep}{10pt}
    \vspace{-10pt}
    \caption{
        \textbf{Datasets used for image classification evaluation.}
        Datasets in highlighted rows do not have an official validation split --
        we use a random held-out subset of the training split.
        EuroSAT and RESISC do not define any splits; we randomly sample non-overlapping splits.
        CLEVR Counts is derived from CLEVR~\citep{johnson2017clevr}
        and SST2 was introduced as an NLP dataset by \citep{socher2013sst}.
    }
    \label{tab:meru_appendix_datasets}
    \begin{tabularx}{\linewidth}{X cccc}
    \toprule
    \textbf{Dataset} & \textbf{Classes} & \textbf{Train} & \textbf{Val} & \textbf{Test} \\
    \midrule
    \rowcolor{Apricot!30} Food-101~\citep{bossard2014food}         & 101 & 68175 & 7575 & 25250 \\
    \rowcolor{Apricot!30} CIFAR-10~\citep{krizhevsky2009cifar}     & 10 & 45000 & 5000 & 10000 \\
    \rowcolor{Apricot!30} CIFAR-100~\citep{krizhevsky2009cifar}    & 100 & 45000 & 5000 & 10000 \\
    \rowcolor{Apricot!30} CUB-2011~\citep{wah2011cub}              & 200 & 4795 & 1199 & 5794 \\
    \rowcolor{Apricot!30} SUN397~\citep{xiao2010sundb}             & 397 & 15880 & 3970 & 19849 \\
    \rowcolor{Apricot!30} Stanford Cars~\citep{krause2013cars}     & 196 & 6515 & 1629 & 8041 \\
    \rowcolor{Apricot!0}  FGVC Aircraft~\citep{maji2013aircraft}   & 100 & 3334 & 3333 & 3333 \\
    \rowcolor{Apricot!0}  DTD~\citep{cimpoi2014dtd}                & 47 & 1880 & 1880 & 1880 \\
    \rowcolor{Apricot!30} Oxf-IIIT Pets~\citep{parkhi2012pets}     & 37 & 2944 & 736 & 3669 \\
    \rowcolor{Apricot!30} Caltech-101~\citep{feifei2004caltech101} & 102 & 2448 & 612 & 6084 \\
    \rowcolor{Apricot!0}  Flowers~\citep{nilsback2008flowers}      & 102 & 1020 & 1020 & 6149 \\
    \rowcolor{Apricot!30} STL-10~\citep{coates2011stl10}           & 10 & 4000 & 1000 & 8000 \\
    \rowcolor{Apricot!0}  EuroSAT~\citep{helber2019eurosat}        & 10 & 5000 & 5000 & 5000 \\
    \rowcolor{Apricot!0}  RESISC~\citep{cheng2017resisc45}         & 45 & 3150 & 3150 & 25200 \\
    \rowcolor{Apricot!0}  Country211~\citep{radford2021clip}       & 211 & 31650 & 10550 & 21100 \\
    \rowcolor{Apricot!30} MNIST~\citep{lecun2010mnist}             & 10 & 48000 & 12000 & 10000 \\
    \rowcolor{Apricot!30} CLEVR Counts~\citep{zhai2019vtab}        & 8 & 4500 & 500 & 5000 \\
    \rowcolor{Apricot!0}  PCAM~\citep{veeling2018pcam}             & 2 & 262144 & 32768 & 32768 \\
    \rowcolor{Apricot!0}  SST2~\citep{radford2021clip}             & 2 & 6920 & 872 & 1821 \\
    \bottomrule
    \end{tabularx}
\end{table*}

%% file: figtabs/appendix_linprobe.tex
\begin{table*}[t]
    \footnotesize
    \setlength{\tabcolsep}{3pt}
    \vspace{-14pt}
    \caption{
        \textbf{Linear probe evaluation.}
        We train a logistic regression classifier on embeddings extracted from the image encoders of CLIP and MERU (before projection layers).
        Note that embeddings from MERU are \emph{not} lifted onto the hyperboloid.
    }
    \label{tab:meru_appendix_linprobe}
    \begin{tabularx}{\linewidth}{c X c ccccccccccccccccccc}
        \toprule
            &&
            \rotatebox[origin=lb]{90}{\smash{Food-101}} &
            \rotatebox[origin=lb]{90}{\smash{CIFAR-10}} &
            \rotatebox[origin=lb]{90}{\smash{CIFAR-100}} &
            \rotatebox[origin=lb]{90}{\smash{CUB}} &
            \rotatebox[origin=lb]{90}{\smash{SUN397}} &
            \rotatebox[origin=lb]{90}{\smash{Cars}} &
            \rotatebox[origin=lb]{90}{\smash{Aircraft}} &
            \rotatebox[origin=lb]{90}{\smash{DTD}} &
            \rotatebox[origin=lb]{90}{\smash{Pets}} &
            \rotatebox[origin=lb]{90}{\smash{Caltech-101}} &
            \rotatebox[origin=lb]{90}{\smash{Flowers}} &
            \rotatebox[origin=lb]{90}{\smash{STL-10}} &
            \rotatebox[origin=lb]{90}{\smash{EuroSAT}} &
            \rotatebox[origin=lb]{90}{\smash{RESISC45}} &
            \rotatebox[origin=lb]{90}{\smash{Country211}} &
            \rotatebox[origin=lb]{90}{\smash{MNIST}} &
            \rotatebox[origin=lb]{90}{\smash{CLEVR}} &
            \rotatebox[origin=lb]{90}{\smash{PCAM}} &
            \rotatebox[origin=lb]{90}{\smash{SST2}} \\
        \midrule
        \cellcolor{Apricot!50}
        & CLIP & 85.3 & 89.6 & 72.3 & 68.8 & 61.1 & 60.5 & 42.2 & 71.2 & 87.9 & 88.4 & 96.2 & 95.5 & 95.7 & 88.1 & 15.0 & 98.5 & 57.5 & 84.6 & 54.9 \\
        \cellcolor{Apricot!50} \multirow{-2}{*}{\shortstack{ViT \\ S/16}}
        & MERU & 85.2 & 89.7 & 70.9 & 69.2 & 59.6 & 58.0 & 43.1 & 70.2 & 87.5 & 85.6 & 95.5 & 95.5 & 95.8 & 87.0 & 14.8 & 98.2 & 56.8 & 84.1 & 54.5 \\
        \midrule
        \cellcolor{Apricot!50}
        & CLIP & 88.4 & 92.2 & 76.5 & 73.2 & 64.7 & 71.1 & 50.4 & 72.6 & 90.2 & 89.6 & 97.3 & 97.1 & 96.9 & 90.0 & 16.7 & 98.9 & 52.7 & 84.4 & 57.6 \\
        \cellcolor{Apricot!50} \multirow{-2}{*}{\shortstack{ViT \\ B/16}}
        & MERU
        & 88.2 & 92.3 & 74.6 & 70.9 & 63.4 & 68.4 & 48.2 & 70.7 & 90.3 & 88.6 & 96.6 & 96.7 & 96.5 & 89.0 & 16.5 & 98.7 & 56.0 & 85.5 & 56.2 \\
        \midrule
        \cellcolor{Apricot!50}
        & CLIP & 89.6 & 95.3 & 80.5 & 75.7 & 66.0 & 75.7 & 54.5 & 75.7 & 92.0 & 92.0 & 97.4 & 97.6 & 96.9 & 90.5 & 17.8 & 99.2 & 55.6 & 87.5 & 56.1 \\
        \cellcolor{Apricot!50} \multirow{-2}{*}{\shortstack{ViT \\ L/16}}
        & MERU & 89.0 & 94.1 & 77.3 & 74.2 & 63.7 & 71.9 & 51.2 & 70.9 & 90.1 & 87.5 & 96.7 & 97.3 & 96.8 & 89.1 & 17.0 & 98.9 & 55.4 & 86.0 & 55.8 \\
        \bottomrule
    \end{tabularx}
\end{table*}

%% file: figtabs/appendix_prompts.tex
\begin{table*}[t!]
    \footnotesize
    \caption{
        \textbf{Prompts used for zero-shot classification (\Cref{subsec:meru_classification_eval}).}
        Most of these prompts are same as \citep{radford2021clip}.
        We modify prompts for some datasets, that significantly improved performance for both MERU and CLIP --
        We did not perform extensive prompt tuning, we simply checked the performance on \texttt{val} splits for our CLIP baseline (\Cref{sec:meru_appendix_clip_baseline}).
        \textbf{NOTE:} Some prompts use the word \texttt{`porn'} as it is included in the subreddit name.
        It does not indicate pornographic content but simply high-quality photographs.
    }
    \label{tab:meru_appendix_prompts}
    \begin{tabularx}{\linewidth}{XXX}
        \midrule
        \rowcolor{Gray!40} \multicolumn{3}{l}{\textbf{ImageNet (our prompts)}} \\
        \midrule
        \rowcolor{Apricot!0}  \texttt{i took a picture : itap of a \{\}.} &
                              \texttt{pics : a bad photo of the \{\}.} &
                              \texttt{pics : a origami \{\}.} \\
        \rowcolor{Apricot!50} \texttt{pics : a photo of the large \{\}.} &
                              \texttt{pics : a \{\} in a video game.} &
                              \texttt{pics : art of the \{\}.} \\
        \rowcolor{Apricot!0}  \texttt{pics : a photo of the small \{\}.} \\
        \midrule
    \end{tabularx}

    \vspace{5pt}

    \renewcommand{\arraystretch}{1.2}
    \begin{tabularx}{0.325\linewidth}[t]{X}
        \midrule
        \rowcolor{Gray!40} \textbf{Food-101 (our prompts)} \\
        \midrule
        \rowcolor{Apricot!0}  \texttt{food : \{\}.} \\
        \rowcolor{Apricot!50} \texttt{food porn : \{\}.} \\
        \midrule

        \rowcolor{Gray!40} \textbf{CIFAR-10 and CIFAR-100} \\
        \midrule
        \rowcolor{Apricot!0}  \texttt{a photo of a \{\}.} \\
        \rowcolor{Apricot!50} \texttt{a blurry photo of a \{\}.} \\
        \rowcolor{Apricot!0}  \texttt{a black and white photo of a \{\}.} \\
        \rowcolor{Apricot!50} \texttt{a low contrast photo of a \{\}.} \\
        \rowcolor{Apricot!0}  \texttt{a high contrast photo of a \{\}.} \\
        \rowcolor{Apricot!50} \texttt{a bad photo of a \{\}.} \\
        \rowcolor{Apricot!0}  \texttt{a good photo of a \{\}.} \\
        \rowcolor{Apricot!50} \texttt{a photo of a small \{\}.} \\
        \rowcolor{Apricot!0}  \texttt{a photo of a big \{\}.} \\
        \rowcolor{Apricot!50} \texttt{a photo of the \{\}.} \\
        \rowcolor{Apricot!0}  \texttt{a blurry photo of the \{\}.} \\
        \rowcolor{Apricot!50} \texttt{a black and white photo of the \{\}.} \\
        \rowcolor{Apricot!0}  \texttt{a low contrast photo of the \{\}.} \\
        \rowcolor{Apricot!50} \texttt{a high contrast photo of the \{\}.} \\
        \rowcolor{Apricot!0}  \texttt{a bad photo of the \{\}.} \\
        \rowcolor{Apricot!50} \texttt{a good photo of the \{\}.} \\
        \rowcolor{Apricot!0}  \texttt{a photo of the small \{\}.} \\
        \rowcolor{Apricot!50} \texttt{a photo of the big \{\}.} \\
        \midrule

        \rowcolor{Gray!40} \textbf{CUB-2011 (our prompts)} \\
        \midrule
        \rowcolor{Apricot!0}  \texttt{bird pics : \{\}.} \\
        \rowcolor{Apricot!50} \texttt{birding : \{\}.} \\
        \rowcolor{Apricot!0}  \texttt{birds : \{\}.} \\
        \rowcolor{Apricot!50} \texttt{bird photography : \{\}.} \\
        \midrule

        \rowcolor{Gray!40} \textbf{SUN397} \\
        \midrule
        \rowcolor{Apricot!0}  \texttt{a photo of a \{\}.} \\
        \rowcolor{Apricot!50} \texttt{a photo of the \{\}.} \\
        \midrule

        \rowcolor{Gray!40} \textbf{Stanford Cars} \\
        \midrule
        \rowcolor{Apricot!0}  \texttt{a photo of a \{\}.} \\
        \rowcolor{Apricot!50} \texttt{a photo of the \{\}.} \\
        \rowcolor{Apricot!0}  \texttt{a photo of my \{\}.} \\
        \rowcolor{Apricot!50} \texttt{i love my \{\}!} \\
        \rowcolor{Apricot!0}  \texttt{a photo of my dirty \{\}.} \\
        \rowcolor{Apricot!50} \texttt{a photo of my clean \{\}.} \\
        \rowcolor{Apricot!0}  \texttt{a photo of my new \{\}.} \\
        \rowcolor{Apricot!50} \texttt{a photo of my old \{\}.} \\
        \midrule

        \rowcolor{Gray!40} \textbf{FGVC Aircraft} \\
        \midrule
        \rowcolor{Apricot!0}  \texttt{a photo of a \{\}, a type of aircraft.} \\
        \rowcolor{Apricot!50} \texttt{a photo of the \{\}, a type of aircraft.} \\
        \midrule
    \end{tabularx}
    \hfill
    \renewcommand{\arraystretch}{1.198}
    \begin{tabularx}{0.325\linewidth}[t]{X}
        \midrule
        \rowcolor{Gray!40} \textbf{DTD (our prompts)} \\
        \midrule
        \rowcolor{Apricot!0}  \texttt{pics : \{\} texture.} \\
        \rowcolor{Apricot!50} \texttt{pics : \{\} pattern.} \\
        \rowcolor{Apricot!0}  \texttt{pics : \{\} thing.} \\
        \rowcolor{Apricot!50} \texttt{pics : this \{\} texture.} \\
        \rowcolor{Apricot!0}  \texttt{pics : this \{\} pattern.} \\
        \rowcolor{Apricot!50} \texttt{pics : this \{\} thing.} \\
        \midrule

        \rowcolor{Gray!40} \textbf{Oxford-IIIT Pets} \\
        \midrule
        \texttt{a photo of a \{\}, a type of pet.} \\
        \midrule

        \rowcolor{Gray!40} \textbf{Caltech-101} \\
        \midrule
        \rowcolor{Apricot!0}  \texttt{a photo of a \{\}.} \\
        \rowcolor{Apricot!50} \texttt{a painting of a \{\}.} \\
        \rowcolor{Apricot!0}  \texttt{a plastic \{\}.} \\
        \rowcolor{Apricot!50} \texttt{a sculpture of a \{\}.} \\
        \rowcolor{Apricot!0}  \texttt{a sketch of a \{\}.} \\
        \rowcolor{Apricot!50} \texttt{a tattoo of a \{\}.} \\
        \rowcolor{Apricot!0}  \texttt{a toy \{\}.} \\
        \rowcolor{Apricot!50} \texttt{a rendition of a \{\}.} \\
        \rowcolor{Apricot!0}  \texttt{a embroidered \{\}.} \\
        \rowcolor{Apricot!50} \texttt{a cartoon \{\}.} \\
        \rowcolor{Apricot!0}  \texttt{a \{\} in a video game.} \\
        \rowcolor{Apricot!50} \texttt{a plushie \{\}.} \\
        \rowcolor{Apricot!0}  \texttt{a origami \{\}.} \\
        \rowcolor{Apricot!50} \texttt{art of a \{\}.} \\
        \rowcolor{Apricot!0}  \texttt{graffiti of a \{\}.} \\
        \rowcolor{Apricot!50} \texttt{a drawing of a \{\}.} \\
        \rowcolor{Apricot!0}  \texttt{a doodle of a \{\}.} \\
        \rowcolor{Apricot!50} \texttt{a photo of the \{\}.} \\
        \rowcolor{Apricot!0}  \texttt{a painting of the \{\}.} \\
        \rowcolor{Apricot!50} \texttt{the plastic \{\}.} \\
        \rowcolor{Apricot!0}  \texttt{a sculpture of the \{\}.} \\
        \rowcolor{Apricot!50} \texttt{a sketch of the \{\}.} \\
        \rowcolor{Apricot!0}  \texttt{a tattoo of the \{\}.} \\
        \rowcolor{Apricot!50} \texttt{the toy \{\}.} \\
        \rowcolor{Apricot!0}  \texttt{a rendition of the \{\}.} \\
        \rowcolor{Apricot!50} \texttt{the embroidered \{\}.} \\
        \rowcolor{Apricot!0}  \texttt{the cartoon \{\}.} \\
        \rowcolor{Apricot!50} \texttt{the \{\} in a video game.} \\
        \rowcolor{Apricot!0}  \texttt{the plushie \{\}.} \\
        \rowcolor{Apricot!50} \texttt{the origami \{\}.} \\
        \rowcolor{Apricot!0}  \texttt{art of the \{\}.} \\
        \rowcolor{Apricot!50} \texttt{graffiti of the \{\}.} \\
        \rowcolor{Apricot!0}  \texttt{a drawing of the \{\}.} \\
        \rowcolor{Apricot!50} \texttt{a doodle of the \{\}.} \\
        \midrule
    \end{tabularx}
    \hfill
    \renewcommand{\arraystretch}{1.17}
    \begin{tabularx}{0.325\linewidth}[t]{X}
        \midrule

        \rowcolor{Gray!40} \textbf{Oxford Flowers (our prompts)} \\
        \midrule
        \texttt{flowers : \{\}.} \\
        \midrule

        \rowcolor{Gray!40} \textbf{STL10} \\
        \midrule
        \rowcolor{Apricot!0}  \texttt{a photo of a \{\}.} \\
        \rowcolor{Apricot!50} \texttt{a photo of the \{\}.} \\
        \midrule

        \rowcolor{Gray!40} \textbf{EuroSAT} \\
        \midrule
        \rowcolor{Apricot!0}  \texttt{a centered satellite photo of \{\}.} \\
        \rowcolor{Apricot!50} \texttt{a centered satellite photo of a \{\}.} \\
        \rowcolor{Apricot!0}  \texttt{a centered satellite photo of the \{\}.} \\
        \midrule

        \rowcolor{Gray!40} \textbf{RESISC} \\
        \midrule
        \rowcolor{Apricot!0}  \texttt{satellite imagery of \{\}.} \\
        \rowcolor{Apricot!50} \texttt{aerial imagery of \{\}.} \\
        \rowcolor{Apricot!0}  \texttt{satellite photo of \{\}.} \\
        \rowcolor{Apricot!50} \texttt{aerial photo of \{\}.} \\
        \rowcolor{Apricot!0}  \texttt{satellite view of \{\}.} \\
        \rowcolor{Apricot!50} \texttt{aerial view of \{\}.} \\
        \rowcolor{Apricot!0}  \texttt{satellite imagery of a \{\}.} \\
        \rowcolor{Apricot!50} \texttt{aerial imagery of a \{\}.} \\
        \rowcolor{Apricot!0}  \texttt{satellite photo of a \{\}.} \\
        \rowcolor{Apricot!50} \texttt{aerial photo of a \{\}.} \\
        \rowcolor{Apricot!0}  \texttt{satellite view of a \{\}.} \\
        \rowcolor{Apricot!50} \texttt{aerial view of a \{\}.} \\
        \rowcolor{Apricot!0}  \texttt{satellite imagery of the \{\}.} \\
        \rowcolor{Apricot!50} \texttt{aerial imagery of the \{\}.} \\
        \rowcolor{Apricot!0}  \texttt{satellite photo of the \{\}.} \\
        \rowcolor{Apricot!50} \texttt{aerial photo of the \{\}.} \\
        \rowcolor{Apricot!0}  \texttt{satellite view of the \{\}.} \\
        \rowcolor{Apricot!50} \texttt{aerial view of the \{\}.} \\
        \midrule

        \rowcolor{Gray!40} \textbf{Country211} \\
        \midrule
        \rowcolor{Apricot!0}  \texttt{a photo i took in \{\}.} \\
        \rowcolor{Apricot!50} \texttt{a photo i took while visiting \{\}.} \\
        \rowcolor{Apricot!0}  \texttt{a photo from my home country of \{\}.} \\
        \rowcolor{Apricot!50} \texttt{a photo from my visit to \{\}.} \\
        \rowcolor{Apricot!0}  \texttt{a photo showing the country of \{\}.} \\
        \midrule

        \rowcolor{Gray!40} \textbf{MNIST} \\
        \midrule
        \rowcolor{Apricot!0}  \texttt{a photo of the number: "\{\}".} \\
        \midrule

        \rowcolor{Gray!40} \textbf{CLEVR} \\
        \midrule
        \rowcolor{Apricot!0}  \texttt{a photo of \{\} objects.} \\
        \midrule

        \rowcolor{Gray!40} \textbf{Patch Camelyon} \\
        \midrule
        \rowcolor{Apricot!0}  \texttt{this is a photo of \{\}.} \\
        \midrule

        \rowcolor{Gray!40} \textbf{Rendered SST2} \\
        \midrule
        \rowcolor{Apricot!0}  \texttt{a \{\} review of a movie.} \\
        \midrule
    \end{tabularx}
\end{table*}

%% file: sections/appendix_traversals.tex
\section{Image traversals: more details and results}
\label{sec:meru_appendix_traversals}

Our qualitative analysis in \Cref{sec:meru_qualitative} involves inferring the learned visual-semantic hierarchy
in the representation space through image traversals.
We performed shortest-path traversal from a given image embedding $\vy$
to the \texttt{[ROOT]} embedding by interpolating 50 equally spaced steps.
At each step, we retrieve text from a set $\mathcal{X}$ of text embeddings (including \texttt{[ROOT]}).
Here we include the precise methodology details to perform image traversals.

MERU and CLIP have different methods for interpolation and nearest-neighbor retrieval
due to the difference in geometric properties of Euclidean and hyperbolic spaces.

\paragraph{Interpolating steps:}
\begin{compactitem}[\hspace{1pt}--]
  \item \textbf{CLIP:} We linearly interpolate between $L^2$ normalized embeddings of $\vy$ and \texttt{[ROOT]},
  and then $L^2$ normalize all \emph{step} embeddings.
  In PyTorch \citep{pytorch}, \texttt{torch.lerp} can perform this linear interpolation.
  \item \textbf{MERU:} We linearly interpolate in the tangent space,
  between $\vv = \text{logm}_\vO (\vy)$ (\cref{eqn:meru_logmap}) and $\vO$ (origin is \texttt{[ROOT]}),
  then lift all \emph{step} embeddings onto the hyperboloid.
\end{compactitem}

\paragraph{Nearest-neighbor text retrieval:}
\begin{compactitem}[\hspace{1pt}--]
  \item \textbf{CLIP:} We select $\vx \in \mathcal{X}$ having the highest \emph{cosine similarity} with the \emph{step} embedding.
  \item \textbf{MERU:} First we create a subset $\mathcal{X}_e \subset \mathcal{X}$ of text embeddings
  that \emph{entail} the given \emph{step} embedding, \ie{} \cref{eqn:meru_entailment} evaluates to 0
  (note that \texttt{[ROOT]} entails everything).
  Then we select $\vx \in \mathcal{X}_e$ having the highest \emph{Lorentzian inner product} with the \emph{step} embedding.
\end{compactitem}

At any given step, the caption associated with the retrieved texct embedding $\vx$ (or \texttt{[ROOT]}) is the retrieved nearest neighbor.
We observed that multiple consecutive \emph{steps} retrieve the same caption,
so our results only display \emph{unique} captions encountered during the traversal.

\paragraph{Caption sources:}
We create the set of text embeddings $\mathcal{X}$ using captions collected from two different sources.

\begin{figure}[t]
  \centering
  \frame{\includegraphics[width=\linewidth]{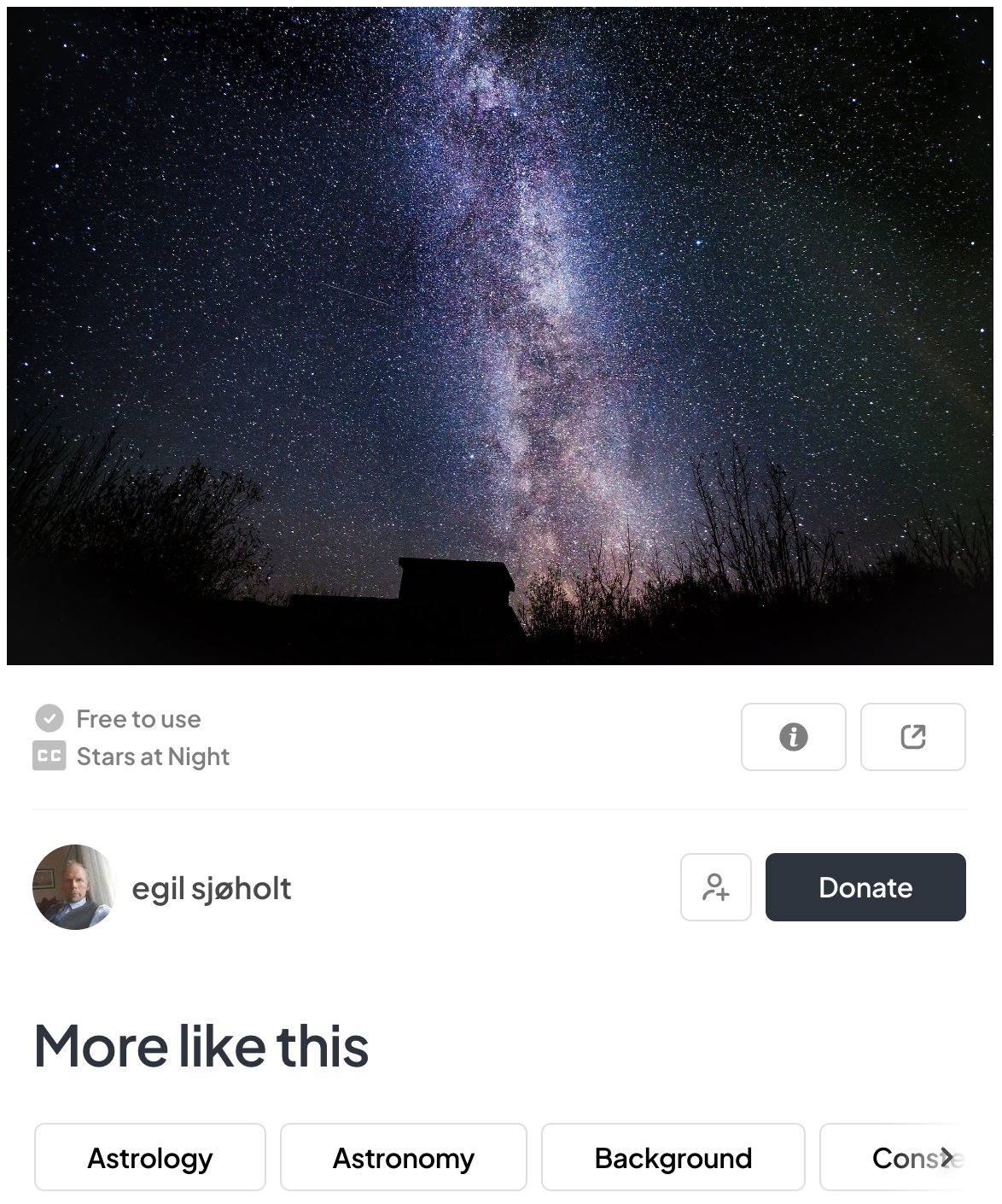}}
  \caption{
    \textbf{\href{https://www.pexels.com}{\texttt{pexels.com}} webpage}
    of an image used in our results.
    We manually collect the closed caption (\emph{CC}) and \emph{`More like this'} tags for all images to create the retrieval set for image traversals.
  }
  \label{fig:meru_appendix_pexels_example}
\end{figure}

\begin{compactitem}[\hspace{1pt}--]
  \item \textbf{\href{https://www.pexels.com}{\texttt{pexels.com}} metadata:}
  We manually collect metadata (\Cref{fig:meru_appendix_pexels_example}), then filter tags to only keep nouns and adjectives (total 750 captions and tags).
  We filter by performing parts-of-speech using the RoBERTa~\citep{liu2019roberta} model (\texttt{en-core-web-trf}) from SpaCy~\cite{spacy} library.
  Finally, we convert tags to captions by filling prompts -- \texttt{`a photo of \{\}.'} for nouns, and \texttt{`this photo is \{\}.'} for adjectives.
  \item \textbf{YFCC dataset:}
  We use the \texttt{text descriptions} of the YFCC-15M subset \citep{radford2021clip}.
  We perform minimal text processing of these captions according to RedCaps~\citep{desai2021redcaps}, to match the training data distribution.
  This involves converting to lowercase, using \texttt{ftfy}~\cite{speer2019ftfy} to strips accents and non-latin characters,
  removing all sub-strings enclosed in brackets (\texttt{(.*)}, \texttt{[.*]}),
  and replacing social media handles (words starting with `@') with a \texttt{$<$usr$>$}.
  We also remove captions having more than 20 tokens (for ease of visualization).
  Finally, we obtain $\approx$8.7M captions.
\end{compactitem}

\paragraph{Results:}
\Cref{fig:meru_main_image_traversals} shows selected qualitative examples with 8 out of 60 images.
On the next pages,
\Cref{fig:meru_appendix_pexels_locations,fig:meru_appendix_pexels_food,fig:meru_appendix_pexels_flora_fauna,fig:meru_appendix_pexels_objects1,fig:meru_appendix_pexels_objects2}
include results with other 52 images.
After the image credits (\Cref{sec:meru_appendix_image_credits}), we display results with YFCC captions.

\clearpage
\onecolumn

\input{figtabs/appendix_traversals_pexels.tex}
\section{Image credits}
\label{sec:meru_appendix_image_credits}

All images displayed in this paper are collected from \href{www.pexels.com}{\texttt{pexels.com}},
a photography website that offers images with permissible usage licenses.
Below is the list of the image source URLs listed in order of their appearance in the paper.
We thank all the photographers for generously sharing these images.

\paragraph{Illustration of the visual-semantic hierarchy (\Cref{fig:meru_intro}).}

\begin{compactitem}[\hspace{1pt}--]
    \item \url{www.pexels.com/photo/adult-yellow-labrador-retriever-standing-on-snow-field-1696589}
    \item \url{www.pexels.com/photo/homeless-cat-fighting-with-dog-on-street-6601811}
    \item \url{www.pexels.com/photo/short-coated-gray-cat-20787}
\end{compactitem}

\paragraph{Image traversals -- results in the main paper (\Cref{fig:meru_main_image_traversals}).}

\begin{compactenum}[\hspace{1pt}(1)]
    \setcounter{enumi}{0}
    \item \url{www.pexels.com/photo/a-bengal-cat-sitting-beside-wheatgrass-on-a-white-surface-7123957}
    \item \url{www.pexels.com/photo/white-horse-running-on-green-field-1996337}
    \item \url{www.pexels.com/photo/photography-of-rainbow-during-cloudy-sky-757239}
    \item \url{www.pexels.com/photo/retro-photo-camera-on-table-7162551}
    \item \url{www.pexels.com/photo/avocado-toast-served-on-white-plate-10464867}
    \item \url{www.pexels.com/photo/photo-of-brooklyn-bridge-new-york-2260783}
    \item \url{www.pexels.com/photo/taj-mahal-through-an-arch-2413613}
    \item \url{www.pexels.com/photo/sydney-opera-house-7088958}
\end{compactenum}

\paragraph{Image traversals -- locations and landmarks (\Cref{fig:meru_appendix_pexels_locations}).}

\begin{compactenum}[\hspace{1pt}(1)]
    \setcounter{enumi}{8}
    \item \url{www.pexels.com/photo/golden-gate-bridge-san-francisco-california-1141853}
    \item \url{www.pexels.com/photo/white-cliffs-of-dover-in-england-9692909}
    \item \url{www.pexels.com/photo/the-famous-fountain-paint-pots-in-yellowstone-national-park-12767016}
    \item \url{www.pexels.com/photo/the-parthenon-temple-ruins-in-athens-greece-14446783}
    \item \url{www.pexels.com/photo/famous-big-ben-under-cloudy-sky-14434677}
    \item \url{www.pexels.com/photo/karlskirche-church-7018621}
    \item \url{www.pexels.com/photo/mt-fuji-3408353}
    \item \url{www.pexels.com/photo/horseshoe-bend-arizona-2563733}
    \item \url{www.pexels.com/photo/stars-at-night-1906667}
    \item \url{www.pexels.com/photo/volcano-erupting-at-night-under-starry-sky-4220967}
    \item \url{www.pexels.com/photo/northern-lights-1933319}
    \item \url{www.pexels.com/photo/attraction-building-city-hotel-415999}
\end{compactenum}

\paragraph{Image traversals -- flora and fauna (\Cref{fig:meru_appendix_pexels_flora_fauna}).}

\begin{compactenum}[\hspace{1pt}(1)]
    \setcounter{enumi}{20}
    \item \url{www.pexels.com/photo/squirrel-up-on-the-snow-covered-tree-15306429}
    \item \url{www.pexels.com/photo/a-seagull-flying-under-blue-sky-12509256}
    \item \url{www.pexels.com/photo/cute-pug-sitting-on-floor-in-white-kitchen-11199295}
    \item \url{www.pexels.com/photo/three-zebras-2118645}
    \item \url{www.pexels.com/photo/monarch-butterfly-perching-on-red-flower-1557208}
    \item \url{www.pexels.com/photo/red-hibiscus-in-bloom-5801054}
    \item \url{www.pexels.com/photo/white-chicken-on-green-grass-field-58902}
    \item \url{www.pexels.com/photo/yellow-blue-and-white-macaw-perched-on-brown-tree-branch-12715261}
    \item \url{www.pexels.com/photo/closeup-photo-of-red-and-white-mushroom-757292}
    \item \url{www.pexels.com/photo/photo-of-jellyfish-lot-underwater-3616240}
    \item \url{www.pexels.com/photo/yellow-labrador-retriever-wearing-red-cap-4588002}
    \item \url{www.pexels.com/photo/an-orca-whale-jumping-out-of-the-water-7767974}
\end{compactenum}

\paragraph{Image traversals -- food and drinks (\Cref{fig:meru_appendix_pexels_food}).}

\begin{compactenum}[\hspace{1pt}(1)]
    \setcounter{enumi}{32}
    \item \url{www.pexels.com/photo/bread-and-coffee-for-breakfast-15891938}
    \item \url{www.pexels.com/photo/grilled-cheese-on-a-plate-14941252}
    \item \url{www.pexels.com/photo/bowl-of-ramen-12984979}
    \item \url{www.pexels.com/photo/green-chili-peppers-and-a-knife-5792428}
    \item \url{www.pexels.com/photo/spinach-caprese-salad-on-white-ceramic-plate-4768996}
    \item \url{www.pexels.com/photo/chocolate-cupcakes-635409}
    \item \url{www.pexels.com/photo/pav-bhaji-dish-on-a-bowl-5410400}
    \item \url{www.pexels.com/photo/clear-glass-bottle-filled-with-broccoli-shake-1346347}
    \item \url{www.pexels.com/photo/vada-pav-15017417}
    \item \url{www.pexels.com/photo/old-fashioned-cocktail-drink-4762719}
    \item \url{www.pexels.com/photo/coffee-in-white-ceramic-teacup-on-white-ceramic-suacer-894696}
    \item \url{www.pexels.com/photo/espresso-martini-in-close-up-photography-15082368}
\end{compactenum}

\paragraph{Image traversals -- objects and scenes (\Cref{fig:meru_appendix_pexels_objects1}).}

\begin{compactenum}[\hspace{1pt}(1)]
    \setcounter{enumi}{44}
    \item \url{www.pexels.com/photo/photograph-of-a-burning-fire-672636}
    \item \url{www.pexels.com/photo/white-clouds-in-blue-sky-8354530}
    \item \url{www.pexels.com/photo/raining-in-the-city-2448749}
    \item \url{www.pexels.com/photo/aerial-view-of-road-in-the-middle-of-trees-1173777}
    \item \url{www.pexels.com/photo/mountain-bike-on-the-beach-10542237}
    \item \url{www.pexels.com/photo/wax-candles-burning-on-ground-14184952}
    \item \url{www.pexels.com/photo/white-wooden-shelf-beside-bed-2062431}
    \item \url{www.pexels.com/photo/stainless-steel-faucet-on-white-ceramic-sink-3761560}
    \item \url{www.pexels.com/photo/jack-o-lantern-with-light-5659699}
    \item \url{www.pexels.com/photo/black-and-white-piano-keys-4077310}
    \item \url{www.pexels.com/photo/assorted-gift-boxes-on-floor-near-christmas-tree-3394779}
    \item \url{www.pexels.com/photo/garden-table-and-chair-14831985}
\end{compactenum}

\paragraph{Image traversals -- objects and scenes (\Cref{fig:meru_appendix_pexels_objects2}).}

\begin{compactenum}[\hspace{1pt}(1)]
    \setcounter{enumi}{56}
    \item \url{www.pexels.com/photo/turned-on-floor-lamp-near-sofa-on-a-library-room-1907784}
    \item \url{www.pexels.com/photo/ripe-pineapple-on-gray-rock-beside-body-of-water-29555}
    \item \url{www.pexels.com/photo/close-up-shot-of-a-cockatiel-13511241}
    \item \url{www.pexels.com/photo/antique-bills-business-cash-210600}
\end{compactenum}

\clearpage

\section*{Image traversals with YFCC captions.}

\input{figtabs/appendix_traversals_yfcc.tex}

\twocolumn

%% file: figtabs/appendix_traversals_pexels.tex
\begin{figure*}[t!]
    \footnotesize
    \newcolumntype{Z}{>{\centering\arraybackslash}X}
    \setlength{\tabcolsep}{1pt}

    \hfill
    \frame{\includegraphics[width=0.18\linewidth]{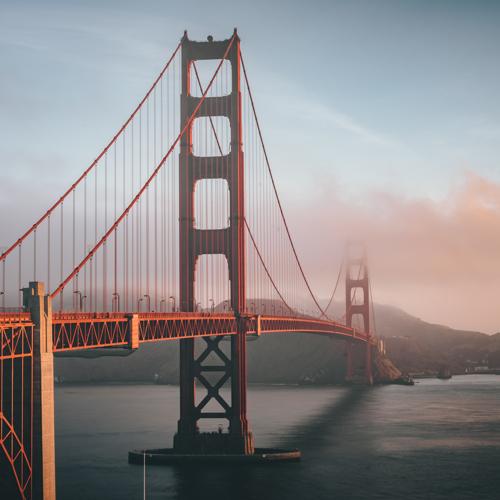}} \hfill \hfill
    \frame{\includegraphics[width=0.18\linewidth]{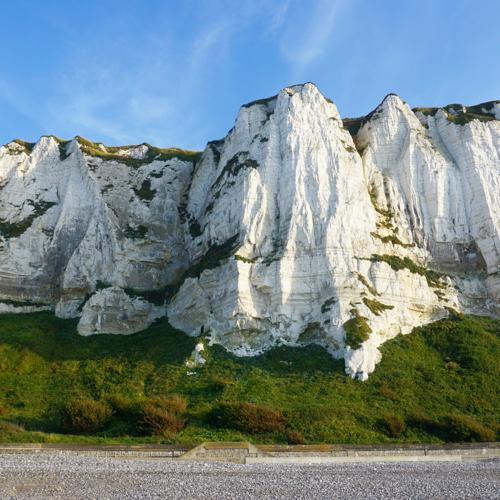}} \hfill \hfill
    \frame{\includegraphics[width=0.18\linewidth]{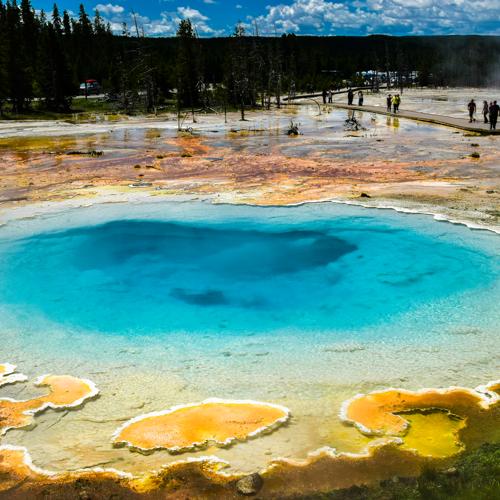}} \hfill \hfill
    \frame{\includegraphics[width=0.18\linewidth]{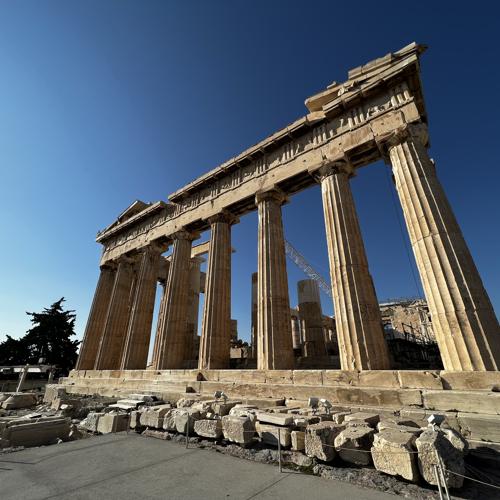}} \hfill \hfill

    \begin{tabularx}{0.24\linewidth}[t]{ZZ}
        \rowcolor{Gray!40} \textbf{MERU} & \textbf{CLIP} \\
        \midrule
        \rowcolor{Apricot!100} \emph{golden gate} & \emph{golden gate bridge, san francisco, california} \\
        \midrule
        \rowcolor{Apricot!80} \emph{san francisco} & \emph{famous landmark} \\
        \midrule
        \rowcolor{Apricot!60} \emph{tourist spot} & $\downarrow$ \\
        \midrule
        \rowcolor{Apricot!40} \emph{photo} & $\downarrow$ \\
        \midrule
        \rowcolor{Apricot!20} \emph{power} & $\downarrow$ \\
        \midrule
        \texttt{\textbf{[ROOT]}} & \texttt{\textbf{[ROOT]}} \\
    \end{tabularx}
    \hfill
    \begin{tabularx}{0.24\linewidth}[t]{ZZ}
        \rowcolor{Gray!40} \textbf{MERU} & \textbf{CLIP} \\
        \midrule
        \rowcolor{Apricot!100} \emph{white cliffs of dover in england} & \emph{white cliffs of dover} \\
        \midrule
        \rowcolor{Apricot!80} \emph{white cliffs of dover} & \emph{cliffs} \\
        \midrule
        \rowcolor{Apricot!60} \emph{white} & \emph{rocky} \\
        \midrule
        \rowcolor{Apricot!40} \emph{coast} & $\downarrow$ \\
        \midrule
        \rowcolor{Apricot!20} \emph{country} & $\downarrow$ \\
        \midrule
        \texttt{\textbf{[ROOT]}} & \texttt{\textbf{[ROOT]}} \\
    \end{tabularx}
    \hfill
    \begin{tabularx}{0.24\linewidth}[t]{ZZ}
        \rowcolor{Gray!40} \textbf{MERU} & \textbf{CLIP} \\
        \midrule
        \rowcolor{Apricot!100} \emph{the famous fountain paint pots in yellowstone national park} & \emph{yellowstone} \\
        \midrule
        \rowcolor{Apricot!73} \emph{yellowstone} & \emph{beauty} \\
        \midrule
        \rowcolor{Apricot!46} \emph{national park} & $\downarrow$ \\
        \midrule
        \rowcolor{Apricot!20} \emph{power} & $\downarrow$ \\
        \midrule
        \texttt{\textbf{[ROOT]}} & \texttt{\textbf{[ROOT]}} \\
    \end{tabularx}
    \hfill
    \begin{tabularx}{0.24\linewidth}[t]{ZZ}
        \rowcolor{Gray!40} \textbf{MERU} & \textbf{CLIP} \\
        \midrule
        \rowcolor{Apricot!100} \emph{the parthenon temple ruins in athens greece} & \emph{the parthenon temple ruins in athens greece} \\
        \midrule
        \rowcolor{Apricot!73} \emph{historical site} & \emph{famous landmark} \\
        \midrule
        \rowcolor{Apricot!46} \emph{architecture} & \emph{low angle shot} \\
        \midrule
        \rowcolor{Apricot!20} \emph{domestic} & $\downarrow$ \\
        \midrule
        \texttt{\textbf{[ROOT]}} & \texttt{\textbf{[ROOT]}} \\
    \end{tabularx}

    \vspace{5pt}
    \hfill
    \frame{\includegraphics[width=0.18\linewidth]{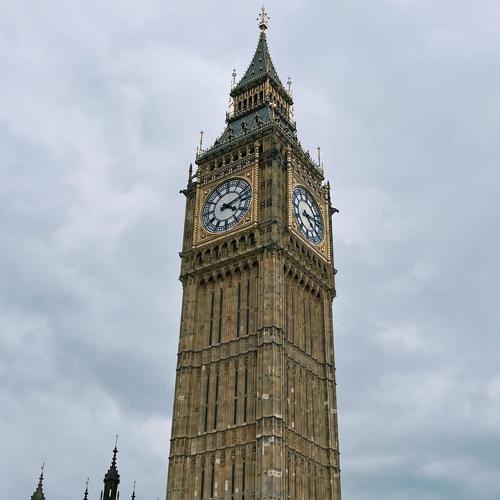}} \hfill \hfill
    \frame{\includegraphics[width=0.18\linewidth]{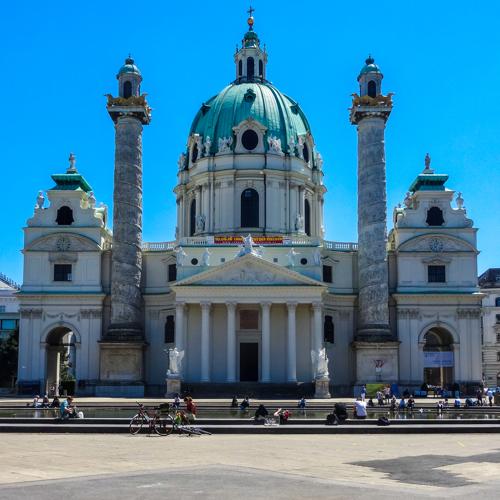}} \hfill \hfill
    \frame{\includegraphics[width=0.18\linewidth]{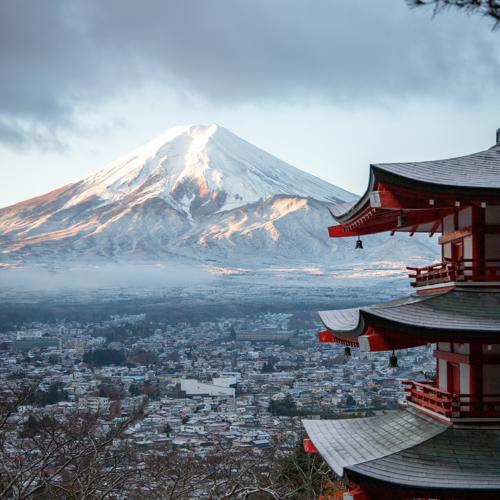}} \hfill \hfill
    \frame{\includegraphics[width=0.18\linewidth]{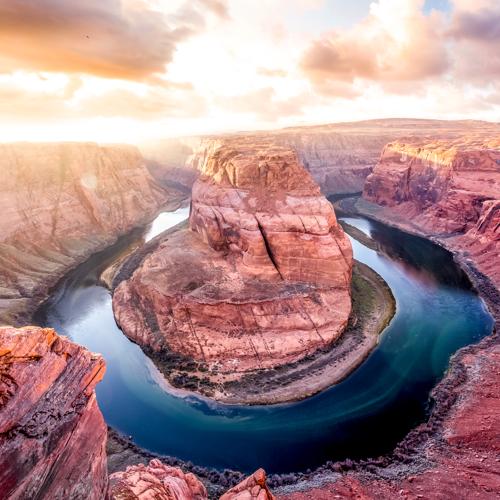}} \hfill \hfill

    \begin{tabularx}{0.24\linewidth}[t]{ZZ}
        \rowcolor{Gray!40} \textbf{MERU} & \textbf{CLIP} \\
        \midrule
        \rowcolor{Apricot!100} \emph{big ben} & \emph{big ben} \\
        \midrule
        \rowcolor{Apricot!60} \emph{holiday} & $\downarrow$ \\
        \midrule
        \rowcolor{Apricot!20} \emph{day} & $\downarrow$ \\
        \midrule
        \texttt{\textbf{[ROOT]}} & \texttt{\textbf{[ROOT]}} \\
    \end{tabularx}
    \hfill
    \begin{tabularx}{0.24\linewidth}[t]{ZZ}
        \rowcolor{Gray!40} \textbf{MERU} & \textbf{CLIP} \\
        \midrule
        \rowcolor{Apricot!100} \emph{karlskirche} & \emph{karlskirche church} \\
        \midrule
        \rowcolor{Apricot!60} \emph{architecture} & \emph{church} \\
        \midrule
        \rowcolor{Apricot!20} \emph{style} & $\downarrow$ \\
        \midrule
        \texttt{\textbf{[ROOT]}} & \texttt{\textbf{[ROOT]}} \\
    \end{tabularx}
    \hfill
    \begin{tabularx}{0.24\linewidth}[t]{ZZ}
        \rowcolor{Gray!40} \textbf{MERU} & \textbf{CLIP} \\
        \midrule
        \rowcolor{Apricot!100} \emph{fuji} & \emph{fuji} \\
        \midrule
        \rowcolor{Apricot!60} \emph{japan} & \emph{cozy} \\
        \midrule
        \rowcolor{Apricot!20} \emph{holiday} & $\downarrow$ \\
        \midrule
        \texttt{\textbf{[ROOT]}} & \texttt{\textbf{[ROOT]}} \\
    \end{tabularx}
    \hfill
    \begin{tabularx}{0.24\linewidth}[t]{ZZ}
        \rowcolor{Gray!40} \textbf{MERU} & \textbf{CLIP} \\
        \midrule
        \rowcolor{Apricot!100} \emph{horseshoe bend} & \emph{horseshoe bend} \\
        \midrule
        \rowcolor{Apricot!60} \emph{outdoors} & \emph{national park} \\
        \midrule
        \rowcolor{Apricot!20} $\downarrow$ & \emph{credit} \\
        \midrule
        \texttt{\textbf{[ROOT]}} & \texttt{\textbf{[ROOT]}} \\
    \end{tabularx}

    \vspace{5pt}
    \hfill
    \frame{\includegraphics[width=0.18\linewidth]{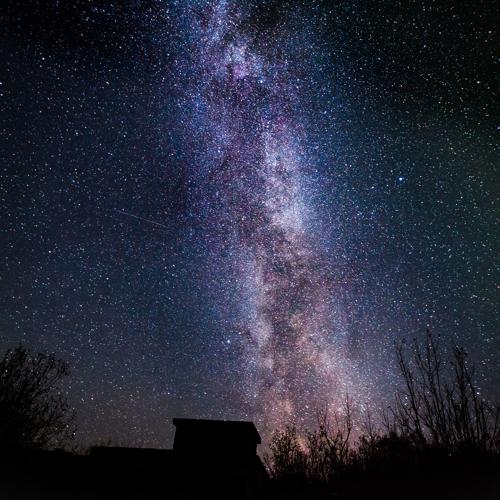}} \hfill \hfill
    \frame{\includegraphics[width=0.18\linewidth]{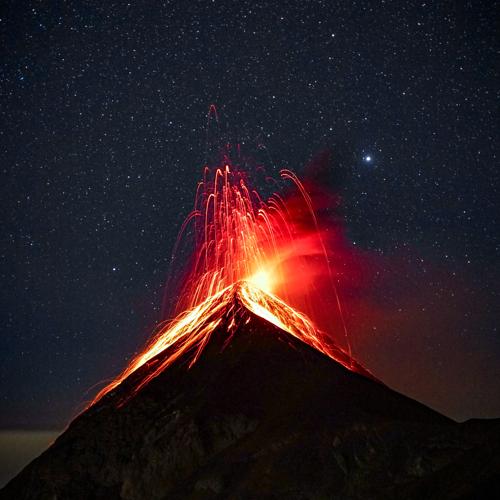}} \hfill \hfill
    \frame{\includegraphics[width=0.18\linewidth]{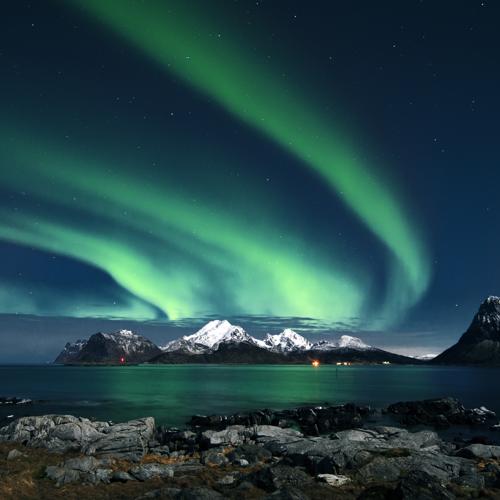}} \hfill \hfill
    \frame{\includegraphics[width=0.18\linewidth]{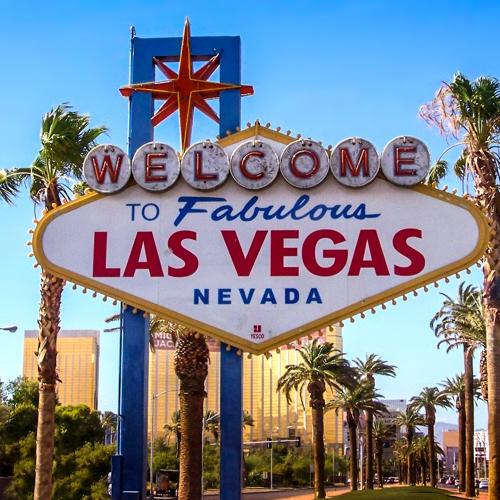}} \hfill \hfill

    \begin{tabularx}{0.24\linewidth}[t]{ZZ}
        \rowcolor{Gray!40} \textbf{MERU} & \textbf{CLIP} \\
        \midrule
        \rowcolor{Apricot!100} \emph{milky way} & $\downarrow$ \\
        \midrule
        \rowcolor{Apricot!20} \emph{rural} & $\downarrow$ \\
        \midrule
        \texttt{\textbf{[ROOT]}} & \texttt{\textbf{[ROOT]}} \\
    \end{tabularx}
    \hfill
    \begin{tabularx}{0.24\linewidth}[t]{ZZ}
        \rowcolor{Gray!40} \textbf{MERU} & \textbf{CLIP} \\
        \midrule
        \rowcolor{Apricot!100} \emph{volcano erupting at night under starry sky} & \emph{volcano erupting at night under starry sky} \\
        \midrule
        \rowcolor{Apricot!60} \emph{active volcano} & \emph{volcanic} \\
        \midrule
        \rowcolor{Apricot!20} \emph{outdoors} & $\downarrow$ \\
        \midrule
        \texttt{\textbf{[ROOT]}} & \texttt{\textbf{[ROOT]}} \\
    \end{tabularx}
    \hfill
    \begin{tabularx}{0.24\linewidth}[t]{ZZ}
        \rowcolor{Gray!40} \textbf{MERU} & \textbf{CLIP} \\
        \midrule
        \rowcolor{Apricot!100} \emph{northern lights norway} & \emph{northern lights norway} \\
        \midrule
        \rowcolor{Apricot!73} \emph{aurora} & \emph{aurora} \\
        \midrule
        \rowcolor{Apricot!46} \emph{scenic} & \emph{outdoors} \\
        \midrule
        \rowcolor{Apricot!20} \emph{outdoor} & $\downarrow$ \\
        \midrule
        \texttt{\textbf{[ROOT]}} & \texttt{\textbf{[ROOT]}} \\
    \end{tabularx}
    \hfill
    \begin{tabularx}{0.24\linewidth}[t]{ZZ}
        \rowcolor{Gray!40} \textbf{MERU} & \textbf{CLIP} \\
        \midrule
        \rowcolor{Apricot!100} \emph{california} & \emph{welcome to fabulous las vegas nevada signage} \\
        \midrule
        \rowcolor{Apricot!20} $\downarrow$ & \emph{famous landmark} \\
        \midrule
        \texttt{\textbf{[ROOT]}} & \texttt{\textbf{[ROOT]}} \\
    \end{tabularx}
    \caption{
        \textbf{Image traversals with MERU and CLIP (locations and landmarks).}
        Retrieved captions are sourced from \href{https://www.pexels.com}{\texttt{pexels.com}} metadata.
        MERU captures a more systematic and fine-grained visual-semantic hierarchy than CLIP --
        trends are same as \Cref{fig:meru_main_image_traversals}.
      }
      \label{fig:meru_appendix_pexels_locations}
\end{figure*}

\clearpage

\begin{figure*}[t!]
    \footnotesize
    \newcolumntype{Z}{>{\centering\arraybackslash}X}
    \setlength{\tabcolsep}{1pt}

    \hfill
    \frame{\includegraphics[width=0.18\linewidth]{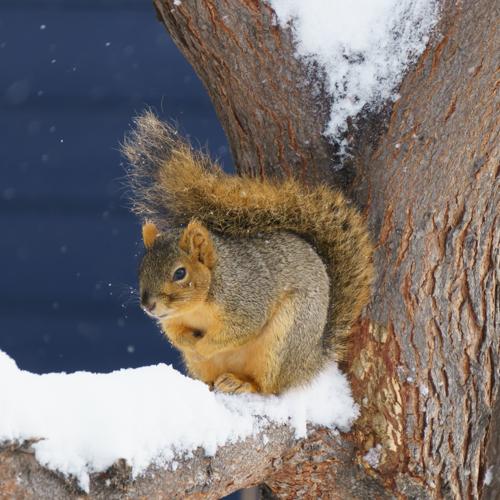}} \hfill \hfill
    \frame{\includegraphics[width=0.18\linewidth]{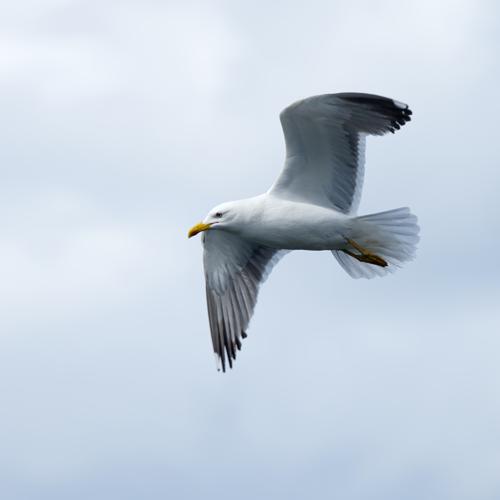}} \hfill \hfill
    \frame{\includegraphics[width=0.18\linewidth]{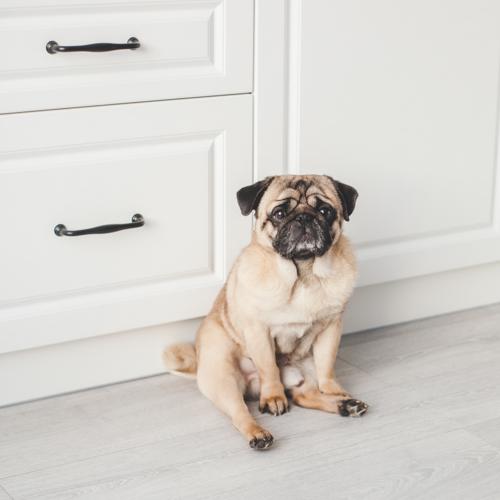}} \hfill \hfill
    \frame{\includegraphics[width=0.18\linewidth]{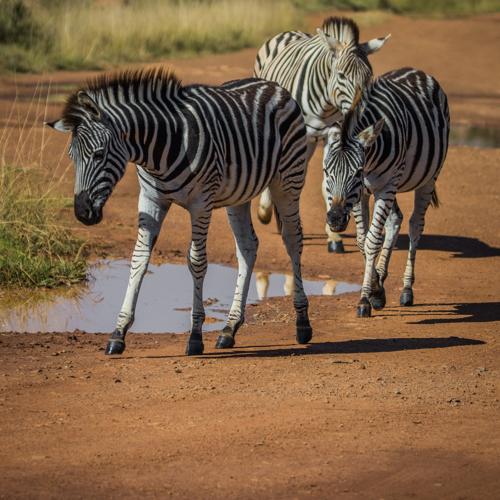}} \hfill \hfill

    \begin{tabularx}{0.24\linewidth}[t]{ZZ}
        \rowcolor{Gray!40} \textbf{MERU} & \textbf{CLIP} \\
        \midrule
        \rowcolor{Apricot!100} \emph{squirrel up on the snow covered tree} & \emph{squirrel up on the snow covered tree} \\
        \midrule
        \rowcolor{Apricot!73} \emph{squirrel} & \emph{squirrel} \\
        \midrule
        \rowcolor{Apricot!46} \emph{wildlife} & $\downarrow$ \\
        \midrule
        \rowcolor{Apricot!20} \emph{fluffy} & $\downarrow$ \\
        \midrule
        \texttt{\textbf{[ROOT]}} & \texttt{\textbf{[ROOT]}} \\
    \end{tabularx}
    \hfill
    \begin{tabularx}{0.24\linewidth}[t]{ZZ}
        \rowcolor{Gray!40} \textbf{MERU} & \textbf{CLIP} \\
        \midrule
        \rowcolor{Apricot!100} \emph{seagull} & \emph{seagull} \\
        \midrule
        \rowcolor{Apricot!80} \emph{bird} & \emph{bird} \\
        \midrule
        \rowcolor{Apricot!60} \emph{air} & $\downarrow$ \\
        \midrule
        \rowcolor{Apricot!40} \emph{coast} & $\downarrow$ \\
        \midrule
        \rowcolor{Apricot!20} \emph{day} & $\downarrow$ \\
        \midrule
        \texttt{\textbf{[ROOT]}} & \texttt{\textbf{[ROOT]}} \\
    \end{tabularx}
    \hfill
    \begin{tabularx}{0.24\linewidth}[t]{ZZ}
        \rowcolor{Gray!40} \textbf{MERU} & \textbf{CLIP} \\
        \midrule
        \rowcolor{Apricot!100} \emph{cute pug sitting on floor in white kitchen} & \emph{cute pug sitting on floor in white kitchen} \\
        \midrule
        \rowcolor{Apricot!73} \emph{pug} & $\downarrow$ \\
        \midrule
        \rowcolor{Apricot!46} \emph{domestic} & $\downarrow$ \\
        \midrule
        \rowcolor{Apricot!20} \emph{little} & $\downarrow$ \\
        \midrule
        \texttt{\textbf{[ROOT]}} & \texttt{\textbf{[ROOT]}} \\
    \end{tabularx}
    \hfill
    \begin{tabularx}{0.24\linewidth}[t]{ZZ}
        \rowcolor{Gray!40} \textbf{MERU} & \textbf{CLIP} \\
        \midrule
        \rowcolor{Apricot!100} \emph{three zebras} & \emph{three zebras} \\
        \midrule
        \rowcolor{Apricot!80} \emph{zebras} & \emph{wild animals} \\
        \midrule
        \rowcolor{Apricot!60} \emph{safari} & $\downarrow$ \\
        \midrule
        \rowcolor{Apricot!40} \emph{animal photography} & $\downarrow$ \\
        \midrule
        \rowcolor{Apricot!20} \emph{wild} & $\downarrow$ \\
        \midrule
        \texttt{\textbf{[ROOT]}} & \texttt{\textbf{[ROOT]}} \\
    \end{tabularx}

    \vspace{5pt}
    \hfill
    \frame{\includegraphics[width=0.18\linewidth]{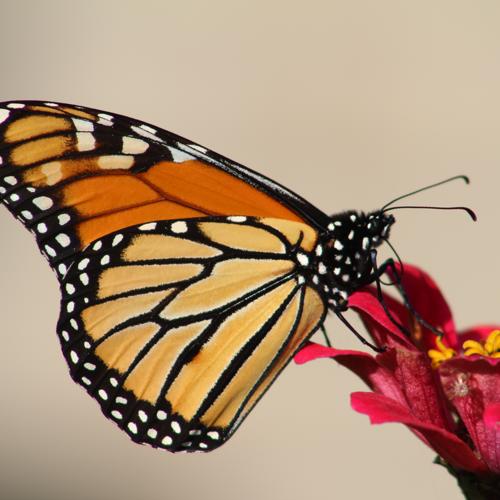}} \hfill \hfill
    \frame{\includegraphics[width=0.18\linewidth]{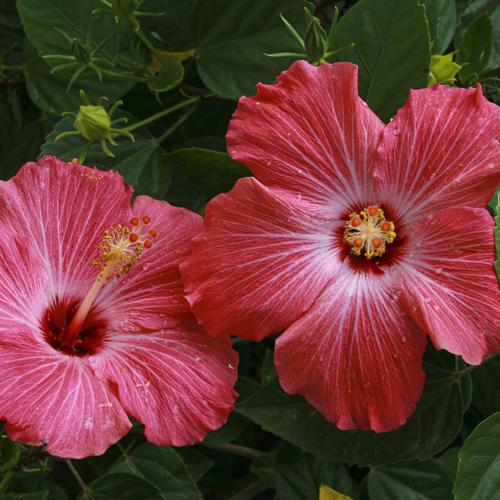}} \hfill \hfill
    \frame{\includegraphics[width=0.18\linewidth]{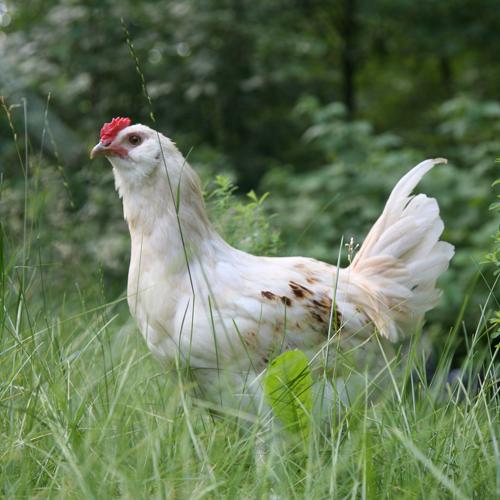}} \hfill \hfill
    \frame{\includegraphics[width=0.18\linewidth]{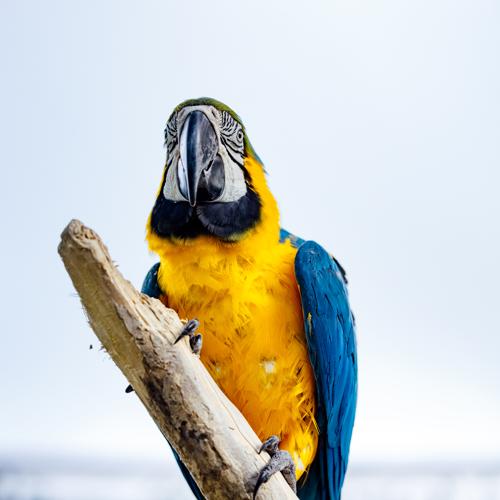}} \hfill \hfill

    \begin{tabularx}{0.24\linewidth}[t]{ZZ}
        \rowcolor{Gray!40} \textbf{MERU} & \textbf{CLIP} \\
        \midrule
        \rowcolor{Apricot!100} \emph{monarch butterfly perching on red flower} & \emph{monarch butterfly} \\
        \midrule
        \rowcolor{Apricot!80} \emph{monarch butterfly} & $\downarrow$ \\
        \midrule
        \rowcolor{Apricot!60} \emph{butterfly} & $\downarrow$ \\
        \midrule
        \rowcolor{Apricot!40} \emph{beauty} & $\downarrow$ \\
        \midrule
        \rowcolor{Apricot!20} \emph{day} & $\downarrow$ \\
        \midrule
        \texttt{\textbf{[ROOT]}} & \texttt{\textbf{[ROOT]}} \\
    \end{tabularx}
    \hfill
    \begin{tabularx}{0.24\linewidth}[t]{ZZ}
        \rowcolor{Gray!40} \textbf{MERU} & \textbf{CLIP} \\
        \midrule
        \rowcolor{Apricot!100} \emph{red hibiscus in bloom} & \emph{red hibiscus in bloom} \\
        \midrule
        \rowcolor{Apricot!73} \emph{hibiscus} & \emph{hibiscus} \\
        \midrule
        \rowcolor{Apricot!46} \emph{bloom} & \emph{blooming flowers} \\
        \midrule
        \rowcolor{Apricot!20} \emph{style} & $\downarrow$ \\
        \midrule
        \texttt{\textbf{[ROOT]}} & \texttt{\textbf{[ROOT]}} \\
    \end{tabularx}
    \hfill
    \begin{tabularx}{0.24\linewidth}[t]{ZZ}
        \rowcolor{Gray!40} \textbf{MERU} & \textbf{CLIP} \\
        \midrule
        \rowcolor{Apricot!100} \emph{white chicken on green grass field} & \emph{white chicken on green grass field} \\
        \midrule
        \rowcolor{Apricot!73} \emph{cockerel} & $\downarrow$ \\
        \midrule
        \rowcolor{Apricot!46} \emph{chicken} & $\downarrow$ \\
        \midrule
        \rowcolor{Apricot!20} \emph{style} & $\downarrow$ \\
        \midrule
        \texttt{\textbf{[ROOT]}} & \texttt{\textbf{[ROOT]}} \\
    \end{tabularx}
    \hfill
    \begin{tabularx}{0.24\linewidth}[t]{ZZ}
        \rowcolor{Gray!40} \textbf{MERU} & \textbf{CLIP} \\
        \midrule
        \rowcolor{Apricot!100} \emph{yellow blue and white macaw perched on brown tree branch} & \emph{yellow blue and white macaw perched on brown tree branch} \\
        \midrule
        \rowcolor{Apricot!73} \emph{parrot} & \emph{parrot} \\
        \midrule
        \rowcolor{Apricot!46} \emph{hungry} & \emph{animal} \\
        \midrule
        \rowcolor{Apricot!20} \emph{female} & $\downarrow$ \\
        \midrule
        \texttt{\textbf{[ROOT]}} & \texttt{\textbf{[ROOT]}} \\
    \end{tabularx}

    \vspace{5pt}
    \hfill
    \frame{\includegraphics[width=0.18\linewidth]{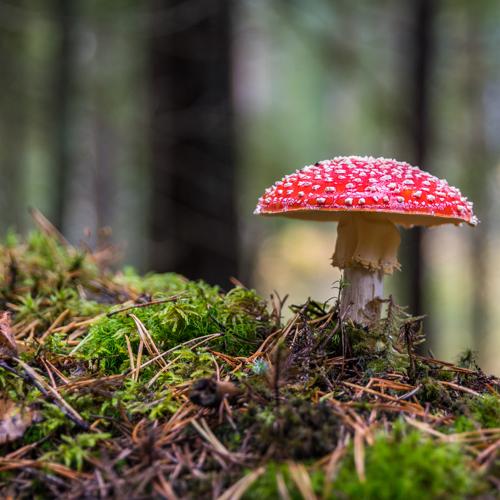}} \hfill \hfill
    \frame{\includegraphics[width=0.18\linewidth]{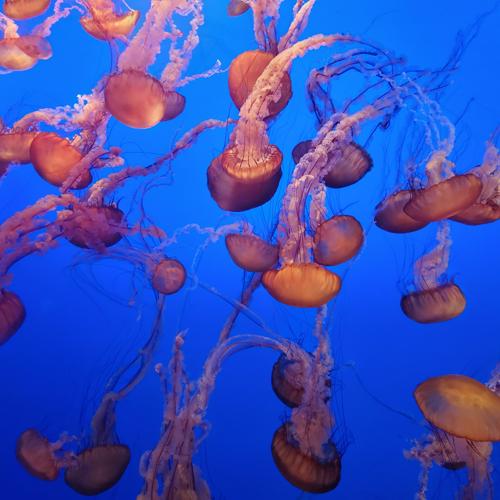}} \hfill \hfill
    \frame{\includegraphics[width=0.18\linewidth]{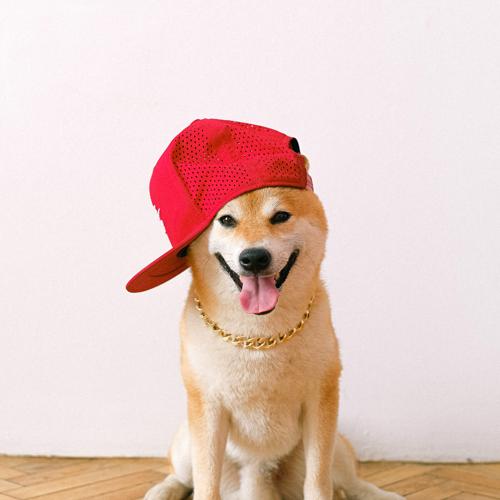}} \hfill \hfill
    \frame{\includegraphics[width=0.18\linewidth]{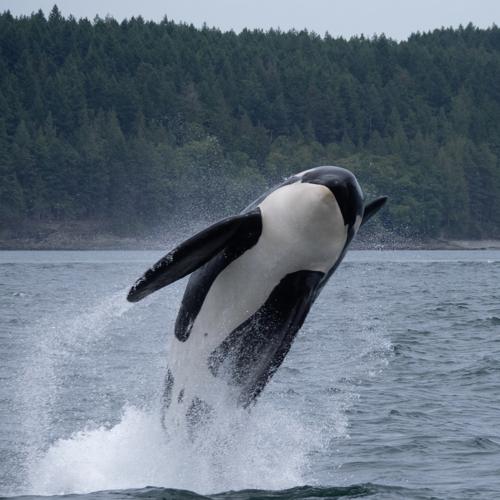}} \hfill \hfill

    \begin{tabularx}{0.24\linewidth}[t]{ZZ}
        \rowcolor{Gray!40} \textbf{MERU} & \textbf{CLIP} \\
        \midrule
        \rowcolor{Apricot!100} \emph{edible agaric} & \emph{edible agaric} \\
        \midrule
        \rowcolor{Apricot!73} \emph{mushroom} & \emph{mushroom} \\
        \midrule
        \rowcolor{Apricot!46} \emph{beauty} & \emph{beauty} \\
        \midrule
        \rowcolor{Apricot!20} \emph{little} & $\downarrow$ \\
        \midrule
        \texttt{\textbf{[ROOT]}} & \texttt{\textbf{[ROOT]}} \\
    \end{tabularx}
    \hfill
    \begin{tabularx}{0.24\linewidth}[t]{ZZ}
        \rowcolor{Gray!40} \textbf{MERU} & \textbf{CLIP} \\
        \midrule
        \rowcolor{Apricot!100} \emph{aquatic animals} & \emph{aquatic animals} \\
        \midrule
        \rowcolor{Apricot!60} \emph{sea life} & \emph{sea life} \\
        \midrule
        \rowcolor{Apricot!20} \emph{style} & \emph{calamity} \\
        \midrule
        \texttt{\textbf{[ROOT]}} & \texttt{\textbf{[ROOT]}} \\
    \end{tabularx}
    \hfill
    \begin{tabularx}{0.24\linewidth}[t]{ZZ}
        \rowcolor{Gray!40} \textbf{MERU} & \textbf{CLIP} \\
        \midrule
        \rowcolor{Apricot!100} \emph{financial} & \emph{adorable} \\
        \midrule
        \rowcolor{Apricot!20} \emph{cute} & $\downarrow$ \\
        \midrule
        \texttt{\textbf{[ROOT]}} & \texttt{\textbf{[ROOT]}} \\
    \end{tabularx}
    \hfill
    \begin{tabularx}{0.24\linewidth}[t]{ZZ}
        \rowcolor{Gray!40} \textbf{MERU} & \textbf{CLIP} \\
        \midrule
        \rowcolor{Apricot!100} \emph{an orca whale jumping out of the water} & \emph{an orca whale jumping out of the water} \\
        \midrule
        \rowcolor{Apricot!20} \emph{whale} & \emph{whale} \\
        \midrule
        \texttt{\textbf{[ROOT]}} & \texttt{\textbf{[ROOT]}} \\
    \end{tabularx}
    \caption{
        \textbf{Image traversals with MERU and CLIP (flora and fauna).}
        Retrieved captions are sourced from \href{https://www.pexels.com}{\texttt{pexels.com}} metadata.
        MERU captures a more systematic and fine-grained visual-semantic hierarchy than CLIP --
        trends are same as \Cref{fig:meru_main_image_traversals}.
      }
      \label{fig:meru_appendix_pexels_flora_fauna}
\end{figure*}

\clearpage

\begin{figure*}[t!]
    \footnotesize
    \newcolumntype{Z}{>{\centering\arraybackslash}X}
    \setlength{\tabcolsep}{1pt}

    \hfill
    \frame{\includegraphics[width=0.18\linewidth]{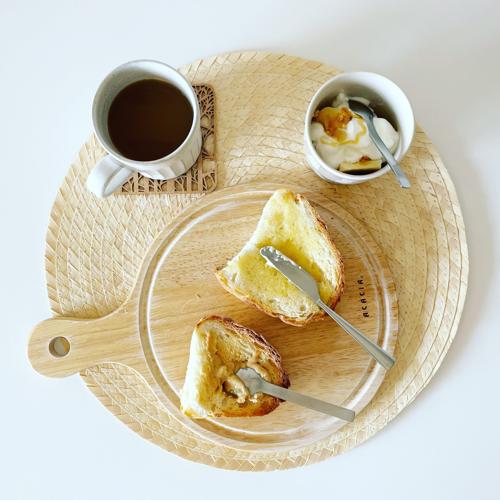}} \hfill \hfill
    \frame{\includegraphics[width=0.18\linewidth]{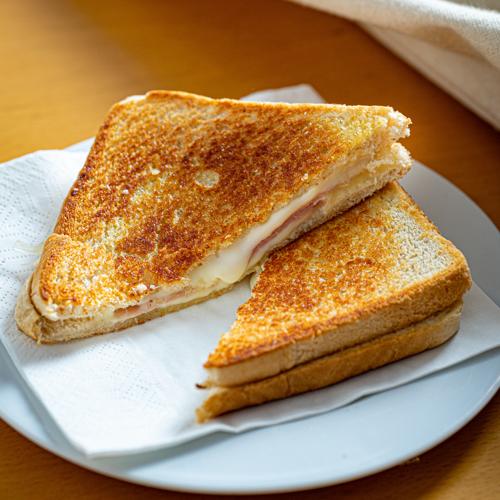}} \hfill \hfill
    \frame{\includegraphics[width=0.18\linewidth]{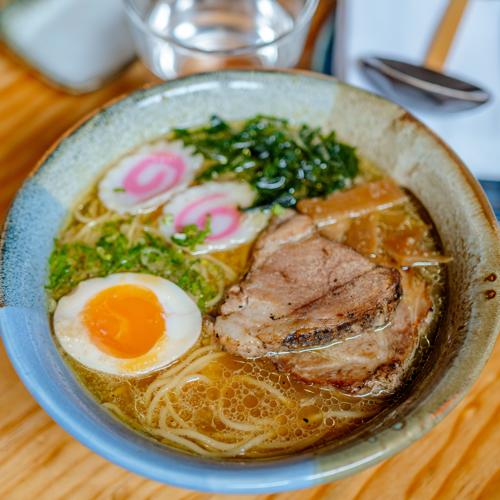}} \hfill \hfill
    \frame{\includegraphics[width=0.18\linewidth]{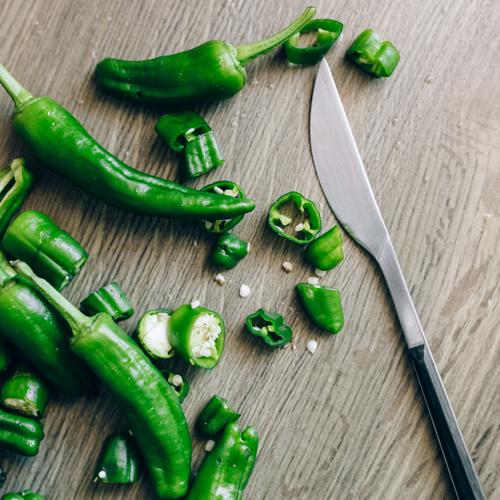}} \hfill \hfill

    \begin{tabularx}{0.24\linewidth}[t]{ZZ}
        \rowcolor{Gray!40} \textbf{MERU} & \textbf{CLIP} \\
        \midrule
        \rowcolor{Apricot!100} \emph{bread and coffee for breakfast} & \emph{bread and coffee for breakfast} \\
        \midrule
        \rowcolor{Apricot!60} \emph{pastry} & $\downarrow$ \\
        \midrule
        \rowcolor{Apricot!20} \emph{art} & $\downarrow$ \\
        \midrule
        \texttt{\textbf{[ROOT]}} & \texttt{\textbf{[ROOT]}} \\
    \end{tabularx}
    \hfill
    \begin{tabularx}{0.24\linewidth}[t]{ZZ}
        \rowcolor{Gray!40} \textbf{MERU} & \textbf{CLIP} \\
        \midrule
        \rowcolor{Apricot!100} \emph{grilled cheese} & \emph{grilled cheese} \\
        \midrule
        \rowcolor{Apricot!73} \emph{lunch} & $\downarrow$ \\
        \midrule
        \rowcolor{Apricot!46} \emph{delicious} & $\downarrow$ \\
        \midrule
        \rowcolor{Apricot!20} \emph{classic} & $\downarrow$ \\
        \midrule
        \texttt{\textbf{[ROOT]}} & \texttt{\textbf{[ROOT]}} \\
    \end{tabularx}
    \hfill
    \begin{tabularx}{0.24\linewidth}[t]{ZZ}
        \rowcolor{Gray!40} \textbf{MERU} & \textbf{CLIP} \\
        \midrule
        \rowcolor{Apricot!100} \emph{bowl of ramen} & \emph{ramen} \\
        \midrule
        \rowcolor{Apricot!60} \emph{local food} & $\downarrow$ \\
        \midrule
        \rowcolor{Apricot!20} \emph{tasty} & $\downarrow$ \\
        \midrule
        \texttt{\textbf{[ROOT]}} & \texttt{\textbf{[ROOT]}} \\
    \end{tabularx}
    \hfill
    \begin{tabularx}{0.24\linewidth}[t]{ZZ}
        \rowcolor{Gray!40} \textbf{MERU} & \textbf{CLIP} \\
        \midrule
        \rowcolor{Apricot!100} \emph{green chili peppers and a knife} & \emph{green chili peppers and a knife} \\
        \midrule
        \rowcolor{Apricot!60} \emph{spicy food} & $\downarrow$ \\
        \midrule
        \rowcolor{Apricot!20} \emph{spicy} & $\downarrow$ \\
        \midrule
        \texttt{\textbf{[ROOT]}} & \texttt{\textbf{[ROOT]}} \\
    \end{tabularx}

    \vspace{5pt}
    \hfill
    \frame{\includegraphics[width=0.18\linewidth]{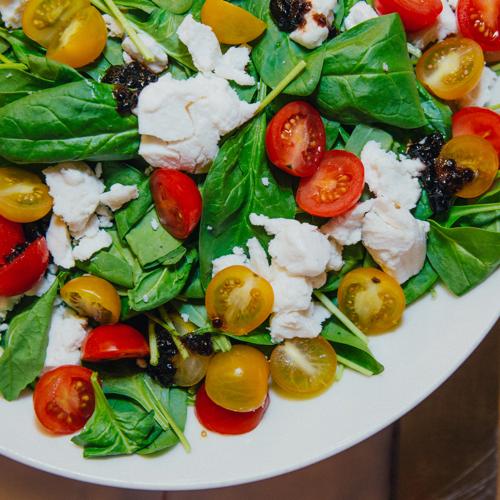}} \hfill \hfill
    \frame{\includegraphics[width=0.18\linewidth]{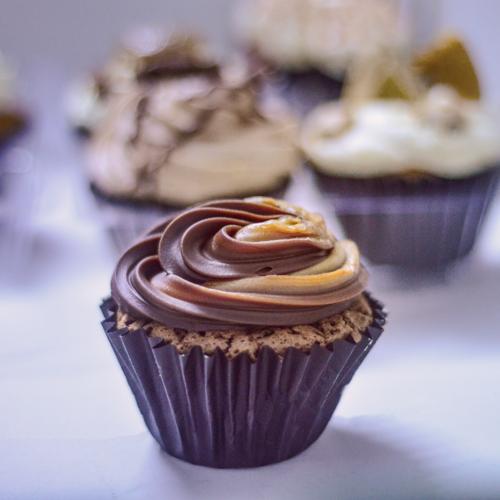}} \hfill \hfill
    \frame{\includegraphics[width=0.18\linewidth]{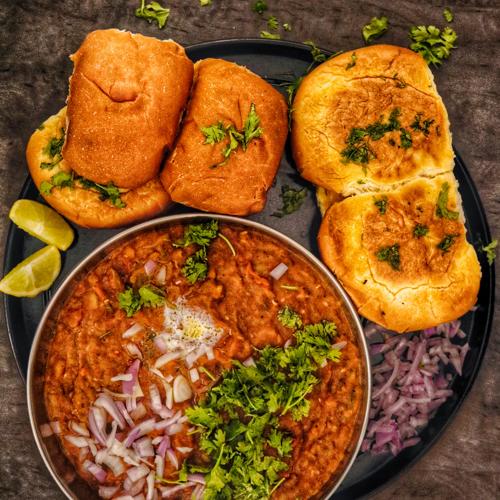}} \hfill \hfill
    \frame{\includegraphics[width=0.18\linewidth]{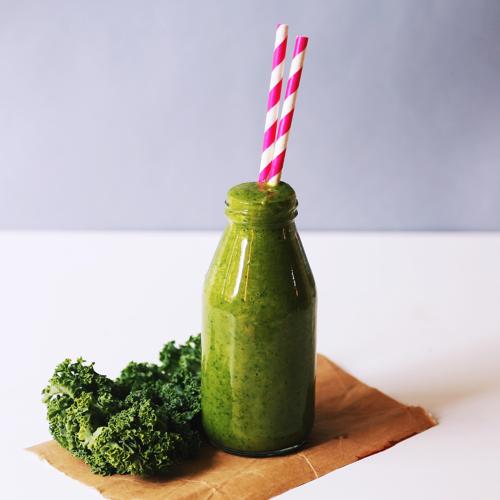}} \hfill \hfill

    \begin{tabularx}{0.24\linewidth}[t]{ZZ}
        \rowcolor{Gray!40} \textbf{MERU} & \textbf{CLIP} \\
        \midrule
        \rowcolor{Apricot!100} \emph{spinach caprese salad} & \emph{spinach caprese salad} \\
        \midrule
        \rowcolor{Apricot!73} \emph{lunch} & \emph{lunch} \\
        \midrule
        \rowcolor{Apricot!46} \emph{homemade} & $\downarrow$ \\
        \midrule
        \rowcolor{Apricot!20} \emph{style} & $\downarrow$ \\
        \midrule
        \texttt{\textbf{[ROOT]}} & \texttt{\textbf{[ROOT]}} \\
    \end{tabularx}
    \hfill
    \begin{tabularx}{0.24\linewidth}[t]{ZZ}
        \rowcolor{Gray!40} \textbf{MERU} & \textbf{CLIP} \\
        \midrule
        \rowcolor{Apricot!100} \emph{cupcakes} & \emph{cupcakes} \\
        \midrule
        \rowcolor{Apricot!84} \emph{chocolate cupcakes} & $\downarrow$ \\
        \midrule
        \rowcolor{Apricot!68} \emph{delicious} & $\downarrow$ \\
        \midrule
        \rowcolor{Apricot!52} \emph{homemade} & $\downarrow$ \\
        \midrule
        \rowcolor{Apricot!36} \emph{clean} & $\downarrow$ \\
        \midrule
        \rowcolor{Apricot!20} \emph{day} & $\downarrow$ \\
        \midrule
        \texttt{\textbf{[ROOT]}} & \texttt{\textbf{[ROOT]}} \\
    \end{tabularx}
    \hfill
    \begin{tabularx}{0.24\linewidth}[t]{ZZ}
        \rowcolor{Gray!40} \textbf{MERU} & \textbf{CLIP} \\
        \midrule
        \rowcolor{Apricot!100} \emph{pav bhaji} & \emph{pav bhaji dish on a bowl} \\
        \midrule
        \rowcolor{Apricot!80} \emph{indian food} & \emph{indian food} \\
        \midrule
        \rowcolor{Apricot!60} \emph{traditional food} & \emph{meal} \\
        \midrule
        \rowcolor{Apricot!40} \emph{local food} & \emph{dinner} \\
        \midrule
        \rowcolor{Apricot!20} \emph{spicy} & $\downarrow$ \\
        \midrule
        \texttt{\textbf{[ROOT]}} & \texttt{\textbf{[ROOT]}} \\
    \end{tabularx}
    \hfill
    \begin{tabularx}{0.24\linewidth}[t]{ZZ}
        \rowcolor{Gray!40} \textbf{MERU} & \textbf{CLIP} \\
        \midrule
        \rowcolor{Apricot!100} \emph{clear glass bottle filled with broccoli shake} & \emph{smoothie} \\
        \midrule
        \rowcolor{Apricot!80} \emph{smoothie} & \emph{homemade} \\
        \midrule
        \rowcolor{Apricot!60} \emph{local food} & \emph{vegetable} \\
        \midrule
        \rowcolor{Apricot!40} \emph{homemade} & $\downarrow$ \\
        \midrule
        \rowcolor{Apricot!20} \emph{spicy} & $\downarrow$ \\
        \midrule
        \texttt{\textbf{[ROOT]}} & \texttt{\textbf{[ROOT]}} \\
    \end{tabularx}

    \vspace{5pt}
    \hfill
    \frame{\includegraphics[width=0.18\linewidth]{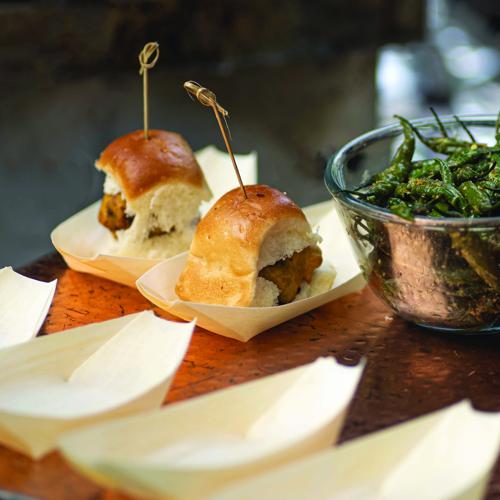}} \hfill \hfill
    \frame{\includegraphics[width=0.18\linewidth]{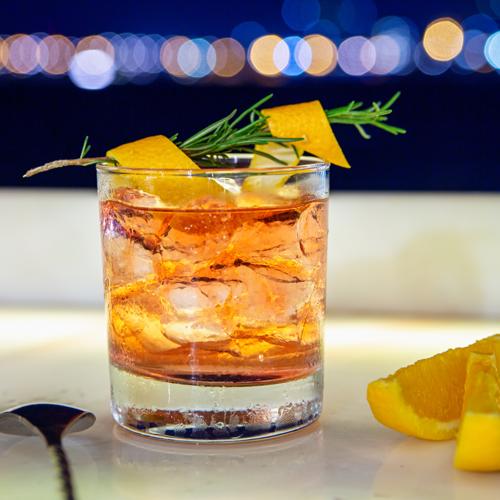}} \hfill \hfill
    \frame{\includegraphics[width=0.18\linewidth]{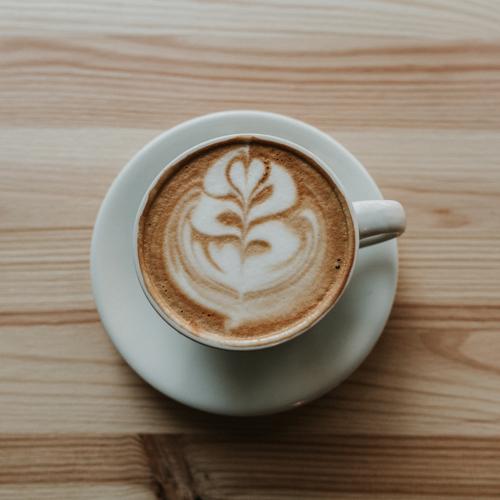}} \hfill \hfill
    \frame{\includegraphics[width=0.18\linewidth]{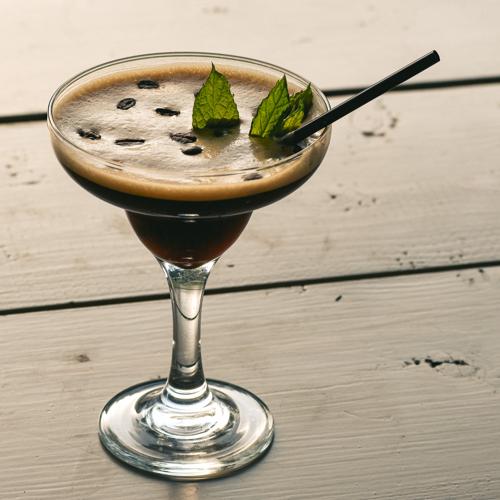}} \hfill \hfill

    \begin{tabularx}{0.24\linewidth}[t]{ZZ}
        \rowcolor{Gray!40} \textbf{MERU} & \textbf{CLIP} \\
        \midrule
        \rowcolor{Apricot!100} \emph{vada pav} & \emph{cheese} \\
        \midrule
        \rowcolor{Apricot!20} \emph{traditional food} & $\downarrow$ \\
        \midrule
        \texttt{\textbf{[ROOT]}} & \texttt{\textbf{[ROOT]}} \\
    \end{tabularx}
    \hfill
    \begin{tabularx}{0.24\linewidth}[t]{ZZ}
        \rowcolor{Gray!40} \textbf{MERU} & \textbf{CLIP} \\
        \midrule
        \rowcolor{Apricot!100} \emph{old fashioned} & \emph{nutrition} \\
        \midrule
        \rowcolor{Apricot!60} \emph{spicy} & $\downarrow$ \\
        \midrule
        \rowcolor{Apricot!20} \emph{style} & $\downarrow$ \\
        \midrule
        \texttt{\textbf{[ROOT]}} & \texttt{\textbf{[ROOT]}} \\
    \end{tabularx}
    \hfill
    \begin{tabularx}{0.24\linewidth}[t]{ZZ}
        \rowcolor{Gray!40} \textbf{MERU} & \textbf{CLIP} \\
        \midrule
        \rowcolor{Apricot!100} \emph{latte} & \emph{latte} \\
        \midrule
        \rowcolor{Apricot!60} \emph{design} & $\downarrow$ \\
        \midrule
        \rowcolor{Apricot!20} \emph{style} & $\downarrow$ \\
        \midrule
        \texttt{\textbf{[ROOT]}} & \texttt{\textbf{[ROOT]}} \\
    \end{tabularx}
    \hfill
    \begin{tabularx}{0.24\linewidth}[t]{ZZ}
        \rowcolor{Gray!40} \textbf{MERU} & \textbf{CLIP} \\
        \midrule
        \rowcolor{Apricot!100} \emph{espresso martini} & $\downarrow$ \\
        \midrule
        \rowcolor{Apricot!73} \emph{cocktail} & $\downarrow$ \\
        \midrule
        \rowcolor{Apricot!46} \emph{dessert} & $\downarrow$ \\
        \midrule
        \rowcolor{Apricot!20} \emph{hot} & $\downarrow$ \\
        \midrule
        \texttt{\textbf{[ROOT]}} & \texttt{\textbf{[ROOT]}} \\
    \end{tabularx}
    \caption{
        \textbf{Image traversals with MERU and CLIP (food and drinks).}
        Retrieved captions are sourced from \href{https://www.pexels.com}{\texttt{pexels.com}} metadata.
        MERU captures a more systematic and fine-grained visual-semantic hierarchy than CLIP --
        trends are same as \Cref{fig:meru_main_image_traversals}.
      }
      \label{fig:meru_appendix_pexels_food}
\end{figure*}

\clearpage

\begin{figure*}[t!]
    \footnotesize
    \newcolumntype{Z}{>{\centering\arraybackslash}X}
    \setlength{\tabcolsep}{1pt}

    \hfill
    \frame{\includegraphics[width=0.18\linewidth]{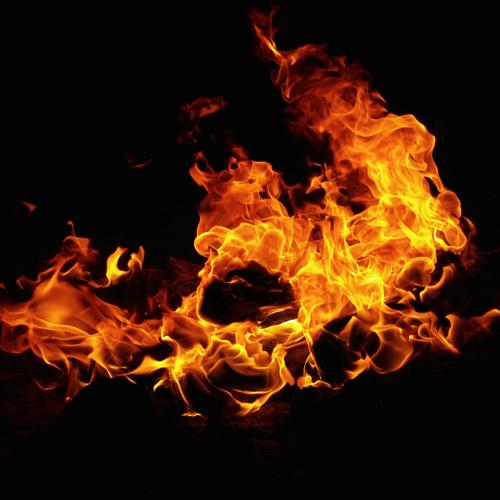}} \hfill \hfill
    \frame{\includegraphics[width=0.18\linewidth]{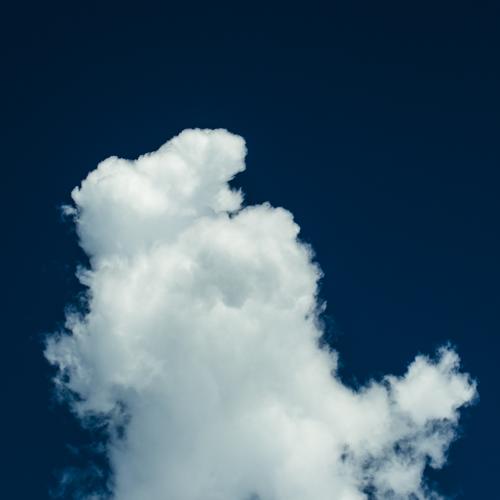}} \hfill \hfill
    \frame{\includegraphics[width=0.18\linewidth]{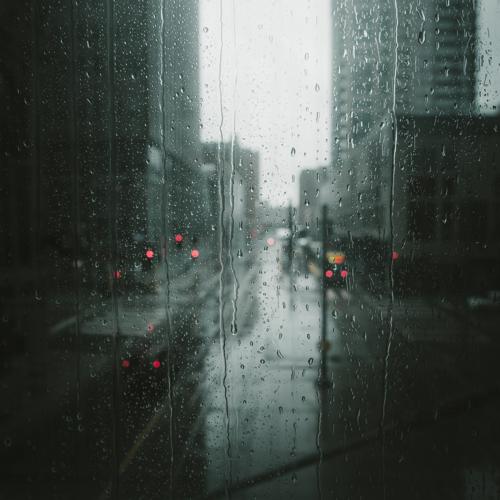}} \hfill \hfill
    \frame{\includegraphics[width=0.18\linewidth]{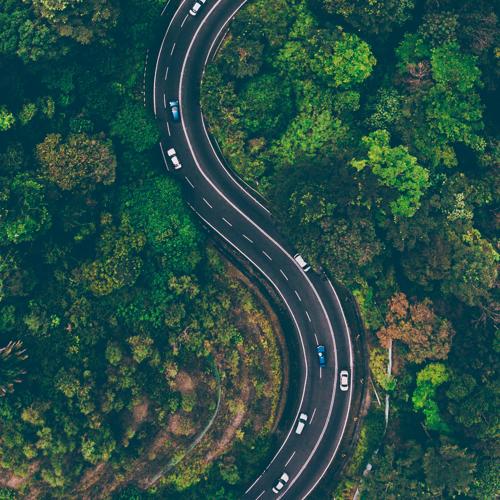}} \hfill \hfill

    \begin{tabularx}{0.24\linewidth}[t]{ZZ}
        \rowcolor{Gray!40} \textbf{MERU} & \textbf{CLIP} \\
        \midrule
        \rowcolor{Apricot!100} \emph{campfire} & \emph{inferno} \\
        \midrule
        \rowcolor{Apricot!73} \emph{fire} & $\downarrow$ \\
        \midrule
        \rowcolor{Apricot!46} \emph{blaze} & $\downarrow$ \\
        \midrule
        \rowcolor{Apricot!20} \emph{hot} & $\downarrow$ \\
        \midrule
        \texttt{\textbf{[ROOT]}} & \texttt{\textbf{[ROOT]}} \\
    \end{tabularx}
    \hfill
    \begin{tabularx}{0.24\linewidth}[t]{ZZ}
        \rowcolor{Gray!40} \textbf{MERU} & \textbf{CLIP} \\
        \midrule
        \rowcolor{Apricot!100} \emph{cumulus} & \emph{cumulus} \\
        \midrule
        \rowcolor{Apricot!80} \emph{white clouds} & $\downarrow$ \\
        \midrule
        \rowcolor{Apricot!60} \emph{clouds} & $\downarrow$ \\
        \midrule
        \rowcolor{Apricot!40} \emph{health} & $\downarrow$ \\
        \midrule
        \rowcolor{Apricot!20} \emph{fluffy} & $\downarrow$ \\
        \midrule
        \texttt{\textbf{[ROOT]}} & \texttt{\textbf{[ROOT]}} \\
    \end{tabularx}
    \hfill
    \begin{tabularx}{0.24\linewidth}[t]{ZZ}
        \rowcolor{Gray!40} \textbf{MERU} & \textbf{CLIP} \\
        \midrule
        \rowcolor{Apricot!100} \emph{raining in the city} & \emph{raining in the city} \\
        \midrule
        \rowcolor{Apricot!73} \emph{weather} & \emph{downtown} \\
        \midrule
        \rowcolor{Apricot!46} \emph{simple} & $\downarrow$ \\
        \midrule
        \rowcolor{Apricot!20} \emph{day} & $\downarrow$ \\
        \midrule
        \texttt{\textbf{[ROOT]}} & \texttt{\textbf{[ROOT]}} \\
    \end{tabularx}
    \hfill
    \begin{tabularx}{0.24\linewidth}[t]{ZZ}
        \rowcolor{Gray!40} \textbf{MERU} & \textbf{CLIP} \\
        \midrule
        \rowcolor{Apricot!100} \emph{road} & \emph{aerial view of road in the middle of trees} \\
        \midrule
        \rowcolor{Apricot!73} \emph{travel} & \emph{aerial shot} \\
        \midrule
        \rowcolor{Apricot!46} \emph{style} & \emph{rural} \\
        \midrule
        \rowcolor{Apricot!20} $\downarrow$ & \emph{clean} \\
        \midrule
        \texttt{\textbf{[ROOT]}} & \texttt{\textbf{[ROOT]}} \\
    \end{tabularx}

    \vspace{5pt}
    \hfill
    \frame{\includegraphics[width=0.18\linewidth]{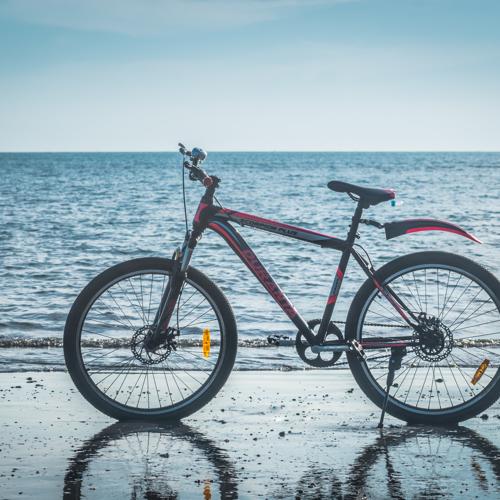}} \hfill \hfill
    \frame{\includegraphics[width=0.18\linewidth]{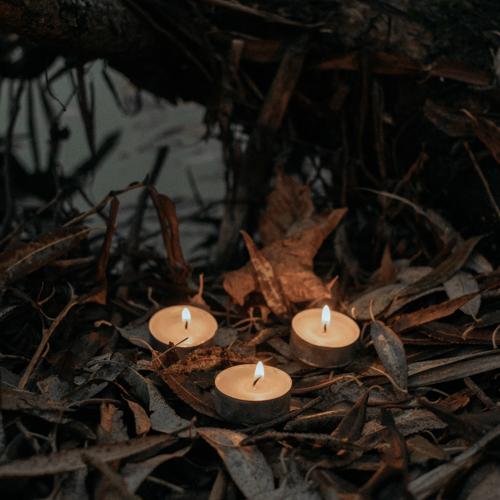}} \hfill \hfill
    \frame{\includegraphics[width=0.18\linewidth]{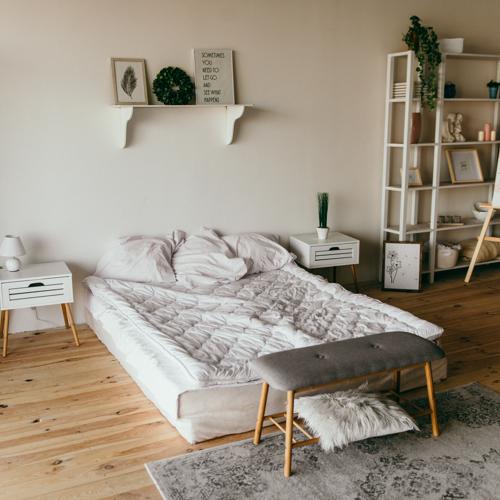}} \hfill \hfill
    \frame{\includegraphics[width=0.18\linewidth]{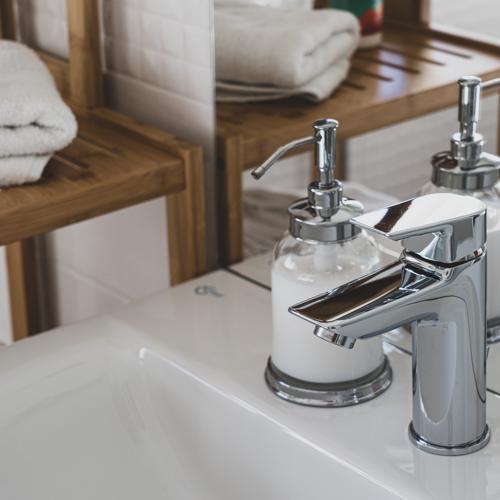}} \hfill \hfill

    \begin{tabularx}{0.24\linewidth}[t]{ZZ}
        \rowcolor{Gray!40} \textbf{MERU} & \textbf{CLIP} \\
        \midrule
        \rowcolor{Apricot!100} \emph{mountain bike on the beach} & \emph{mountain bike on the beach} \\
        \midrule
        \rowcolor{Apricot!73} \emph{analog} & \emph{bicycle} \\
        \midrule
        \rowcolor{Apricot!46} \emph{retro} & $\downarrow$ \\
        \midrule
        \rowcolor{Apricot!20} \emph{style} & $\downarrow$ \\
        \midrule
        \texttt{\textbf{[ROOT]}} & \texttt{\textbf{[ROOT]}} \\
    \end{tabularx}
    \hfill
    \begin{tabularx}{0.24\linewidth}[t]{ZZ}
        \rowcolor{Gray!40} \textbf{MERU} & \textbf{CLIP} \\
        \midrule
        \rowcolor{Apricot!100} \emph{lights} & \emph{white heart shaped candle on dried leaves} \\
        \midrule
        \rowcolor{Apricot!60} \emph{evening} & \emph{holiday} \\
        \midrule
        \rowcolor{Apricot!20} \emph{day} & $\downarrow$ \\
        \midrule
        \texttt{\textbf{[ROOT]}} & \texttt{\textbf{[ROOT]}} \\
    \end{tabularx}
    \hfill
    \begin{tabularx}{0.24\linewidth}[t]{ZZ}
        \rowcolor{Gray!40} \textbf{MERU} & \textbf{CLIP} \\
        \midrule
        \rowcolor{Apricot!100} \emph{bedroom} & $\downarrow$ \\
        \midrule
        \rowcolor{Apricot!20} \emph{clean} & $\downarrow$ \\
        \midrule
        \texttt{\textbf{[ROOT]}} & \texttt{\textbf{[ROOT]}} \\
    \end{tabularx}
    \hfill
    \begin{tabularx}{0.24\linewidth}[t]{ZZ}
        \rowcolor{Gray!40} \textbf{MERU} & \textbf{CLIP} \\
        \midrule
        \rowcolor{Apricot!100} \emph{clean bathroom} & \emph{stainless steel faucet on white ceramic sink} \\
        \midrule
        \rowcolor{Apricot!60} \emph{investment} & $\downarrow$ \\
        \midrule
        \rowcolor{Apricot!20} \emph{clean} & $\downarrow$ \\
        \midrule
        \texttt{\textbf{[ROOT]}} & \texttt{\textbf{[ROOT]}} \\
    \end{tabularx}

    \vspace{5pt}
    \hfill
    \frame{\includegraphics[width=0.18\linewidth]{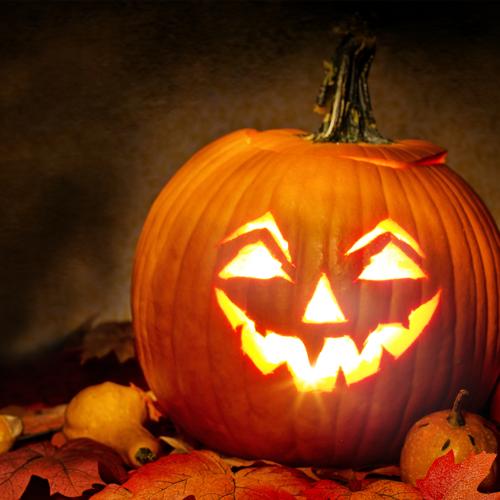}} \hfill \hfill
    \frame{\includegraphics[width=0.18\linewidth]{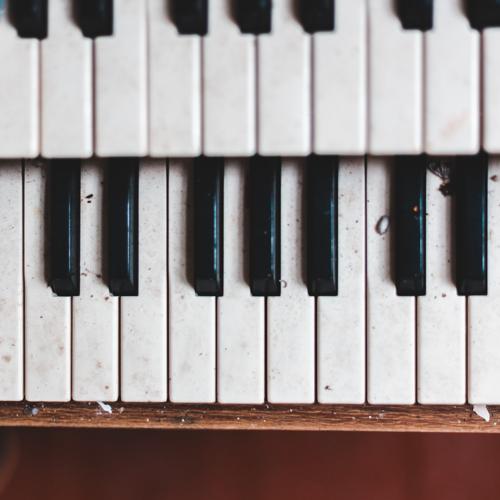}} \hfill \hfill
    \frame{\includegraphics[width=0.18\linewidth]{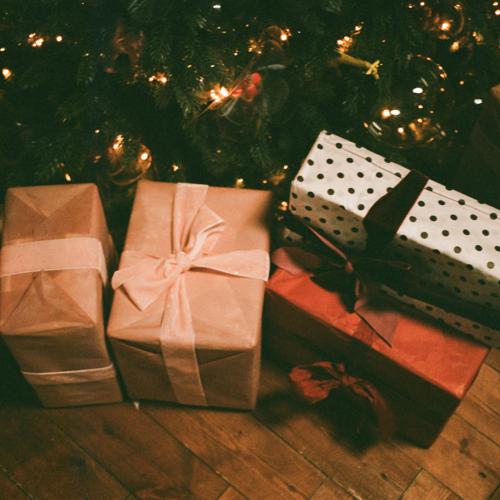}} \hfill \hfill
    \frame{\includegraphics[width=0.18\linewidth]{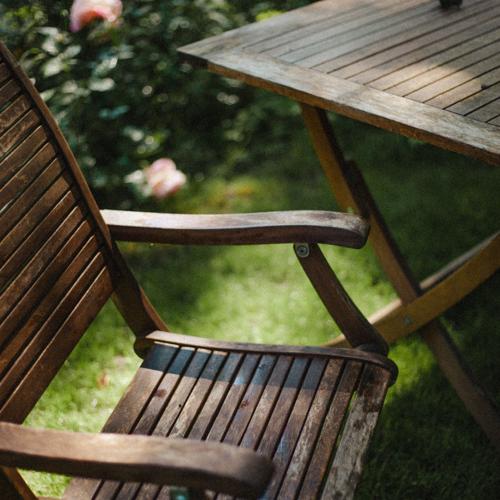}} \hfill \hfill

    \begin{tabularx}{0.24\linewidth}[t]{ZZ}
        \rowcolor{Gray!40} \textbf{MERU} & \textbf{CLIP} \\
        \midrule
        \rowcolor{Apricot!100} \emph{jack o lantern with light} & \emph{jack o lantern with light} \\
        \midrule
        \rowcolor{Apricot!73} \emph{carved pumpkin} & $\downarrow$ \\
        \midrule
        \rowcolor{Apricot!46} \emph{halloween} & $\downarrow$ \\
        \midrule
        \rowcolor{Apricot!20} \emph{hot} & $\downarrow$ \\
        \midrule
        \texttt{\textbf{[ROOT]}} & \texttt{\textbf{[ROOT]}} \\
    \end{tabularx}
    \hfill
    \begin{tabularx}{0.24\linewidth}[t]{ZZ}
        \rowcolor{Gray!40} \textbf{MERU} & \textbf{CLIP} \\
        \midrule
        \rowcolor{Apricot!100} \emph{piano keys} & \emph{musical instrument} \\
        \midrule
        \rowcolor{Apricot!80} \emph{keyboard} & \emph{music} \\
        \midrule
        \rowcolor{Apricot!60} \emph{analog} & $\downarrow$ \\
        \midrule
        \rowcolor{Apricot!40} \emph{vintage} & $\downarrow$ \\
        \midrule
        \rowcolor{Apricot!20} \emph{style} & $\downarrow$ \\
        \midrule
        \texttt{\textbf{[ROOT]}} & \texttt{\textbf{[ROOT]}} \\
    \end{tabularx}
    \hfill
    \begin{tabularx}{0.24\linewidth}[t]{ZZ}
        \rowcolor{Gray!40} \textbf{MERU} & \textbf{CLIP} \\
        \midrule
        \rowcolor{Apricot!100} \emph{assorted gift boxes on floor near christmas tree} & \emph{christmas presents} \\
        \midrule
        \rowcolor{Apricot!60} \emph{christmas presents} & \emph{christmas gifts} \\
        \midrule
        \rowcolor{Apricot!20} \emph{christmas gifts} & $\downarrow$ \\
        \midrule
        \texttt{\textbf{[ROOT]}} & \texttt{\textbf{[ROOT]}} \\
    \end{tabularx}
    \hfill
    \begin{tabularx}{0.24\linewidth}[t]{ZZ}
        \rowcolor{Gray!40} \textbf{MERU} & \textbf{CLIP} \\
        \midrule
        \rowcolor{Apricot!100} \emph{garden table and chair} & \emph{seat} \\
        \midrule
        \rowcolor{Apricot!73} \emph{table} & $\downarrow$ \\
        \midrule
        \rowcolor{Apricot!46} \emph{design} & $\downarrow$ \\
        \midrule
        \rowcolor{Apricot!20} \emph{comfort} & $\downarrow$ \\
        \midrule
        \texttt{\textbf{[ROOT]}} & \texttt{\textbf{[ROOT]}} \\
    \end{tabularx}
    \caption{
        \textbf{Image traversals with MERU and CLIP (objects and scenes).}
        Retrieved captions are sourced from \href{https://www.pexels.com}{\texttt{pexels.com}} metadata.
        MERU captures a more systematic and fine-grained visual-semantic hierarchy than CLIP --
        trends are same as \Cref{fig:meru_main_image_traversals}.
      }
      \label{fig:meru_appendix_pexels_objects1}
\end{figure*}

\clearpage

\begin{figure*}[t!]
    \footnotesize
    \newcolumntype{Z}{>{\centering\arraybackslash}X}
    \setlength{\tabcolsep}{1pt}

    \hfill
    \frame{\includegraphics[width=0.18\linewidth]{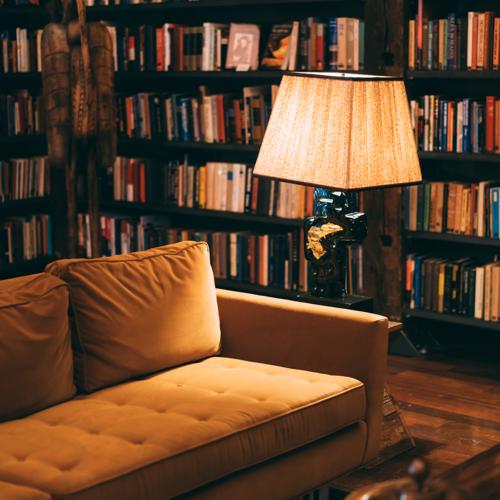}} \hfill \hfill
    \frame{\includegraphics[width=0.18\linewidth]{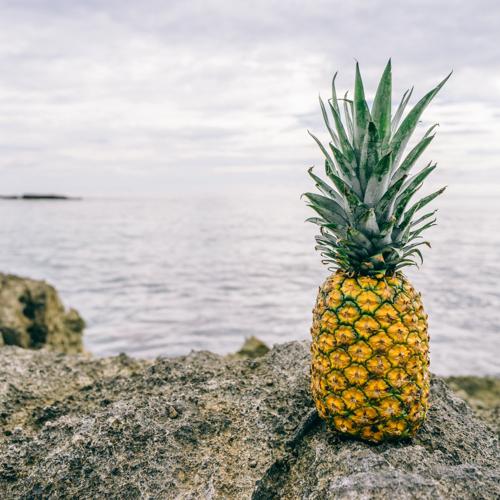}} \hfill \hfill
    \frame{\includegraphics[width=0.18\linewidth]{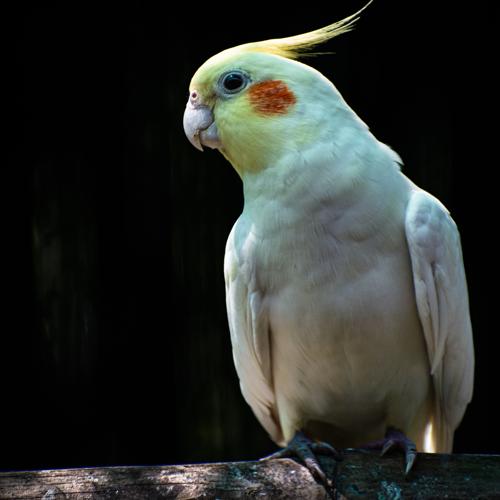}} \hfill \hfill
    \frame{\includegraphics[width=0.18\linewidth]{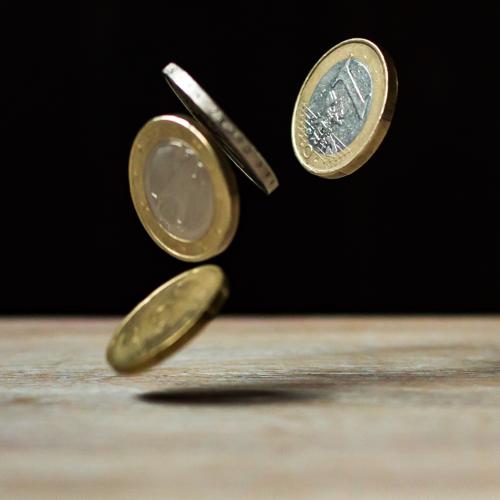}} \hfill \hfill

    \begin{tabularx}{0.24\linewidth}[t]{ZZ}
        \rowcolor{Gray!40} \textbf{MERU} & \textbf{CLIP} \\
        \midrule
        \rowcolor{Apricot!100} \emph{turned on floor lamp near sofa on a library room} & \emph{bookshelves} \\
        \midrule
        \rowcolor{Apricot!86} \emph{books} & $\downarrow$ \\
        \midrule
        \rowcolor{Apricot!73} \emph{bookshelves} & $\downarrow$ \\
        \midrule
        \rowcolor{Apricot!60} \emph{comfort room} & $\downarrow$ \\
        \midrule
        \rowcolor{Apricot!46} \emph{cozy} & $\downarrow$ \\
        \midrule
        \rowcolor{Apricot!20} \emph{style} & $\downarrow$ \\
        \midrule
        \texttt{\textbf{[ROOT]}} & \texttt{\textbf{[ROOT]}} \\
    \end{tabularx}
    \hfill
    \begin{tabularx}{0.24\linewidth}[t]{ZZ}
        \rowcolor{Gray!40} \textbf{MERU} & \textbf{CLIP} \\
        \midrule
        \rowcolor{Apricot!100} \emph{pineapple} & \emph{ripe pineapple on gray rock beside body of water} \\
        \midrule
        \rowcolor{Apricot!73} \emph{inspiration} & \emph{pineapple} \\
        \midrule
        \rowcolor{Apricot!46} \emph{health} & \emph{calamity} \\
        \midrule
        \rowcolor{Apricot!20} \emph{little} & $\downarrow$ \\
        \midrule
        \texttt{\textbf{[ROOT]}} & \texttt{\textbf{[ROOT]}} \\
    \end{tabularx}
    \hfill
    \begin{tabularx}{0.24\linewidth}[t]{ZZ}
        \rowcolor{Gray!40} \textbf{MERU} & \textbf{CLIP} \\
        \midrule
        \rowcolor{Apricot!100} \emph{cockatiel} & \emph{cockatiel} \\
        \midrule
        \rowcolor{Apricot!20} \emph{female} & $\downarrow$ \\
        \midrule
        \texttt{\textbf{[ROOT]}} & \texttt{\textbf{[ROOT]}} \\
    \end{tabularx}
    \hfill
    \begin{tabularx}{0.24\linewidth}[t]{ZZ}
        \rowcolor{Gray!40} \textbf{MERU} & \textbf{CLIP} \\
        \midrule
        \rowcolor{Apricot!100} \emph{currency} & \emph{euro} \\
        \midrule
        \rowcolor{Apricot!20} \emph{simple} & \emph{revenue} \\
        \midrule
        \texttt{\textbf{[ROOT]}} & \texttt{\textbf{[ROOT]}} \\
    \end{tabularx}
    \caption{
        \textbf{Image traversals (objects and scenes).}
        Retrieved captions are sourced from \href{https://www.pexels.com}{\texttt{pexels.com}} metadata.
        MERU captures a more systematic and fine-grained visual-semantic hierarchy than CLIP --
        trends are same as \Cref{fig:meru_main_image_traversals}.
    }
    \label{fig:meru_appendix_pexels_objects2}
\end{figure*}

%% file: figtabs/appendix_traversals_yfcc.tex
\begin{figure*}[h!]
    \footnotesize
    \newcolumntype{Z}{>{\centering\arraybackslash}X}
    \setlength{\tabcolsep}{2pt}

    \vspace{5pt}
    \adjustbox{valign=t}{
        \begin{minipage}{0.18\linewidth}
            \centering
            \frame{\includegraphics[width=\linewidth]{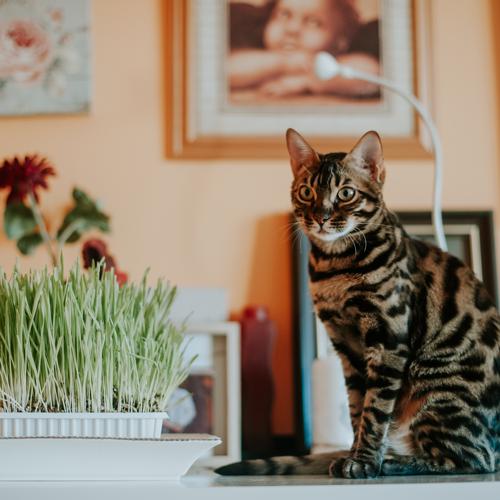}}
            (1)
        \end{minipage}
    }
    \hfill
    \begin{tabularx}{0.8\linewidth}[t]{ZZ}
        \rowcolor{Gray!40} \textbf{MERU} & \textbf{CLIP} \\
        \midrule
        \rowcolor{Apricot!120} \emph{leopard and stig have a beautiful piano at their home.} & \emph{loki is a 1 year old bengal cat.} \\
        \midrule
        \rowcolor{Apricot!103} \emph{merlin wasn't impressed to leave the last house and his precious cat grass} & $\downarrow$ \\
        \midrule
        \rowcolor{Apricot!86} \emph{my parents cat 'barry' loves being photographed!} & $\downarrow$ \\
        \midrule
        \rowcolor{Apricot!70} \emph{house cat posing} & $\downarrow$ \\
        \midrule
        \rowcolor{Apricot!53} \emph{mr . bo-majed} & $\downarrow$ \\
        \midrule
        \rowcolor{Apricot!36} \emph{our cat, our love our third member of our family. :)} & $\downarrow$ \\
        \midrule
        \rowcolor{Apricot!20} \emph{why are you taking pictures? it's dilo don't fill it up. :)} & $\downarrow$ \\
        \midrule
        \texttt{\textbf{[ROOT]}} & \texttt{\textbf{[ROOT]}} \\
    \end{tabularx}

    \vspace{5pt}
    \adjustbox{valign=t}{
        \begin{minipage}{0.18\linewidth}
            \centering
            \frame{\includegraphics[width=\linewidth]{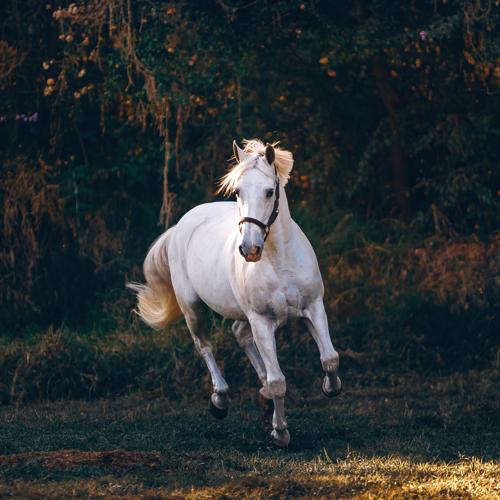}}
            (2)
        \end{minipage}
    }
    \hfill
    \begin{tabularx}{0.8\linewidth}[t]{ZZ}
        \rowcolor{Gray!40} \textbf{MERU} & \textbf{CLIP} \\
        \midrule
        \rowcolor{Apricot!120} \emph{caught my attention by the beautiful light cascading on a grass behind this fellow.} & \emph{pity about the camera shake in the evening light} \\
        \midrule
        \rowcolor{Apricot!107} \emph{the focus is all wrong , but the white on the tail and the tongue are pretty cool.} & $\downarrow$ \\
        \midrule
        \rowcolor{Apricot!95} \emph{just a goofy white guy.} & $\downarrow$ \\
        \midrule
        \rowcolor{Apricot!82} \emph{he was an active one, running to and fro.} & $\downarrow$ \\
        \midrule
        \rowcolor{Apricot!70} \emph{but then, she was happy to pose for me} & $\downarrow$ \\
        \midrule
        \rowcolor{Apricot!57} \emph{if she were a race horse her name would be poopbiscuit.} & $\downarrow$ \\
        \midrule
        \rowcolor{Apricot!45} \emph{dorky photo is dorky} & $\downarrow$ \\
        \midrule
        \rowcolor{Apricot!32} \emph{she looks so leery of the camera in this photo.} & $\downarrow$ \\
        \midrule
        \rowcolor{Apricot!20} \emph{this is only luky.} & $\downarrow$ \\
        \midrule
        \texttt{\textbf{[ROOT]}} & \texttt{\textbf{[ROOT]}} \\
    \end{tabularx}

    \vspace{5pt}
    \adjustbox{valign=t}{
        \begin{minipage}{0.18\linewidth}
            \centering
            \frame{\includegraphics[width=\linewidth]{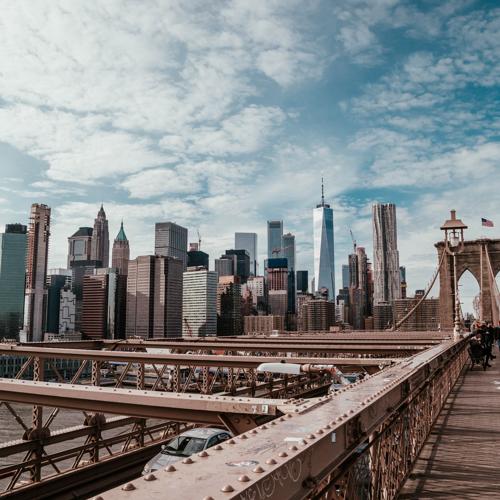}}
            (3)
        \end{minipage}
    }
    \hfill
    \begin{tabularx}{0.8\linewidth}[t]{ZZ}
        \rowcolor{Gray!40} \textbf{MERU} & \textbf{CLIP} \\
        \midrule
        \rowcolor{Apricot!120} \emph{going across brooklyn bridge on the way to brooklyn 3 likes on instagram} & \emph{shot from the manhattan end of the brooklyn bridge} \\
        \midrule
        \rowcolor{Apricot!105} \emph{manhattan depuis le brooklyn bridge park, a brooklyn.} & \emph{much more scenic to walk on than the brooklyn bridge} \\
        \midrule
        \rowcolor{Apricot!91} \emph{bridge, manhattan skyline} & \emph{new york new york!} \\
        \midrule
        \rowcolor{Apricot!77} \emph{shot from near the middle of the brooklyn bridge.} & $\downarrow$ \\
        \midrule
        \rowcolor{Apricot!62} \emph{this city goes on forever} & $\downarrow$ \\
        \midrule
        \rowcolor{Apricot!48} \emph{the city that never sleeps} & $\downarrow$ \\
        \midrule
        \rowcolor{Apricot!34} \emph{the city that never sleeps...it can't.} & $\downarrow$ \\
        \midrule
        \rowcolor{Apricot!20} \emph{it can be seen from most places in the city} & $\downarrow$ \\
        \midrule
        \texttt{\textbf{[ROOT]}} & \texttt{\textbf{[ROOT]}} \\
    \end{tabularx}

    \vspace{5pt}
    \adjustbox{valign=t}{
        \begin{minipage}{0.18\linewidth}
            \centering
            \frame{\includegraphics[width=\linewidth]{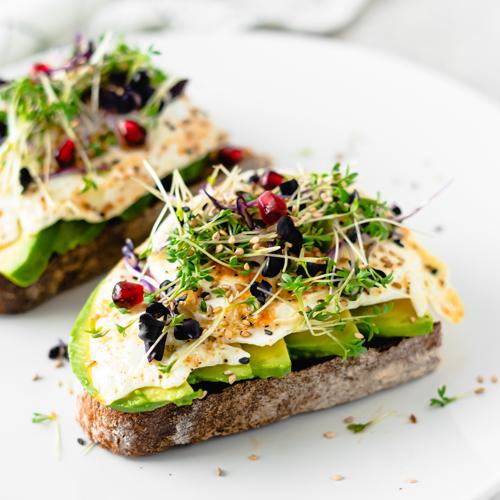}}
            (4)
        \end{minipage}
    }
    \hfill
    \begin{tabularx}{0.8\linewidth}[t]{ZZ}
        \rowcolor{Gray!40} \textbf{MERU} & \textbf{CLIP} \\
        \midrule
        \rowcolor{Apricot!120} \emph{avocado, roasted garlic and sriracha on light rye \& raisins sourdough.} & \emph{la tartine} \\
        \midrule
        \rowcolor{Apricot!95} \emph{with avocado slices - yum!} & \emph{a toast to the new place} \\
        \midrule
        \rowcolor{Apricot!70} \emph{vaguely-healthy} & $\downarrow$ \\
        \midrule
        \rowcolor{Apricot!45} \emph{on bread, probably not what you should do with it, but it was a good meal} & $\downarrow$ \\
        \midrule
        \rowcolor{Apricot!20} \emph{nice if you like this sort of thing.} & $\downarrow$ \\
        \midrule
        \texttt{\textbf{[ROOT]}} & \texttt{\textbf{[ROOT]}} \\
    \end{tabularx}
\end{figure*}

\clearpage

\begin{figure*}[h]
    \footnotesize
    \newcolumntype{Z}{>{\centering\arraybackslash}X}
    \setlength{\tabcolsep}{2pt}

    \vspace{5pt}
    \adjustbox{valign=t}{
        \begin{minipage}{0.18\linewidth}
            \centering
            \frame{\includegraphics[width=\linewidth]{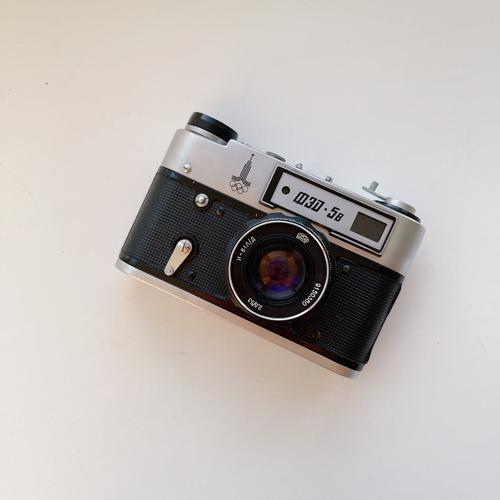}}
            (5)
        \end{minipage}
    }
    \hfill
    \begin{tabularx}{0.8\linewidth}[t]{ZZ}
        \rowcolor{Gray!40} \textbf{MERU} & \textbf{CLIP} \\
        \midrule
        \rowcolor{Apricot!120} \emph{avenida paulista. fisheye 2 kodak elite chrome 100} & \emph{leica m7 with voigtlander zoom finder and dsptch camera strap} \\
        \midrule
        \rowcolor{Apricot!95} \emph{rolleiflex kodak portra 160 epson v500 scanner} & \emph{rollei minidigi} \\
        \midrule
        \rowcolor{Apricot!70} \emph{...of my brand new shiny 7.1mp camera.} & \emph{zeiss ikon icarex, 02/2010} \\
        \midrule
        \rowcolor{Apricot!45} \emph{camera de 5mp} & \emph{faux lomo from www.dumpr.net - photo fun} \\
        \midrule
        \rowcolor{Apricot!20} \emph{i'm a lumix camera fan now.} & \emph{zeiss-ikon} \\
        \midrule
        \texttt{\textbf{[ROOT]}} & \texttt{\textbf{[ROOT]}} \\
    \end{tabularx}

    \vspace{5pt}
    \adjustbox{valign=t}{
        \begin{minipage}{0.18\linewidth}
            \centering
            \frame{\includegraphics[width=\linewidth]{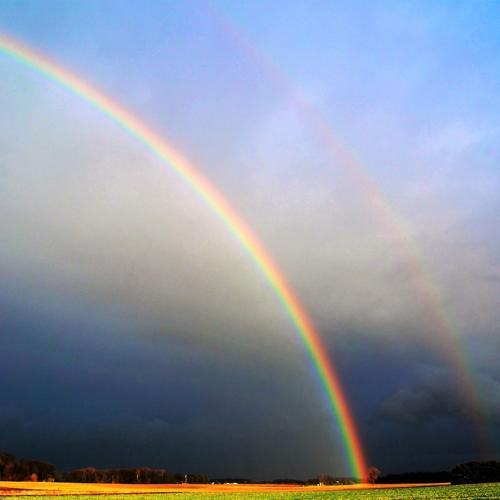}}
            (6)
        \end{minipage}
    }
    \hfill
    \begin{tabularx}{0.8\linewidth}[t]{ZZ}
        \rowcolor{Gray!40} \textbf{MERU} & \textbf{CLIP} \\
        \midrule
        \rowcolor{Apricot!120} \emph{double rainbows in our field were too good to pass up photographing} & \emph{double rainbows in our field were too good to pass up photographing} \\
        \midrule
        \rowcolor{Apricot!103} \emph{whoa... double rainbow} & \emph{is that... a double rainbow? ;-)} \\
        \midrule
        \rowcolor{Apricot!86} \emph{is that... a double rainbow? ;-)} & \emph{what does it mean!} \\
        \midrule
        \rowcolor{Apricot!70} \emph{what does it mean of this picture?} & $\downarrow$ \\
        \midrule
        \rowcolor{Apricot!53} \emph{only god could create something so beautiful.} & $\downarrow$ \\
        \midrule
        \rowcolor{Apricot!36} \emph{this is a good one to end with. reminds me of the woman in this picture.} & $\downarrow$ \\
        \midrule
        \rowcolor{Apricot!20} \emph{look out for that right one.} & $\downarrow$ \\
        \midrule
        \texttt{\textbf{[ROOT]}} & \texttt{\textbf{[ROOT]}} \\
    \end{tabularx}

    \vspace{5pt}
    \adjustbox{valign=t}{
        \begin{minipage}{0.18\linewidth}
            \centering
            \frame{\includegraphics[width=\linewidth]{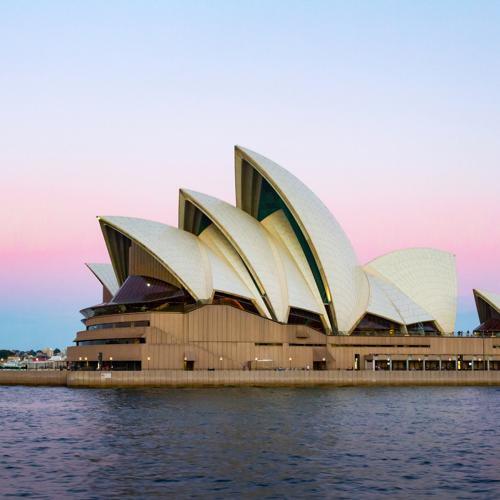}}
            (7)
        \end{minipage}
    }
    \hfill
    \begin{tabularx}{0.8\linewidth}[t]{ZZ}
        \rowcolor{Gray!40} \textbf{MERU} & \textbf{CLIP} \\
        \midrule
        \rowcolor{Apricot!120} \emph{sydney opera house, october 2012.} & \emph{gros plan sur l'opera de sydney} \\
        \midrule
        \rowcolor{Apricot!105} \emph{sydney opera house see where this picture was taken.} & \emph{you can just make out the opera house in the far left.} \\
        \midrule
        \rowcolor{Apricot!91} \emph{from the new opera house} & \emph{my sydney} \\
        \midrule
        \rowcolor{Apricot!77} \emph{i think this is the last one i have of the opera house.} & $\downarrow$ \\
        \midrule
        \rowcolor{Apricot!62} \emph{oh, and some opera house, too.} & $\downarrow$ \\
        \midrule
        \rowcolor{Apricot!48} \emph{just next to the famous opera house} & $\downarrow$ \\
        \midrule
        \rowcolor{Apricot!34} \emph{from horseshoe bay.} & $\downarrow$ \\
        \midrule
        \rowcolor{Apricot!20} \emph{taken from the donau.} & $\downarrow$ \\
        \midrule
        \texttt{\textbf{[ROOT]}} & \texttt{\textbf{[ROOT]}} \\
    \end{tabularx}

    \vspace{5pt}
    \adjustbox{valign=t}{
        \begin{minipage}{0.18\linewidth}
            \centering
            \frame{\includegraphics[width=\linewidth]{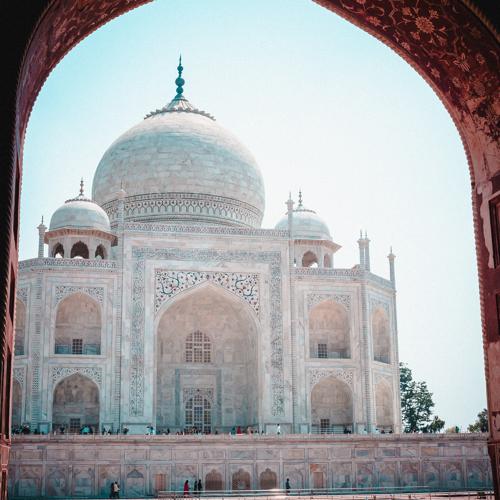}}
            (8)
        \end{minipage}
    }
    \hfill
    \begin{tabularx}{0.8\linewidth}[t]{ZZ}
        \rowcolor{Gray!40} \textbf{MERU} & \textbf{CLIP} \\
        \midrule
        \rowcolor{Apricot!120} \emph{captured this during my visit to taj mahal, seems like it still inspires young hearts..........} & \emph{a weekend adventure to agra to see the taj mahal, see also afternoon and night. luxury.} \\
        \midrule
        \rowcolor{Apricot!107} \emph{the royal mausoleum on the grounds of `iolani palace} & \emph{this pre-dates the taj mahal} \\
        \midrule
        \rowcolor{Apricot!95} \emph{$<$usr$>$ taj} & \emph{you couldn't photograph inside the tombs, so this is all i can show} \\
        \midrule
        \rowcolor{Apricot!82} \emph{rotunda at nmai} & \emph{the beauty of age, the mark of wisdom} \\
        \midrule
        \rowcolor{Apricot!70} \emph{outside of yet another palace} & \emph{don't remember where.} \\
        \midrule
        \rowcolor{Apricot!57} \emph{taken from city palace.} & $\downarrow$ \\
        \midrule
        \rowcolor{Apricot!45} \emph{photography ii} & $\downarrow$ \\
        \midrule
        \rowcolor{Apricot!32} \emph{it can be seen from most places in the city} & $\downarrow$ \\
        \midrule
        \rowcolor{Apricot!20} \emph{kla photography} & $\downarrow$ \\
        \midrule
        \texttt{\textbf{[ROOT]}} & \texttt{\textbf{[ROOT]}} \\
    \end{tabularx}
\end{figure*}

\clearpage

\begin{figure*}[h]
    \footnotesize
    \newcolumntype{Z}{>{\centering\arraybackslash}X}
    \setlength{\tabcolsep}{2pt}

    \vspace{5pt}
    \adjustbox{valign=t}{
        \begin{minipage}{0.18\linewidth}
            \centering
            \frame{\includegraphics[width=\linewidth]{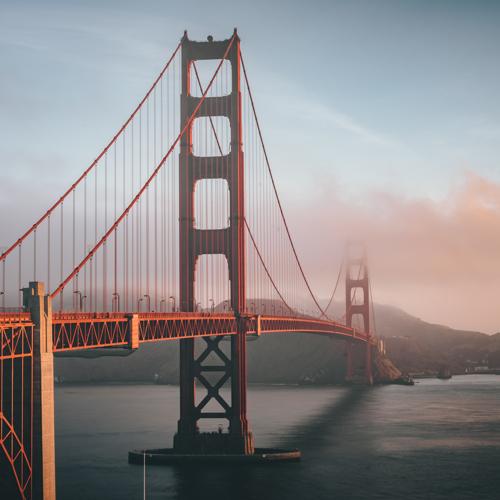}}
            (9)
        \end{minipage}
    }
    \hfill
    \begin{tabularx}{0.8\linewidth}[t]{ZZ}
        \rowcolor{Gray!40} \textbf{MERU} & \textbf{CLIP} \\
        \midrule
        \rowcolor{Apricot!120} \emph{a high dynamic range shot of the golden gate bridge on a foggy afternoon} & \emph{golden gate bridge thru photoshop lightroom} \\
        \midrule
        \rowcolor{Apricot!107} \emph{working on the perfect golden gate shot.} & \emph{golden gates} \\
        \midrule
        \rowcolor{Apricot!95} \emph{golden gate iii} & \emph{no...it's not the golden gate ;)} \\
        \midrule
        \rowcolor{Apricot!82} \emph{everyone who's been to sf has to take this photo at least once in their lives, right?} & \emph{by gusf bit.ly/17hga6r} \\
        \midrule
        \rowcolor{Apricot!70} \emph{just got back from sf. will post more on my photoblog: ohad.me} & \emph{the independent sf} \\
        \midrule
        \rowcolor{Apricot!57} \emph{<3 sf} & \emph{thinking about painting this makes my shoulder hurt.} \\
        \midrule
        \rowcolor{Apricot!45} \emph{come out and play sf} & \emph{still searching for the shot around here.} \\
        \midrule
        \rowcolor{Apricot!32} \emph{back from golden bay} & $\downarrow$ \\
        \midrule
        \rowcolor{Apricot!20} \emph{without fog} & $\downarrow$ \\
        \midrule
        \texttt{\textbf{[ROOT]}} & \texttt{\textbf{[ROOT]}} \\
    \end{tabularx}

    \vspace{5pt}
    \adjustbox{valign=t}{
        \begin{minipage}{0.18\linewidth}
            \centering
            \frame{\includegraphics[width=\linewidth]{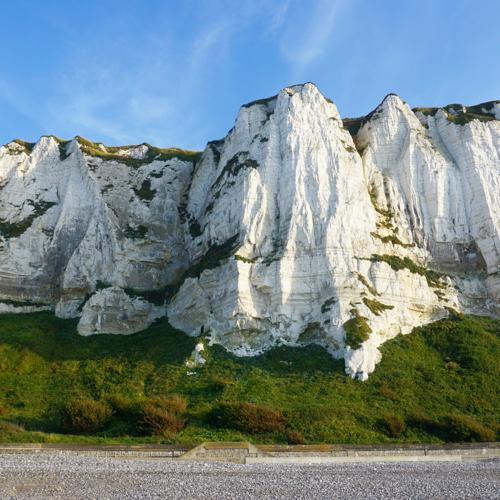}}
            (10)
        \end{minipage}
    }
    \hfill
    \begin{tabularx}{0.8\linewidth}[t]{ZZ}
        \rowcolor{Gray!40} \textbf{MERU} & \textbf{CLIP} \\
        \midrule
        \rowcolor{Apricot!120} \emph{white cliffs of dover. august, maybe 2004?} & \emph{coloured sand cliffs of alum bay, isle of wight, 1 may 2012.} \\
        \midrule
        \rowcolor{Apricot!107} \emph{calcite cumbria england} & \emph{the cliffs are made of limestone.} \\
        \midrule
        \rowcolor{Apricot!95} \emph{poland rocks.} & \emph{falkenberg from the south} \\
        \midrule
        \rowcolor{Apricot!82} \emph{white point natural area} & \emph{at juta village} \\
        \midrule
        \rowcolor{Apricot!70} \emph{balderstone close} & \emph{it's pretty rocky there.} \\
        \midrule
        \rowcolor{Apricot!57} \emph{nepomuk rocks...} & $\downarrow$ \\
        \midrule
        \rowcolor{Apricot!45} \emph{l'eglise de giverny} & $\downarrow$ \\
        \midrule
        \rowcolor{Apricot!32} \emph{one point if you can tell me where this was taken.} & $\downarrow$ \\
        \midrule
        \rowcolor{Apricot!20} \emph{also some kind of guenon, methinks.} & $\downarrow$ \\
        \midrule
        \texttt{\textbf{[ROOT]}} & \texttt{\textbf{[ROOT]}} \\
    \end{tabularx}

    \vspace{5pt}
    \adjustbox{valign=t}{
        \begin{minipage}{0.18\linewidth}
            \centering
            \frame{\includegraphics[width=\linewidth]{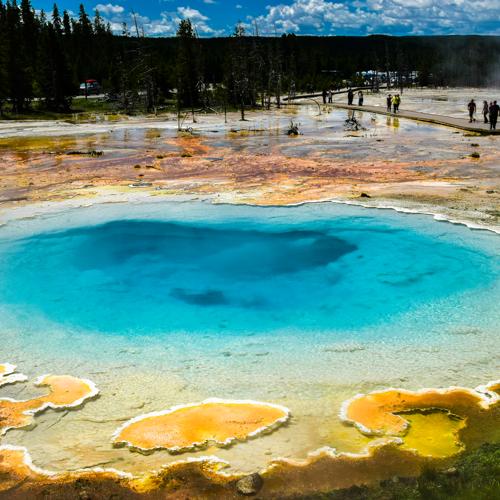}}
            (11)
        \end{minipage}
    }
    \hfill
    \begin{tabularx}{0.8\linewidth}[t]{ZZ}
        \rowcolor{Gray!40} \textbf{MERU} & \textbf{CLIP} \\
        \midrule
        \rowcolor{Apricot!120} \emph{at yellowstone national park there are geyser pools called painted pots because of the colors they exude.} & \emph{yellowstone - noth entrance} \\
        \midrule
        \rowcolor{Apricot!107} \emph{yellowstone - noth entrance} & \emph{...like i was, how yellowstone got its name} \\
        \midrule
        \rowcolor{Apricot!95} \emph{no trip to yellowstone is complete without it} & \emph{i don't remember where this one was. it was striking.} \\
        \midrule
        \rowcolor{Apricot!82} \emph{people enjoy the hot spring, even at this time.} & \emph{ain't he a beaut?} \\
        \midrule
        \rowcolor{Apricot!70} \emph{there are hot springs around here somewhere...} & $\downarrow$ \\
        \midrule
        \rowcolor{Apricot!57} \emph{wy'east} & $\downarrow$ \\
        \midrule
        \rowcolor{Apricot!45} \emph{so many places that were stunning to look at.} & $\downarrow$ \\
        \midrule
        \rowcolor{Apricot!32} \emph{there are some special places in the earth. this is one !} & $\downarrow$ \\
        \midrule
        \rowcolor{Apricot!20} \emph{with this photo... it's almost like taking a vacation just looking at this.} & $\downarrow$ \\
        \midrule
        \texttt{\textbf{[ROOT]}} & \texttt{\textbf{[ROOT]}} \\
    \end{tabularx}

    \vspace{5pt}
    \adjustbox{valign=t}{
        \begin{minipage}{0.18\linewidth}
            \centering
            \frame{\includegraphics[width=\linewidth]{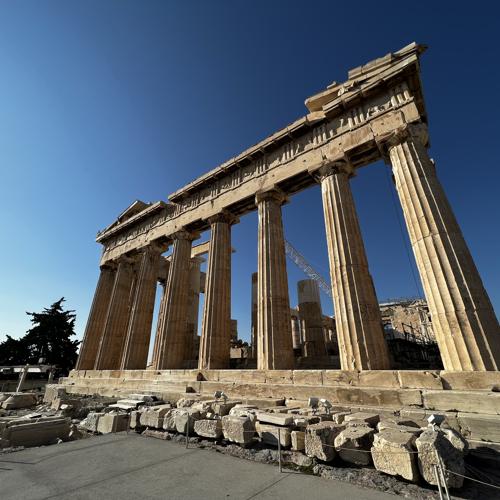}}
            (12)
        \end{minipage}
    }
    \hfill
    \begin{tabularx}{0.8\linewidth}[t]{ZZ}
        \rowcolor{Gray!40} \textbf{MERU} & \textbf{CLIP} \\
        \midrule
        \rowcolor{Apricot!120} \emph{the new parthenon museum, next to the acropolis.} & \emph{athens archaeological site of the acropolis the parthenon} \\
        \midrule
        \rowcolor{Apricot!105} \emph{this is the magnificent temple of zeus, located in the center of athens} & \emph{temple of jupiter and ruins - selinunte} \\
        \midrule
        \rowcolor{Apricot!91} \emph{lil-bit bigger than the athens arch} & \emph{ruins from roman time} \\
        \midrule
        \rowcolor{Apricot!77} \emph{roman building , later used as royal residence .} & \emph{can't remember where this was} \\
        \midrule
        \rowcolor{Apricot!62} \emph{roman fort/settlement.} & $\downarrow$ \\
        \midrule
        \rowcolor{Apricot!48} \emph{built in roman times} & $\downarrow$ \\
        \midrule
        \rowcolor{Apricot!34} \emph{we will miss our old place.} & $\downarrow$ \\
        \midrule
        \rowcolor{Apricot!20} \emph{at quimbledon.} & $\downarrow$ \\
        \midrule
        \texttt{\textbf{[ROOT]}} & \texttt{\textbf{[ROOT]}} \\
    \end{tabularx}
\end{figure*}

\clearpage

\begin{figure*}[h]
    \footnotesize
    \newcolumntype{Z}{>{\centering\arraybackslash}X}
    \setlength{\tabcolsep}{2pt}

    \vspace{5pt}
    \adjustbox{valign=t}{
        \begin{minipage}{0.18\linewidth}
            \centering
            \frame{\includegraphics[width=\linewidth]{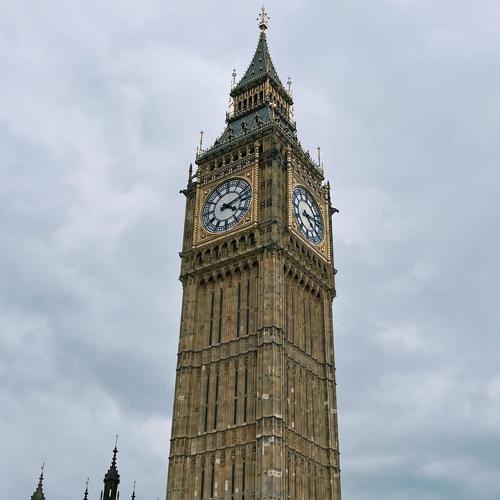}}
            (13)
        \end{minipage}
    }
    \hfill
    \begin{tabularx}{0.8\linewidth}[t]{ZZ}
        \rowcolor{Gray!40} \textbf{MERU} & \textbf{CLIP} \\
        \midrule
        \rowcolor{Apricot!120} \emph{obligatory big ben shot} & \emph{big ben i el parlament} \\
        \midrule
        \rowcolor{Apricot!100} \emph{it's what they call a 'big clock' could be thought of as 'big ben'} & \emph{or big ben, to his friends} \\
        \midrule
        \rowcolor{Apricot!80} \emph{yeah, i know: big ben is the bell, not the clock tower...} & $\downarrow$ \\
        \midrule
        \rowcolor{Apricot!60} \emph{the famous tower} & $\downarrow$ \\
        \midrule
        \rowcolor{Apricot!40} \emph{i pretty much only took a photo of this because it was in english} & $\downarrow$ \\
        \midrule
        \rowcolor{Apricot!20} \emph{guess what time i took this picture .} & $\downarrow$ \\
        \midrule
        \texttt{\textbf{[ROOT]}} & \texttt{\textbf{[ROOT]}} \\
    \end{tabularx}

    \vspace{5pt}
    \adjustbox{valign=t}{
        \begin{minipage}{0.18\linewidth}
            \centering
            \frame{\includegraphics[width=\linewidth]{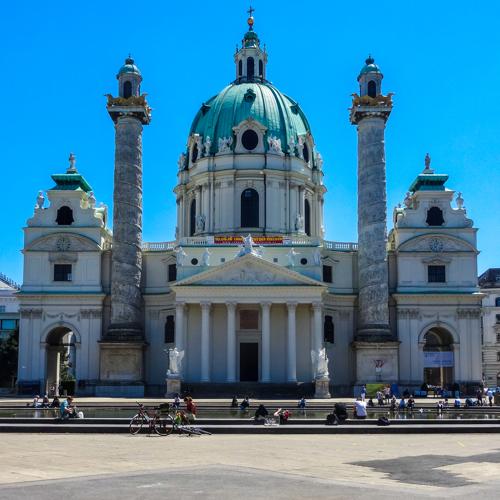}}
            (14)
        \end{minipage}
    }
    \hfill
    \begin{tabularx}{0.8\linewidth}[t]{ZZ}
        \rowcolor{Gray!40} \textbf{MERU} & \textbf{CLIP} \\
        \midrule
        \rowcolor{Apricot!120} \emph{karlskirche, just on the outer ring in vienna .} & \emph{kazan, kremin, annunciation cathedral 02/25/2007} \\
        \midrule
        \rowcolor{Apricot!95} \emph{vienna, austria - st. charles church} & \emph{bulgaria, sofia, st. nikolai russian church} \\
        \midrule
        \rowcolor{Apricot!70} \emph{wawel cathedral} & \emph{almost like a cathedral. a cathedral to transit.} \\
        \midrule
        \rowcolor{Apricot!45} \emph{near st. george's cathedral} & \emph{as beautiful as any cathedral} \\
        \midrule
        \rowcolor{Apricot!20} \emph{from my old pda.} & \emph{not far from the cathedral} \\
        \midrule
        \texttt{\textbf{[ROOT]}} & \texttt{\textbf{[ROOT]}} \\
    \end{tabularx}

    \vspace{5pt}
    \adjustbox{valign=t}{
        \begin{minipage}{0.18\linewidth}
            \centering
            \frame{\includegraphics[width=\linewidth]{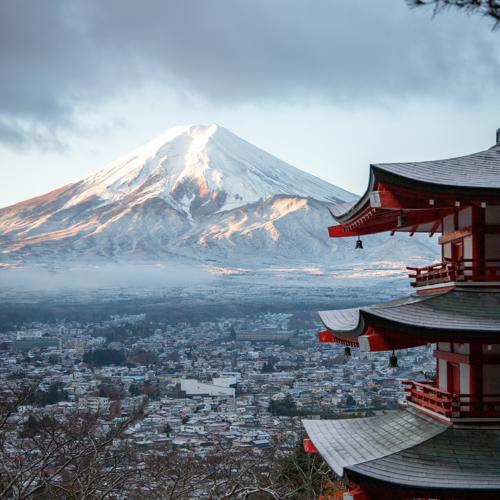}}
            (15)
        \end{minipage}
    }
    \hfill
    \begin{tabularx}{0.8\linewidth}[t]{ZZ}
        \rowcolor{Gray!40} \textbf{MERU} & \textbf{CLIP} \\
        \midrule
        \rowcolor{Apricot!120} \emph{who needs mt.fuji} & \emph{fuji provia 100} \\
        \midrule
        \rowcolor{Apricot!95} \emph{fuji} & \emph{fuji-q highlands} \\
        \midrule
        \rowcolor{Apricot!70} \emph{mt haba} & \emph{fuji f30} \\
        \midrule
        \rowcolor{Apricot!45} \emph{mount.} & \emph{fuji-san in the background...} \\
        \midrule
        \rowcolor{Apricot!20} \emph{at quimbledon.} & \emph{fairmount, in} \\
        \midrule
        \texttt{\textbf{[ROOT]}} & \texttt{\textbf{[ROOT]}} \\
    \end{tabularx}

    \vspace{5pt}
    \adjustbox{valign=t}{
        \begin{minipage}{0.18\linewidth}
            \centering
            \frame{\includegraphics[width=\linewidth]{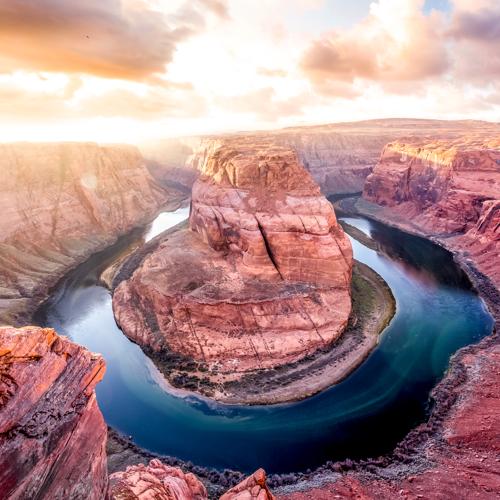}}
            (16)
        \end{minipage}
    }
    \hfill
    \begin{tabularx}{0.8\linewidth}[t]{ZZ}
        \rowcolor{Gray!40} \textbf{MERU} & \textbf{CLIP} \\
        \midrule
        \rowcolor{Apricot!120} \emph{a single exposure of horseshoe bend at sunrise. certainly one of my favorites from the trip.} & \emph{yes, there is that backdrop of horseshoe bend :)} \\
        \midrule
        \rowcolor{Apricot!103} \emph{horseshoe bend} & \emph{searching for the one ring} \\
        \midrule
        \rowcolor{Apricot!86} \emph{bend over} & \emph{$<$usr$>$,usa} \\
        \midrule
        \rowcolor{Apricot!70} \emph{looking back at horseshoe canyon} & $\downarrow$ \\
        \midrule
        \rowcolor{Apricot!53} \emph{horseshoe bay... this is as close to paradise as you can get!!!!} & $\downarrow$ \\
        \midrule
        \rowcolor{Apricot!36} \emph{canyon country, specifically} & $\downarrow$ \\
        \midrule
        \rowcolor{Apricot!20} \emph{if you use my photo please post a link and let me know.} & $\downarrow$ \\
        \midrule
        \texttt{\textbf{[ROOT]}} & \texttt{\textbf{[ROOT]}} \\
    \end{tabularx}

    \vspace{5pt}
    \adjustbox{valign=t}{
        \begin{minipage}{0.18\linewidth}
            \centering
            \frame{\includegraphics[width=\linewidth]{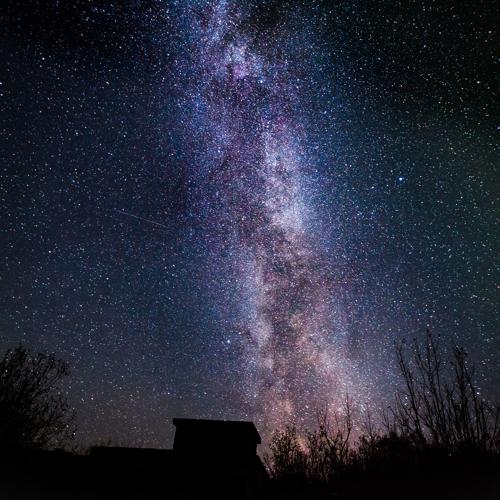}}
            (17)
        \end{minipage}
    }
    \hfill
    \begin{tabularx}{0.8\linewidth}[t]{ZZ}
        \rowcolor{Gray!40} \textbf{MERU} & \textbf{CLIP} \\
        \midrule
        \rowcolor{Apricot!120} \emph{the milk way over bleriot ferry provincial park near drumheller, alberta.} & \emph{the milky way as it appeared above the farmhouse in grey county - 30 sec exposure} \\
        \midrule
        \rowcolor{Apricot!86} \emph{the south-western part of the milky way} & \emph{outside a house from the austmarka region} \\
        \midrule
        \rowcolor{Apricot!53} \emph{we were in quite a rural place, although there were still lights on the horizon.} & \emph{this was just a couple of miles from the farmhouse we stayed in.} \\
        \midrule
        \rowcolor{Apricot!20} \emph{keeping the peace while bush was in town} & $\downarrow$ \\
        \midrule
        \texttt{\textbf{[ROOT]}} & \texttt{\textbf{[ROOT]}} \\
    \end{tabularx}
\end{figure*}

\clearpage

\begin{figure*}[h]
    \footnotesize
    \newcolumntype{Z}{>{\centering\arraybackslash}X}
    \setlength{\tabcolsep}{2pt}

    \vspace{5pt}
    \adjustbox{valign=t}{
        \begin{minipage}{0.18\linewidth}
            \centering
            \frame{\includegraphics[width=\linewidth]{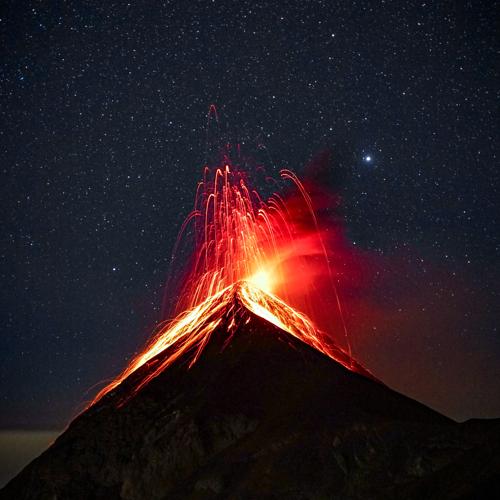}}
            (18)
        \end{minipage}
    }
    \hfill
    \begin{tabularx}{0.8\linewidth}[t]{ZZ}
        \rowcolor{Gray!40} \textbf{MERU} & \textbf{CLIP} \\
        \midrule
        \rowcolor{Apricot!120} \emph{lava as seen through night shot of volcan arenal} & \emph{lava as seen through night shot of volcan arenal} \\
        \midrule
        \rowcolor{Apricot!105} \emph{the majestic villarica volcano} & \emph{volcan osorno} \\
        \midrule
        \rowcolor{Apricot!91} \emph{nice photo of us in front of an active volcano.} & \emph{volcanic origin} \\
        \midrule
        \rowcolor{Apricot!77} \emph{still an active volcano} & \emph{the volcano!} \\
        \midrule
        \rowcolor{Apricot!62} \emph{around the khorgo volcano.} & \emph{with volcan lanin in the background} \\
        \midrule
        \rowcolor{Apricot!48} \emph{mt stromlo} & $\downarrow$ \\
        \midrule
        \rowcolor{Apricot!34} \emph{i'm not sure where we were for this shot.} & $\downarrow$ \\
        \midrule
        \rowcolor{Apricot!20} \emph{for some reason, i think this photo is great!} & $\downarrow$ \\
        \midrule
        \texttt{\textbf{[ROOT]}} & \texttt{\textbf{[ROOT]}} \\
    \end{tabularx}

    \vspace{5pt}
    \adjustbox{valign=t}{
        \begin{minipage}{0.18\linewidth}
            \centering
            \frame{\includegraphics[width=\linewidth]{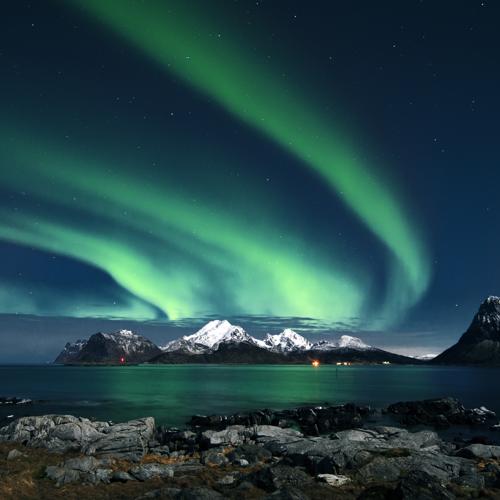}}
            (19)
        \end{minipage}
    }
    \hfill
    \begin{tabularx}{0.8\linewidth}[t]{ZZ}
        \rowcolor{Gray!40} \textbf{MERU} & \textbf{CLIP} \\
        \midrule
        \rowcolor{Apricot!120} \emph{arcs of the northen lights over the mountains neat troms, norway,} & \emph{northern lights on the otherside of patreksfjorur} \\
        \midrule
        \rowcolor{Apricot!95} \emph{northern lights, norway} & \emph{aurora boreale a kangerlussuaq} \\
        \midrule
        \rowcolor{Apricot!70} \emph{in the night see where this picture was taken.} & \emph{to a cinema the aurora} \\
        \midrule
        \rowcolor{Apricot!45} \emph{from the north. see where this picture was taken.} & \emph{the village at the end of the world} \\
        \midrule
        \rowcolor{Apricot!20} \emph{sometimes something just looks out of place !} & \emph{over a year's worth of photos here.} \\
        \midrule
        \texttt{\textbf{[ROOT]}} & \texttt{\textbf{[ROOT]}} \\
    \end{tabularx}

    \vspace{5pt}
    \adjustbox{valign=t}{
        \begin{minipage}{0.18\linewidth}
            \centering
            \frame{\includegraphics[width=\linewidth]{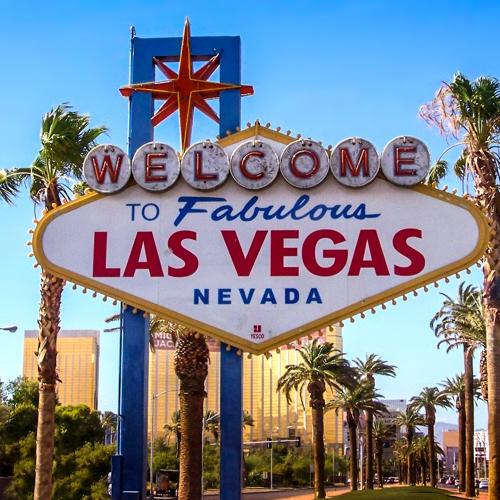}}
            (20)
        \end{minipage}
    }
    \hfill
    \begin{tabularx}{0.8\linewidth}[t]{ZZ}
        \rowcolor{Gray!40} \textbf{MERU} & \textbf{CLIP} \\
        \midrule
        \rowcolor{Apricot!120} \emph{cozy cone motel sign with tower of terror in background. california's adventure park at disneyland resort.} & \emph{adam taylor ollie over sign kodak: iso 200} \\
        \midrule
        \rowcolor{Apricot!95} \emph{my favorite tourist attraction in la.} & \emph{enjoying jason scott's talk.} \\
        \midrule
        \rowcolor{Apricot!70} \emph{if this place didn't scream la, i don't know what does.} & \emph{lost in las vegas- max ruckman} \\
        \midrule
        \rowcolor{Apricot!45} \emph{funny, this place was empty.} & \emph{hollywood rip, ride, rockit} \\
        \midrule
        \rowcolor{Apricot!20} \emph{photo : l.g.} & $\downarrow$ \\
        \midrule
        \texttt{\textbf{[ROOT]}} & \texttt{\textbf{[ROOT]}} \\
    \end{tabularx}

    \vspace{5pt}
    \adjustbox{valign=t}{
        \begin{minipage}{0.18\linewidth}
            \centering
            \frame{\includegraphics[width=\linewidth]{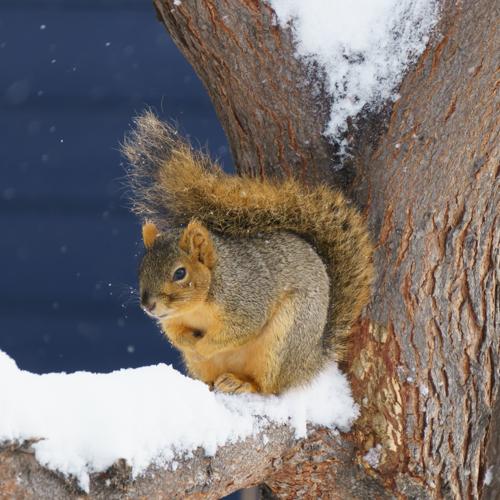}}
            (21)
        \end{minipage}
    }
    \hfill
    \begin{tabularx}{0.8\linewidth}[t]{ZZ}
        \rowcolor{Gray!40} \textbf{MERU} & \textbf{CLIP} \\
        \midrule
        \rowcolor{Apricot!120} \emph{a squirrel enjoying the snow on a not-very cold day.} & \emph{a squirrel enjoying the snow on a not-very cold day.} \\
        \midrule
        \rowcolor{Apricot!100} \emph{winter male still coming to food after the snow.} & \emph{winter male still coming to food after the snow.} \\
        \midrule
        \rowcolor{Apricot!80} \emph{i don't usually see these type of squirrels down here. loved this little guy. :)} & $\downarrow$ \\
        \midrule
        \rowcolor{Apricot!60} \emph{the last nut, my dear!} & $\downarrow$ \\
        \midrule
        \rowcolor{Apricot!40} \emph{it's kinda fuzzy. but i love this picture for some reason.} & $\downarrow$ \\
        \midrule
        \rowcolor{Apricot!20} \emph{i had to take one picture, okay?} & $\downarrow$ \\
        \midrule
        \texttt{\textbf{[ROOT]}} & \texttt{\textbf{[ROOT]}} \\
    \end{tabularx}

    \vspace{5pt}
    \adjustbox{valign=t}{
        \begin{minipage}{0.18\linewidth}
            \centering
            \frame{\includegraphics[width=\linewidth]{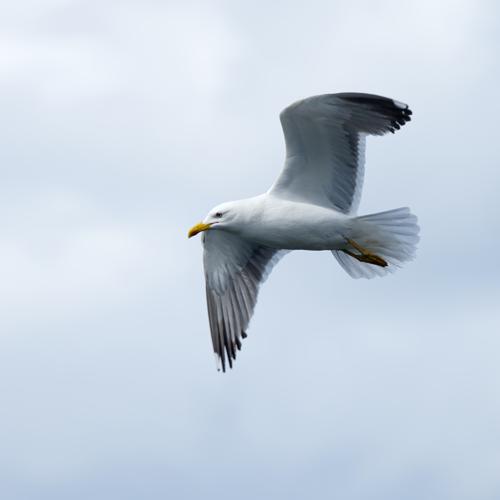}}
            (22)
        \end{minipage}
    }
    \hfill
    \begin{tabularx}{0.8\linewidth}[t]{ZZ}
        \rowcolor{Gray!40} \textbf{MERU} & \textbf{CLIP} \\
        \midrule
        \rowcolor{Apricot!120} \emph{a gull in flight in stratford} & \emph{a common or arctic tern flying above the scottish peninsula of kintyre.} \\
        \midrule
        \rowcolor{Apricot!107} \emph{gull on the wing} & \emph{more bird shots at dyrholaey.} \\
        \midrule
        \rowcolor{Apricot!95} \emph{we also saw gull-billed but never close enough to photo.} & \emph{wouldn't be a trip without at least one picture of a seagull} \\
        \midrule
        \rowcolor{Apricot!82} \emph{taken at little gull islands - b, little gull islands} & \emph{seagull!} \\
        \midrule
        \rowcolor{Apricot!70} \emph{taken with the seagull} & \emph{some bird thing.} \\
        \midrule
        \rowcolor{Apricot!57} \emph{i was running after the seagull as i took this photo.} & \emph{it took patience to get this shot since the stupid bird kept looking away} \\
        \midrule
        \rowcolor{Apricot!45} \emph{not a very nice bird, but still interesting to take pictures of...} & $\downarrow$ \\
        \midrule
        \rowcolor{Apricot!32} \emph{i took one westward shot, just to see it} & $\downarrow$ \\
        \midrule
        \rowcolor{Apricot!20} \emph{i keep taking this photograph} & $\downarrow$ \\
        \midrule
        \texttt{\textbf{[ROOT]}} & \texttt{\textbf{[ROOT]}} \\
    \end{tabularx}
\end{figure*}

\clearpage

\begin{figure*}[h]
    \footnotesize
    \newcolumntype{Z}{>{\centering\arraybackslash}X}
    \setlength{\tabcolsep}{2pt}

    \vspace{5pt}
    \adjustbox{valign=t}{
        \begin{minipage}{0.18\linewidth}
            \centering
            \frame{\includegraphics[width=\linewidth]{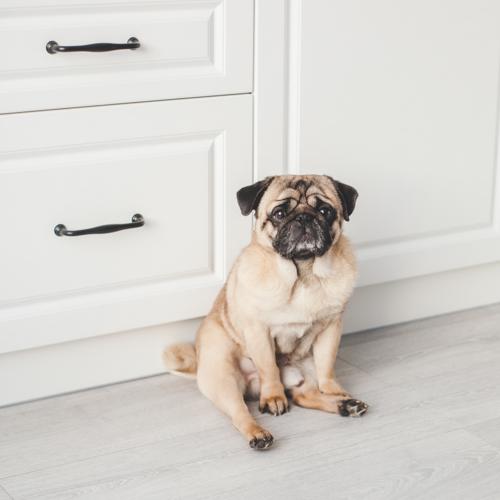}}
            (23)
        \end{minipage}
    }
    \hfill
    \begin{tabularx}{0.8\linewidth}[t]{ZZ}
        \rowcolor{Gray!40} \textbf{MERU} & \textbf{CLIP} \\
        \midrule
        \rowcolor{Apricot!120} \emph{margaret willie sanborn-sebo harvey henry sanborn pug dog framed} & \emph{sacha my friend and companion patiently waiting for dad to finish taking photos !!!} \\
        \midrule
        \rowcolor{Apricot!103} \emph{everyday bear takes up this position as he waits for his mom to make his food.} & \emph{eli begged mommy to take some photos} \\
        \midrule
        \rowcolor{Apricot!86} \emph{patiently waiting for the photographer to get his "shot".} & $\downarrow$ \\
        \midrule
        \rowcolor{Apricot!70} \emph{"this picture isn't going to work, and i'm going to show you why..."} & $\downarrow$ \\
        \midrule
        \rowcolor{Apricot!53} \emph{thinking to himself, "what's missing?"} & $\downarrow$ \\
        \midrule
        \rowcolor{Apricot!36} \emph{this is us... trying to be sultry.} & $\downarrow$ \\
        \midrule
        \rowcolor{Apricot!20} \emph{to do nothing or to do something.} & $\downarrow$ \\
        \midrule
        \texttt{\textbf{[ROOT]}} & \texttt{\textbf{[ROOT]}} \\
    \end{tabularx}

    \vspace{5pt}
    \adjustbox{valign=t}{
        \begin{minipage}{0.18\linewidth}
            \centering
            \frame{\includegraphics[width=\linewidth]{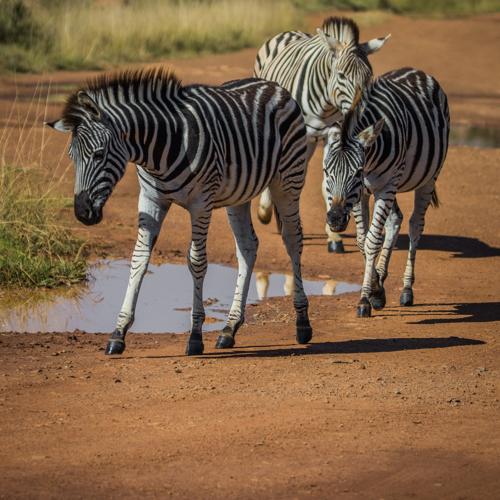}}
            (24)
        \end{minipage}
    }
    \hfill
    \begin{tabularx}{0.8\linewidth}[t]{ZZ}
        \rowcolor{Gray!40} \textbf{MERU} & \textbf{CLIP} \\
        \midrule
        \rowcolor{Apricot!120} \emph{zebras, ruaha national park} & \emph{zebras at kidepo national park, uganda.} \\
        \midrule
        \rowcolor{Apricot!108} \emph{zebra fouls with their mohicans, south africa} & \emph{zebra fouls with their mohicans, south africa} \\
        \midrule
        \rowcolor{Apricot!97} \emph{the zebras} & \emph{zebras at the watering hole} \\
        \midrule
        \rowcolor{Apricot!86} \emph{photographer: simone kuipers} & \emph{three zebras} \\
        \midrule
        \rowcolor{Apricot!75} \emph{photographer: mark antos } & \emph{good things come in threes. apparently, so do zebras.} \\
        \midrule
        \rowcolor{Apricot!64} \emph{from photo safari, take 2 .} & \emph{the group comes around the bend.} \\
        \midrule
        \rowcolor{Apricot!53} \emph{at oudja} & \emph{wild animals} \\
        \midrule
        \rowcolor{Apricot!42} \emph{at quimbledon.} & \emph{more straglers} \\
        \midrule
        \rowcolor{Apricot!31} $\downarrow$ & \emph{of the group} \\
        \midrule
        \rowcolor{Apricot!20} $\downarrow$ & \emph{just a little to the left of the middle} \\
        \midrule
        \texttt{\textbf{[ROOT]}} & \texttt{\textbf{[ROOT]}} \\
    \end{tabularx}

    \vspace{5pt}
    \adjustbox{valign=t}{
        \begin{minipage}{0.18\linewidth}
            \centering
            \frame{\includegraphics[width=\linewidth]{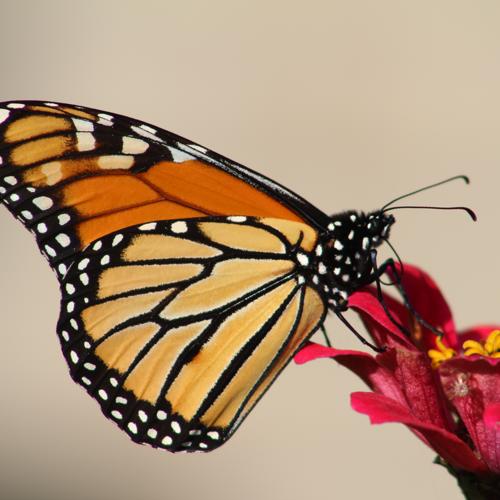}}
            (25)
        \end{minipage}
    }
    \hfill
    \begin{tabularx}{0.8\linewidth}[t]{ZZ}
        \rowcolor{Gray!40} \textbf{MERU} & \textbf{CLIP} \\
        \midrule
        \rowcolor{Apricot!120} \emph{monarch, danaus plexippus . shot in waimanalo, hawaii.} & \emph{a monarch pauses for a drink at a butterfly bush.} \\
        \midrule
        \rowcolor{Apricot!107} \emph{taken at the desert botanical garden in phoenix, arizona, during its seasonal monarch butterfly exhibit.} & \emph{monarch in a standard profile} \\
        \midrule
        \rowcolor{Apricot!95} \emph{i'm exercising to capture butterflies.} & \emph{the monarch} \\
        \midrule
        \rowcolor{Apricot!82} \emph{visitor to the butterfly tree,} & $\downarrow$ \\
        \midrule
        \rowcolor{Apricot!70} \emph{monarch} & $\downarrow$ \\
        \midrule
        \rowcolor{Apricot!57} \emph{butterflies are always free, so enjoy as many of them as you want.} & $\downarrow$ \\
        \midrule
        \rowcolor{Apricot!45} \emph{butterfly photography} & $\downarrow$ \\
        \midrule
        \rowcolor{Apricot!32} \emph{i tried to get more shots but it flew away.} & $\downarrow$ \\
        \midrule
        \rowcolor{Apricot!20} \emph{insect porn} & $\downarrow$ \\
        \midrule
        \texttt{\textbf{[ROOT]}} & \texttt{\textbf{[ROOT]}} \\
    \end{tabularx}

    \vspace{5pt}
    \adjustbox{valign=t}{
        \begin{minipage}{0.18\linewidth}
            \centering
            \frame{\includegraphics[width=\linewidth]{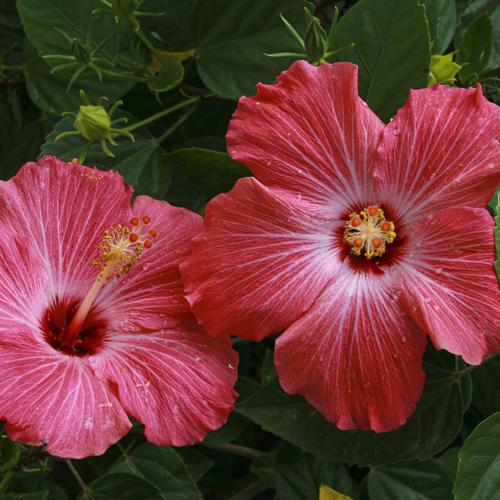}}
            (26)
        \end{minipage}
    }
    \hfill
    \begin{tabularx}{0.8\linewidth}[t]{ZZ}
        \rowcolor{Gray!40} \textbf{MERU} & \textbf{CLIP} \\
        \midrule
        \rowcolor{Apricot!120} \emph{i love the big blooms of the hibiscus with their bold colour.} & \emph{some hibiscus-like blossoms beside the visitor center at moody gardens in galveston, tx} \\
        \midrule
        \rowcolor{Apricot!100} \emph{these looked like hibiscus, but i think they are something else.} & \emph{after many years nurturing this back to life, the recent heat and rainfall have produced more spectacular blooms.} \\
        \midrule
        \rowcolor{Apricot!80} \emph{only a few blooms this time of year....} & \emph{this beautiful species may be the hibiscus according to my wife but i am not so sure,} \\
        \midrule
        \rowcolor{Apricot!60} \emph{much better bloom this time...} & \emph{these looked like hibiscus, but i think they are something else.} \\
        \midrule
        \rowcolor{Apricot!40} \emph{i usually dispise flower photos, but i actually like this one.} & \emph{looked like they had a lot of nice camera's and video gear. :-)} \\
        \midrule
        \rowcolor{Apricot!20} \emph{this is one is good.} & \emph{flourishing.} \\
        \midrule
        \texttt{\textbf{[ROOT]}} & \texttt{\textbf{[ROOT]}} \\
    \end{tabularx}
\end{figure*}

\clearpage

\begin{figure*}[h]
    \footnotesize
    \newcolumntype{Z}{>{\centering\arraybackslash}X}
    \setlength{\tabcolsep}{2pt}

    \vspace{5pt}
    \adjustbox{valign=t}{
        \begin{minipage}{0.18\linewidth}
            \centering
            \frame{\includegraphics[width=\linewidth]{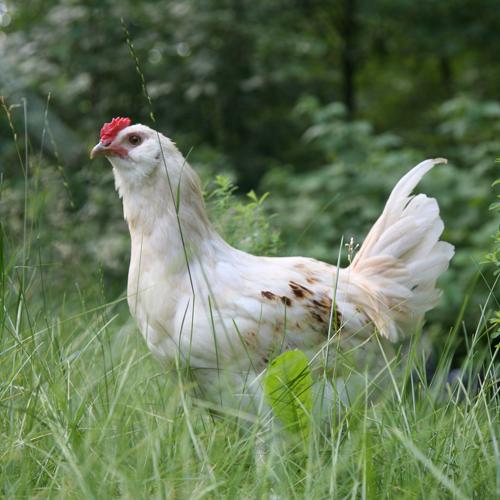}}
            (27)
        \end{minipage}
    }
    \hfill
    \begin{tabularx}{0.8\linewidth}[t]{ZZ}
        \rowcolor{Gray!40} \textbf{MERU} & \textbf{CLIP} \\
        \midrule
        \rowcolor{Apricot!120} \emph{crele old english game bantam cockerel} & \emph{read "the beautiful yellow bird - a cautionary tale for new mps" only at acid rabbi.} \\
        \midrule
        \rowcolor{Apricot!100} \emph{sabrina is running and twirling to the bachanalia song from the samson and delila opera.} & \emph{flaunt your assets!} \\
        \midrule
        \rowcolor{Apricot!80} \emph{really interesting gadgetar! posted by second life resident ina centaur. visit kaneohe.} & \emph{most of the girls had gone to college with mary, whose husband was competing.} \\
        \midrule
        \rowcolor{Apricot!60} \emph{harvey henry sanborn marie elizabeth campbell} & $\downarrow$ \\
        \midrule
        \rowcolor{Apricot!40} \emph{sam is showing good form and follow through here.} & $\downarrow$ \\
        \midrule
        \rowcolor{Apricot!20} \emph{this is only luky.} & $\downarrow$ \\
        \midrule
        \texttt{\textbf{[ROOT]}} & \texttt{\textbf{[ROOT]}} \\
    \end{tabularx}

    \vspace{5pt}
    \adjustbox{valign=t}{
        \begin{minipage}{0.18\linewidth}
            \centering
            \frame{\includegraphics[width=\linewidth]{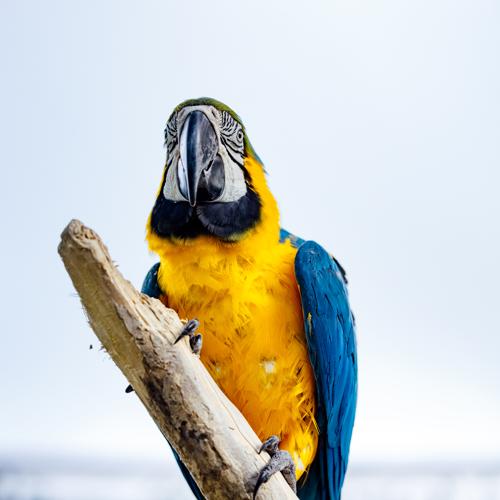}}
            (28)
        \end{minipage}
    }
    \hfill
    \begin{tabularx}{0.8\linewidth}[t]{ZZ}
        \rowcolor{Gray!40} \textbf{MERU} & \textbf{CLIP} \\
        \midrule
        \rowcolor{Apricot!120} \emph{blue-and-yellow macaw at the zooparque itatiba.} & \emph{i like this portrait of this blue and yellow macaw, because of the black background...} \\
        \midrule
        \rowcolor{Apricot!103} \emph{one of my fav birds to shoot at the zoo} & \emph{from parrot island} \\
        \midrule
        \rowcolor{Apricot!86} \emph{i like these birds, especially when you can see the yellow of their eyes!} & \emph{$<$usr$>$ bird sanctuary.} \\
        \midrule
        \rowcolor{Apricot!70} \emph{there are more colorful birds, but there are few birds with as much character.} & \emph{zero post processing} \\
        \midrule
        \rowcolor{Apricot!53} \emph{a beautiful bird, and was quite happy to pose for me.} & \emph{one of a bunch} \\
        \midrule
        \rowcolor{Apricot!36} \emph{never had a bird pose for well before!} & $\downarrow$ \\
        \midrule
        \rowcolor{Apricot!20} \emph{hi!some more photos...ana} & $\downarrow$ \\
        \midrule
        \texttt{\textbf{[ROOT]}} & \texttt{\textbf{[ROOT]}} \\
    \end{tabularx}

    \vspace{5pt}
    \adjustbox{valign=t}{
        \begin{minipage}{0.18\linewidth}
            \centering
            \frame{\includegraphics[width=\linewidth]{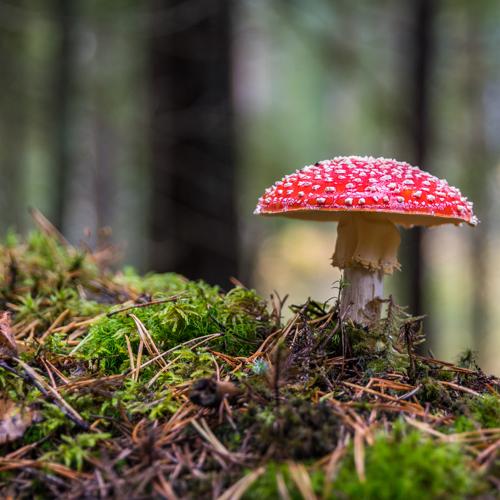}}
            (29)
        \end{minipage}
    }
    \hfill
    \begin{tabularx}{0.8\linewidth}[t]{ZZ}
        \rowcolor{Gray!40} \textbf{MERU} & \textbf{CLIP} \\
        \midrule
        \rowcolor{Apricot!120} \emph{fly agaric in the forest with a little spider.} & \emph{freshly popped amanita muscaria in the forest.} \\
        \midrule
        \rowcolor{Apricot!100} \emph{all alone on the forest floor} & \emph{a fly agaric on the rise.} \\
        \midrule
        \rowcolor{Apricot!80} \emph{from a recent new forest trip} & \emph{or "fly agaric". "if your viking gets to choose."} \\
        \midrule
        \rowcolor{Apricot!60} \emph{i like this photo, so here it is too.} & \emph{reminded me a bit of alice in wonderland} \\
        \midrule
        \rowcolor{Apricot!40} $\downarrow$ & \emph{reminded me of alice in wonderland} \\
        \midrule
        \rowcolor{Apricot!20} $\downarrow$ & \emph{this one goes out to forest love.} \\
        \midrule
        \texttt{\textbf{[ROOT]}} & \texttt{\textbf{[ROOT]}} \\
    \end{tabularx}

    \vspace{5pt}
    \adjustbox{valign=t}{
        \begin{minipage}{0.18\linewidth}
            \centering
            \frame{\includegraphics[width=\linewidth]{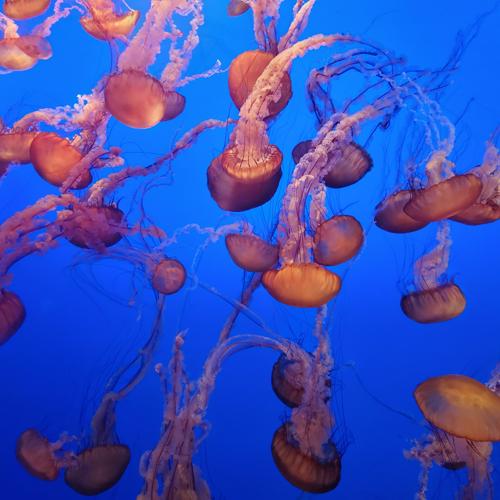}}
            (30)
        \end{minipage}
    }
    \hfill
    \begin{tabularx}{0.8\linewidth}[t]{ZZ}
        \rowcolor{Gray!40} \textbf{MERU} & \textbf{CLIP} \\
        \midrule
        \rowcolor{Apricot!120} \emph{taken at mystic aquarium, ct} & \emph{taken by michael i love watching jellyfish. wish these pics had turned out better.} \\
        \midrule
        \rowcolor{Apricot!103} \emph{i don't really like jellyfish, but they are beautiful.} & \emph{jelly.} \\
        \midrule
        \rowcolor{Apricot!86} \emph{something about this reminds me of every photo i've ever taken of jellyfish} & $\downarrow$ \\
        \midrule
        \rowcolor{Apricot!70} \emph{it is really very cool to be able to see them...} & $\downarrow$ \\
        \midrule
        \rowcolor{Apricot!53} \emph{it did not feel like an aquarium when i took the picture} & $\downarrow$ \\
        \midrule
        \rowcolor{Apricot!36} \emph{really not much they let you see here.} & $\downarrow$ \\
        \midrule
        \rowcolor{Apricot!20} \emph{that i met.} & $\downarrow$ \\
        \midrule
        \texttt{\textbf{[ROOT]}} & \texttt{\textbf{[ROOT]}} \\
    \end{tabularx}

    \vspace{5pt}
    \adjustbox{valign=t}{
        \begin{minipage}{0.18\linewidth}
            \centering
            \frame{\includegraphics[width=\linewidth]{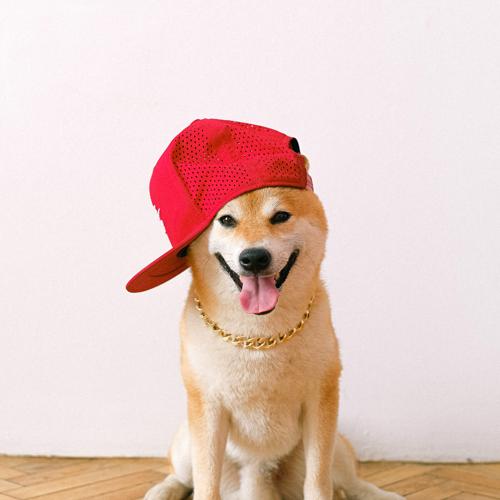}}
            (31)
        \end{minipage}
    }
    \hfill
    \begin{tabularx}{0.8\linewidth}[t]{ZZ}
        \rowcolor{Gray!40} \textbf{MERU} & \textbf{CLIP} \\
        \midrule
        \rowcolor{Apricot!120} \emph{in new hat from adji} & \emph{includes my twit hat- wink} \\
        \midrule
        \rowcolor{Apricot!103} \emph{oui, girl friend.} & \emph{in new hat from adji} \\
        \midrule
        \rowcolor{Apricot!86} \emph{the hat actually belongs to honey :)} & \emph{:) he's still in japan right now... i miss him.} \\
        \midrule
        \rowcolor{Apricot!70} \emph{fashion photo session} & $\downarrow$ \\
        \midrule
        \rowcolor{Apricot!53} \emph{with new hat} & $\downarrow$ \\
        \midrule
        \rowcolor{Apricot!36} \emph{a very fit lady.} & $\downarrow$ \\
        \midrule
        \rowcolor{Apricot!20} \emph{the boy has style!} & $\downarrow$ \\
        \midrule
        \texttt{\textbf{[ROOT]}} & \texttt{\textbf{[ROOT]}} \\
    \end{tabularx}
\end{figure*}

\clearpage

\begin{figure*}[h]
    \footnotesize
    \newcolumntype{Z}{>{\centering\arraybackslash}X}
    \setlength{\tabcolsep}{2pt}

    \vspace{5pt}
    \adjustbox{valign=t}{
        \begin{minipage}{0.18\linewidth}
            \centering
            \frame{\includegraphics[width=\linewidth]{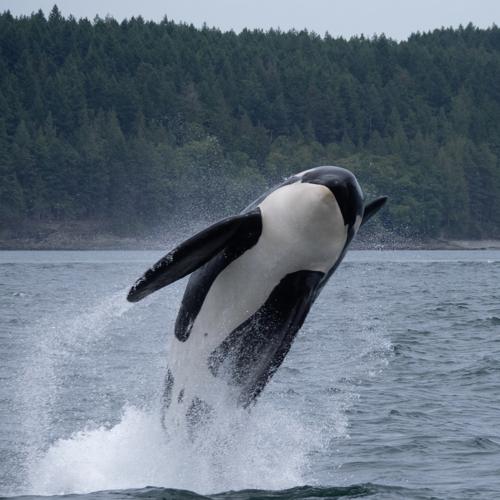}}
            (32)
        \end{minipage}
    }
    \hfill
    \begin{tabularx}{0.8\linewidth}[t]{ZZ}
        \rowcolor{Gray!40} \textbf{MERU} & \textbf{CLIP} \\
        \midrule
        \rowcolor{Apricot!120} \emph{humpback coming up for air near juneau, alaska} & \emph{orca, craig, alaska, tongass nf. usfs francisco sanchez} \\
        \midrule
        \rowcolor{Apricot!103} \emph{here, this orca swims about before the show starts.} & \emph{named for its shape. not for the occasional whales found in its waters.} \\
        \midrule
        \rowcolor{Apricot!86} \emph{an orca during the show "believe" in sea world .} & \emph{a bit shakey!} \\
        \midrule
        \rowcolor{Apricot!70} \emph{humpback} & \emph{takin' off.} \\
        \midrule
        \rowcolor{Apricot!53} \emph{a shamu in action} & $\downarrow$ \\
        \midrule
        \rowcolor{Apricot!36} \emph{pretend you see a whale and i'll take your photo.} & $\downarrow$ \\
        \midrule
        \rowcolor{Apricot!20} \emph{cropitornot shot this one} & $\downarrow$ \\
        \midrule
        \texttt{\textbf{[ROOT]}} & \texttt{\textbf{[ROOT]}} \\
    \end{tabularx}

    \vspace{5pt}
    \adjustbox{valign=t}{
        \begin{minipage}{0.18\linewidth}
            \centering
            \frame{\includegraphics[width=\linewidth]{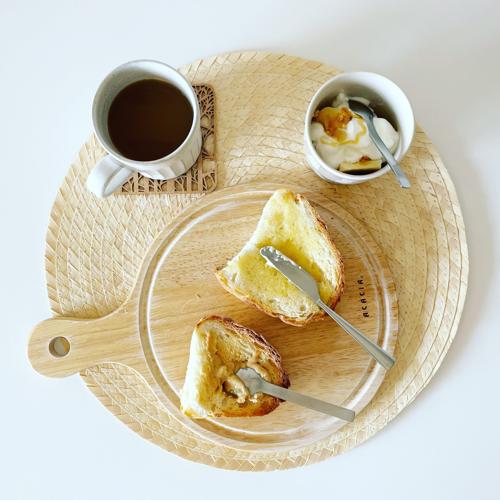}}
            (33)
        \end{minipage}
    }
    \hfill
    \begin{tabularx}{0.8\linewidth}[t]{ZZ}
        \rowcolor{Gray!40} \textbf{MERU} & \textbf{CLIP} \\
        \midrule
        \rowcolor{Apricot!120} \emph{a pastry \& a coffee for breakfast yummy} & \emph{today's breakfast: toast with honey, egg and black coffee} \\
        \midrule
        \rowcolor{Apricot!95} \emph{a right and proper afternoon tea \#1} & \emph{afternoon cream tea} \\
        \midrule
        \rowcolor{Apricot!70} \emph{made to perfection. much needed morning tea!} & \emph{have a nice cup of tea and a sandwich.} \\
        \midrule
        \rowcolor{Apricot!45} \emph{some photos from me when i working at home} & $\downarrow$ \\
        \midrule
        \rowcolor{Apricot!20} \emph{sobrepeso en la proa} & $\downarrow$ \\
        \midrule
        \texttt{\textbf{[ROOT]}} & \texttt{\textbf{[ROOT]}} \\
    \end{tabularx}

    \vspace{5pt}
    \adjustbox{valign=t}{
        \begin{minipage}{0.18\linewidth}
            \centering
            \frame{\includegraphics[width=\linewidth]{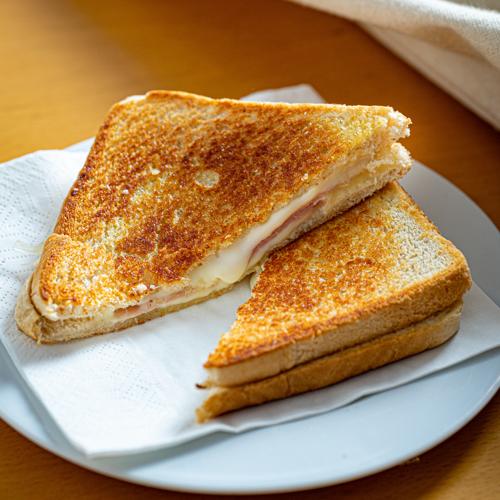}}
            (34)
        \end{minipage}
    }
    \hfill
    \begin{tabularx}{0.8\linewidth}[t]{ZZ}
        \rowcolor{Gray!40} \textbf{MERU} & \textbf{CLIP} \\
        \midrule
        \rowcolor{Apricot!120} \emph{'tooey's delight' gc16aga} & \emph{dominic samonte \& stephanie estabillo} \\
        \midrule
        \rowcolor{Apricot!105} \emph{30 seconds into gina's five-minute toast.} & \emph{chicken and cheese sandwich.} \\
        \midrule
        \rowcolor{Apricot!91} \emph{the plain toast was, um, plain.} & $\downarrow$ \\
        \midrule
        \rowcolor{Apricot!77} \emph{with grilled cheese sandwich.} & $\downarrow$ \\
        \midrule
        \rowcolor{Apricot!62} \emph{i don't normally like toasted sandwiches, but this one was delicious!} & $\downarrow$ \\
        \midrule
        \rowcolor{Apricot!48} \emph{on bread, probably not what you should do with it, but it was a good meal} & $\downarrow$ \\
        \midrule
        \rowcolor{Apricot!34} \emph{call center del club cantv} & $\downarrow$ \\
        \midrule
        \rowcolor{Apricot!20} \emph{so. good.} & $\downarrow$ \\
        \midrule
        \texttt{\textbf{[ROOT]}} & \texttt{\textbf{[ROOT]}} \\
    \end{tabularx}

    \vspace{5pt}
    \adjustbox{valign=t}{
        \begin{minipage}{0.18\linewidth}
            \centering
            \frame{\includegraphics[width=\linewidth]{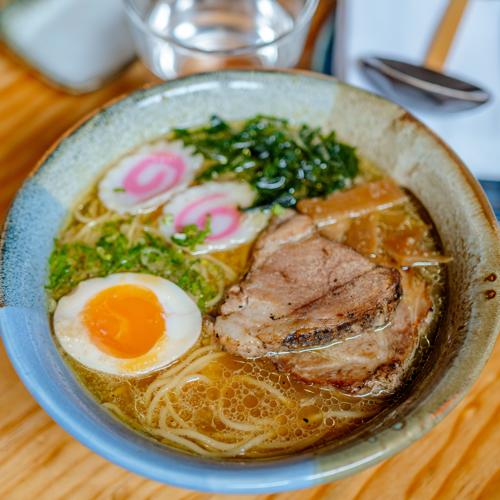}}
            (35)
        \end{minipage}
    }
    \hfill
    \begin{tabularx}{0.8\linewidth}[t]{ZZ}
        \rowcolor{Gray!40} \textbf{MERU} & \textbf{CLIP} \\
        \midrule
        \rowcolor{Apricot!120} \emph{at momofuku} & \emph{where the magic happens at momofuku noodle bar} \\
        \midrule
        \rowcolor{Apricot!95} \emph{the ramen got pwned.} & \emph{@ terakawa ramen 4sq.com/ekseer} \\
        \midrule
        \rowcolor{Apricot!70} \emph{siawase ramen} & \emph{ramen exploration - still looking} \\
        \midrule
        \rowcolor{Apricot!45} \emph{this i ate, and it was great!} & \emph{with michael cotta} \\
        \midrule
        \rowcolor{Apricot!20} \emph{i ate some of this.} & $\downarrow$ \\
        \midrule
        \texttt{\textbf{[ROOT]}} & \texttt{\textbf{[ROOT]}} \\
    \end{tabularx}

    \vspace{5pt}
    \adjustbox{valign=t}{
        \begin{minipage}{0.18\linewidth}
            \centering
            \frame{\includegraphics[width=\linewidth]{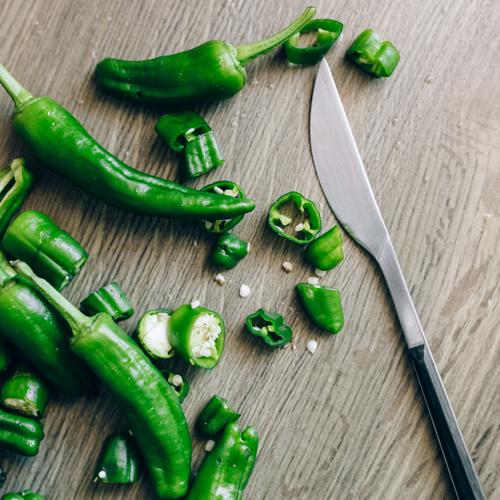}}
            (36)
        \end{minipage}
    }
    \hfill
    \begin{tabularx}{0.8\linewidth}[t]{ZZ}
        \rowcolor{Gray!40} \textbf{MERU} & \textbf{CLIP} \\
        \midrule
        \rowcolor{Apricot!120} \emph{tiny hot peppers from the freezer} & \emph{poblanos to be roasted} \\
        \midrule
        \rowcolor{Apricot!103} \emph{the ones with jalapenos are the ones that are ruling} & \emph{they're best as peppers from the garden!} \\
        \midrule
        \rowcolor{Apricot!86} \emph{add the chillies and cook for another minute.} & \emph{do nothing gardening in action!} \\
        \midrule
        \rowcolor{Apricot!70} \emph{getting ready for some chili-dippin'} & $\downarrow$ \\
        \midrule
        \rowcolor{Apricot!53} \emph{yo tengo el poder!} & $\downarrow$ \\
        \midrule
        \rowcolor{Apricot!36} \emph{toying around with a f/1.8} & $\downarrow$ \\
        \midrule
        \rowcolor{Apricot!20} \emph{you canno tmake this stuff up} & $\downarrow$ \\
        \midrule
        \texttt{\textbf{[ROOT]}} & \texttt{\textbf{[ROOT]}} \\
    \end{tabularx}
\end{figure*}

\clearpage

\begin{figure*}[h]
    \footnotesize
    \newcolumntype{Z}{>{\centering\arraybackslash}X}
    \setlength{\tabcolsep}{2pt}

    \vspace{5pt}
    \adjustbox{valign=t}{
        \begin{minipage}{0.18\linewidth}
            \centering
            \frame{\includegraphics[width=\linewidth]{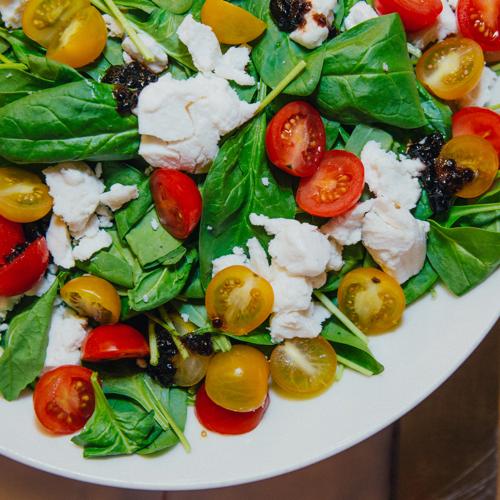}}
            (37)
        \end{minipage}
    }
    \hfill
    \begin{tabularx}{0.8\linewidth}[t]{ZZ}
        \rowcolor{Gray!40} \textbf{MERU} & \textbf{CLIP} \\
        \midrule
        \rowcolor{Apricot!120} \emph{last nights' salad, cropped} & \emph{the first of many caprese salads while we traveled.} \\
        \midrule
        \rowcolor{Apricot!95} \emph{nice salad, summery, fresh.} & \emph{for a homegrown salad} \\
        \midrule
        \rowcolor{Apricot!70} \emph{contemp salad} & \emph{another salad :)} \\
        \midrule
        \rowcolor{Apricot!45} \emph{i made the salad} & $\downarrow$ \\
        \midrule
        \rowcolor{Apricot!20} \emph{it's a verso/kveton thing.} & $\downarrow$ \\
        \midrule
        \texttt{\textbf{[ROOT]}} & \texttt{\textbf{[ROOT]}} \\
    \end{tabularx}

    \vspace{5pt}
    \adjustbox{valign=t}{
        \begin{minipage}{0.18\linewidth}
            \centering
            \frame{\includegraphics[width=\linewidth]{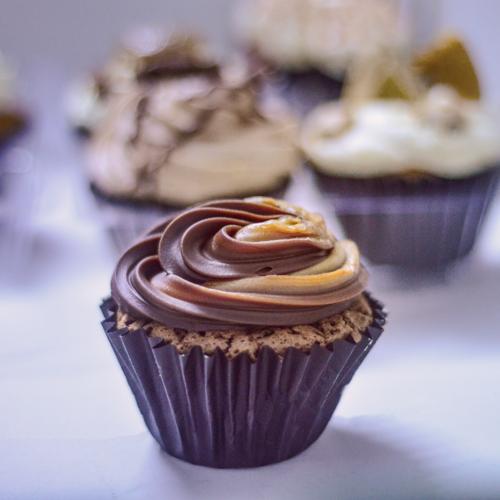}}
            (38)
        \end{minipage}
    }
    \hfill
    \begin{tabularx}{0.8\linewidth}[t]{ZZ}
        \rowcolor{Gray!40} \textbf{MERU} & \textbf{CLIP} \\
        \midrule
        \rowcolor{Apricot!120} \emph{awesome guinness \& chocolate cupcakes by kari stewart our wonderful studio manager} & \emph{i made cupcakes and took a million photos...} \\
        \midrule
        \rowcolor{Apricot!105} \emph{peanut butter cupcakes with whipped chocolate ganache.} & \emph{from the chocolate lady} \\
        \midrule
        \rowcolor{Apricot!91} \emph{vegan cupcakes. chocolate, coffee and cinnamon.} & $\downarrow$ \\
        \midrule
        \rowcolor{Apricot!77} \emph{coffee and chocolate: a "can't miss" cupcake...} & $\downarrow$ \\
        \midrule
        \rowcolor{Apricot!62} \emph{including a flourless chocolate cupcake!} & $\downarrow$ \\
        \midrule
        \rowcolor{Apricot!48} \emph{chocolate makes everything better. thanks $<$usr$>$} & $\downarrow$ \\
        \midrule
        \rowcolor{Apricot!34} \emph{chocolate-y goodness!} & $\downarrow$ \\
        \midrule
        \rowcolor{Apricot!20} \emph{this week's take, brought over by a friend.} & $\downarrow$ \\
        \midrule
        \texttt{\textbf{[ROOT]}} & \texttt{\textbf{[ROOT]}} \\
    \end{tabularx}

    \vspace{5pt}
    \adjustbox{valign=t}{
        \begin{minipage}{0.18\linewidth}
            \centering
            \frame{\includegraphics[width=\linewidth]{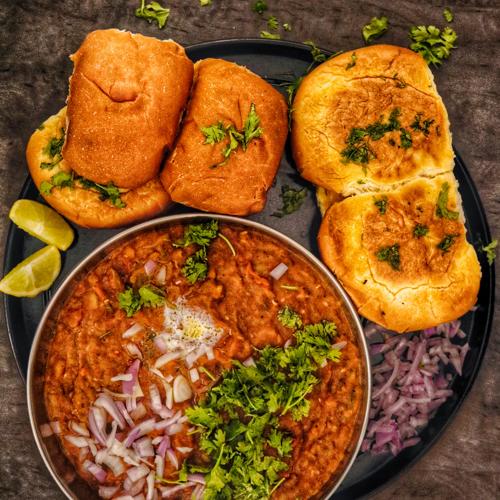}}
            (39)
        \end{minipage}
    }
    \hfill
    \begin{tabularx}{0.8\linewidth}[t]{ZZ}
        \rowcolor{Gray!40} \textbf{MERU} & \textbf{CLIP} \\
        \midrule
        \rowcolor{Apricot!120} \emph{new delhi's best cholle bhature} & \emph{pav bhaji appetizer, \$5 dinner time only tirupathi bhimas, milpitas} \\
        \midrule
        \rowcolor{Apricot!103} \emph{en tlaquepaque, jal.} & \emph{dal monte la motta .} \\
        \midrule
        \rowcolor{Apricot!86} \emph{west indian food. you can't help but smile when you eat it .} & \emph{the famous curry mile, taken on saturday 5th march} \\
        \midrule
        \rowcolor{Apricot!70} \emph{$<$usr$>$ buse dal lof} & \emph{sloppy everything} \\
        \midrule
        \rowcolor{Apricot!53} \emph{paraje guer aike} & \emph{the midas, great food} \\
        \midrule
        \rowcolor{Apricot!36} \emph{manoush} & $\downarrow$ \\
        \midrule
        \rowcolor{Apricot!20} \emph{with sambuca} & $\downarrow$ \\
        \midrule
        \texttt{\textbf{[ROOT]}} & \texttt{\textbf{[ROOT]}} \\
    \end{tabularx}

    \vspace{5pt}
    \adjustbox{valign=t}{
        \begin{minipage}{0.18\linewidth}
            \centering
            \frame{\includegraphics[width=\linewidth]{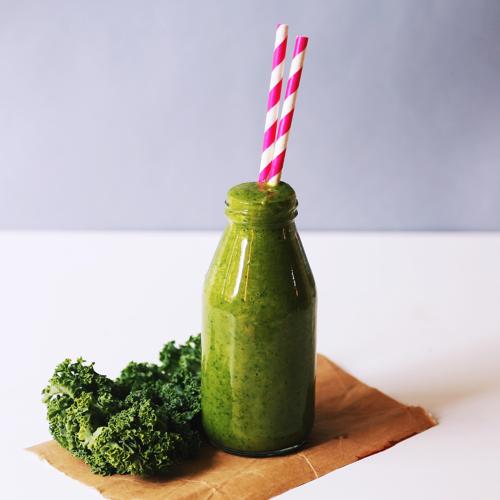}}
            (40)
        \end{minipage}
    }
    \hfill
    \begin{tabularx}{0.8\linewidth}[t]{ZZ}
        \rowcolor{Gray!40} \textbf{MERU} & \textbf{CLIP} \\
        \midrule
        \rowcolor{Apricot!120} \emph{matcha green tea ice blended} & \emph{chimichurri sauce recipe} \\
        \midrule
        \rowcolor{Apricot!100} \emph{yes i had a pesto drink} & \emph{trust me, i tried to make this less green.} \\
        \midrule
        \rowcolor{Apricot!80} \emph{hot/fermented} & \emph{make some good ones!} \\
        \midrule
        \rowcolor{Apricot!60} \emph{that's right, i've been experimenting. trying to keep things fresh.} & $\downarrow$ \\
        \midrule
        \rowcolor{Apricot!40} \emph{ondel-ondel} & $\downarrow$ \\
        \midrule
        \rowcolor{Apricot!20} \emph{one love hi pawa} & $\downarrow$ \\
        \midrule
        \texttt{\textbf{[ROOT]}} & \texttt{\textbf{[ROOT]}} \\
    \end{tabularx}

    \vspace{5pt}
    \adjustbox{valign=t}{
        \begin{minipage}{0.18\linewidth}
            \centering
            \frame{\includegraphics[width=\linewidth]{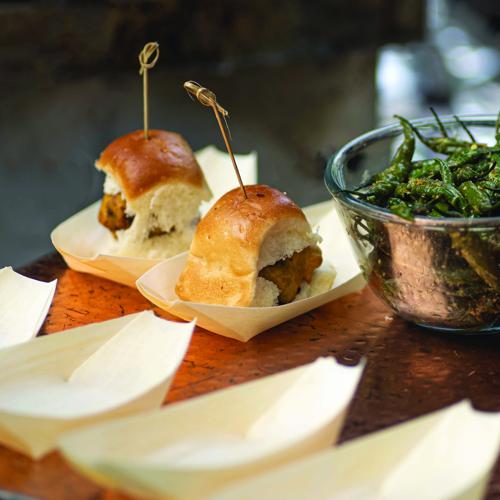}}
            (41)
        \end{minipage}
    }
    \hfill
    \begin{tabularx}{0.8\linewidth}[t]{ZZ}
        \rowcolor{Gray!40} \textbf{MERU} & \textbf{CLIP} \\
        \midrule
        \rowcolor{Apricot!120} \emph{bread with rosemary and garlic infused olive oil at jaleo, a tapas restaurant in my neighborhood.} & \emph{sliders served at lee roy selmon's restaurant.} \\
        \midrule
        \rowcolor{Apricot!103} \emph{with rosemary and parmigiano... our mellow new year's menu} & \emph{biergarten} \\
        \midrule
        \rowcolor{Apricot!86} \emph{bread with olive oil and vinegar} & \emph{sliders \& greens} \\
        \midrule
        \rowcolor{Apricot!82} \emph{my welcome brunch to vienna..a cheese party} & $\downarrow$ \\
        \midrule
        \rowcolor{Apricot!70} \emph{they serve it with some sort of sauce . this is their version of "bread".} & $\downarrow$ \\
        \midrule
        \rowcolor{Apricot!53} \emph{a good shot of how the bread should look.} & $\downarrow$ \\
        \midrule
        \rowcolor{Apricot!36} \emph{the bread is real, i think. at least, not glass.} & $\downarrow$ \\
        \midrule
        \rowcolor{Apricot!20} \emph{trying out some bread} & $\downarrow$ \\
        \midrule
        \texttt{\textbf{[ROOT]}} & \texttt{\textbf{[ROOT]}} \\
    \end{tabularx}
\end{figure*}

\clearpage

\begin{figure*}[h]
    \footnotesize
    \newcolumntype{Z}{>{\centering\arraybackslash}X}
    \setlength{\tabcolsep}{2pt}

    \vspace{5pt}
    \adjustbox{valign=t}{
        \begin{minipage}{0.18\linewidth}
            \centering
            \frame{\includegraphics[width=\linewidth]{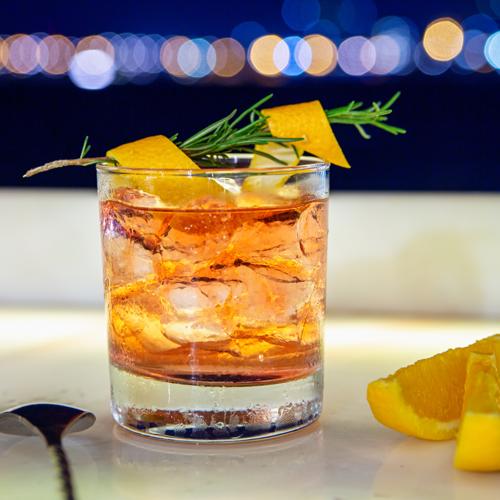}}
            (42)
        \end{minipage}
    }
    \hfill
    \begin{tabularx}{0.8\linewidth}[t]{ZZ}
        \rowcolor{Gray!40} \textbf{MERU} & \textbf{CLIP} \\
        \midrule
        \rowcolor{Apricot!120} \emph{adam johnston} & \emph{sparkling apple juice my blog: mikaeladanvers.com} \\
        \midrule
        \rowcolor{Apricot!103} \emph{smoked salmon vodka.} & \emph{nombre sells the accesories} \\
        \midrule
        \rowcolor{Apricot!86} \emph{it's so cold in here! look at the frosty vodka} & \emph{taste like it's fermenting into alcohol.} \\
        \midrule
        \rowcolor{Apricot!70} \emph{warm vodka. i'm still cringing.} & $\downarrow$ \\
        \midrule
        \rowcolor{Apricot!53} \emph{tasty, tasty cocktails} & $\downarrow$ \\
        \midrule
        \rowcolor{Apricot!36} \emph{made with hugin} & $\downarrow$ \\
        \midrule
        \rowcolor{Apricot!20} \emph{off for cocktails} & $\downarrow$ \\
        \midrule
        \texttt{\textbf{[ROOT]}} & \texttt{\textbf{[ROOT]}} \\
    \end{tabularx}

    \vspace{5pt}
    \adjustbox{valign=t}{
        \begin{minipage}{0.18\linewidth}
            \centering
            \frame{\includegraphics[width=\linewidth]{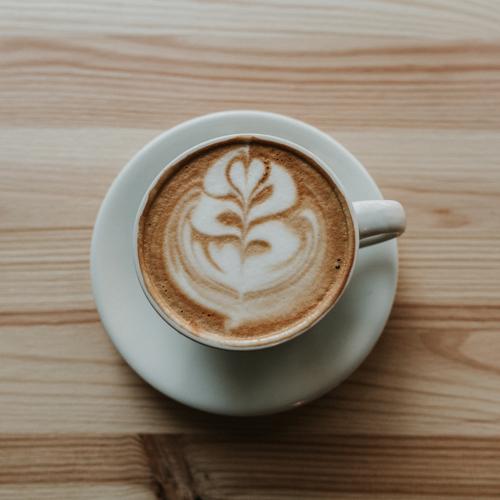}}
            (43)
        \end{minipage}
    }
    \hfill
    \begin{tabularx}{0.8\linewidth}[t]{ZZ}
        \rowcolor{Gray!40} \textbf{MERU} & \textbf{CLIP} \\
        \midrule
        \rowcolor{Apricot!120} \emph{it feels like winter again \#coffeemornings} & \emph{a latte from blue bottle coffee in oakland, ca.} \\
        \midrule
        \rowcolor{Apricot!100} \emph{by phil o'kane aka icedcoffee} & \emph{sweet cold coffee of destiny especially for $<$usr$>$ :-) 1 likes on instagram} \\
        \midrule
        \rowcolor{Apricot!80} \emph{1 comments on instagram: sorelle\_de\_latte: you looked great!} & $\downarrow$ \\
        \midrule
        \rowcolor{Apricot!60} \emph{heh, maybe my latte art chops aren't so bad after all :-)} & $\downarrow$ \\
        \midrule
        \rowcolor{Apricot!40} \emph{cheers $<$usr$>$ ;)} & $\downarrow$ \\
        \midrule
        \rowcolor{Apricot!20} \emph{i likes this one more, i think.} & $\downarrow$ \\
        \midrule
        \texttt{\textbf{[ROOT]}} & \texttt{\textbf{[ROOT]}} \\
    \end{tabularx}

    \vspace{5pt}
    \adjustbox{valign=t}{
        \begin{minipage}{0.18\linewidth}
            \centering
            \frame{\includegraphics[width=\linewidth]{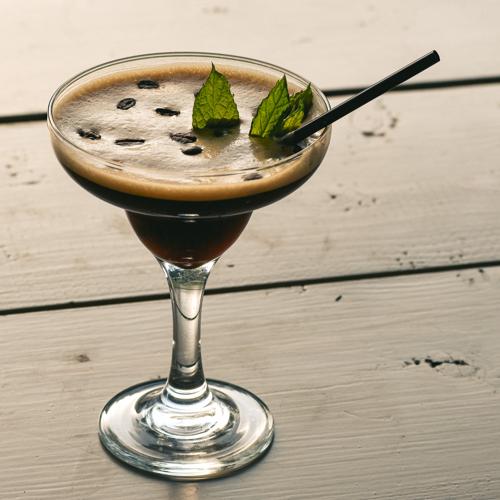}}
            (44)
        \end{minipage}
    }
    \hfill
    \begin{tabularx}{0.8\linewidth}[t]{ZZ}
        \rowcolor{Gray!40} \textbf{MERU} & \textbf{CLIP} \\
        \midrule
        \rowcolor{Apricot!120} \emph{ceu do mapia} & \emph{caffe ladro at 5 corners in edmonds. americanvirus.com} \\
        \midrule
        \rowcolor{Apricot!100} \emph{with a leaf, at the caffe espresso.} & \emph{chocolate catalan donkey with dinosaur egs, at eastern time} \\
        \midrule
        \rowcolor{Apricot!80} \emph{bourbon \& branch} & $\downarrow$ \\
        \midrule
        \rowcolor{Apricot!60} \emph{the braan} & $\downarrow$ \\
        \midrule
        \rowcolor{Apricot!40} \emph{enjoying the riff raff} & $\downarrow$ \\
        \midrule
        \rowcolor{Apricot!20} \emph{b b: the braes} & $\downarrow$ \\
        \midrule
        \texttt{\textbf{[ROOT]}} & \texttt{\textbf{[ROOT]}} \\
    \end{tabularx}

    \vspace{5pt}
    \adjustbox{valign=t}{
        \begin{minipage}{0.18\linewidth}
            \centering
            \frame{\includegraphics[width=\linewidth]{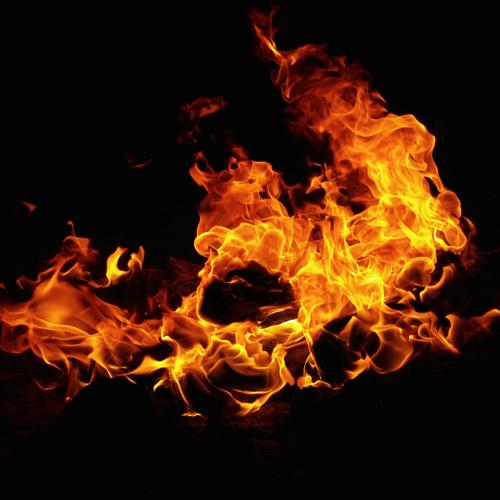}}
            (45)
        \end{minipage}
    }
    \hfill
    \begin{tabularx}{0.8\linewidth}[t]{ZZ}
        \rowcolor{Gray!40} \textbf{MERU} & \textbf{CLIP} \\
        \midrule
        \rowcolor{Apricot!120} \emph{i was well pleased that i managed to capture the flambe moment ;o)} & \emph{usfws photo/heather webb} \\
        \midrule
        \rowcolor{Apricot!95} \emph{everyone was taking pictures of the fire} & $\downarrow$ \\
        \midrule
        \rowcolor{Apricot!70} \emph{note to self: need to make more photos of fire again.} & $\downarrow$ \\
        \midrule
        \rowcolor{Apricot!45} \emph{this was one of the other interesting things. fire is always fun ;)} & $\downarrow$ \\
        \midrule
        \rowcolor{Apricot!20} \emph{i so wanted to photograph this.} & $\downarrow$ \\
        \midrule
        \texttt{\textbf{[ROOT]}} & \texttt{\textbf{[ROOT]}} \\
    \end{tabularx}

    \vspace{5pt}
    \adjustbox{valign=t}{
        \begin{minipage}{0.18\linewidth}
            \centering
            \frame{\includegraphics[width=\linewidth]{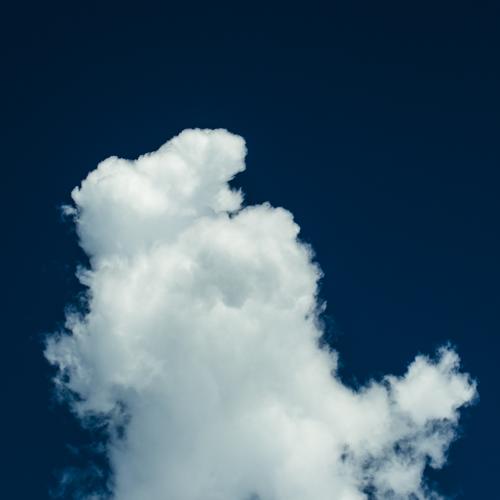}}
            (46)
        \end{minipage}
    }
    \hfill
    \begin{tabularx}{0.8\linewidth}[t]{ZZ}
        \rowcolor{Gray!40} \textbf{MERU} & \textbf{CLIP} \\
        \midrule
        \rowcolor{Apricot!120} \emph{i caught a cloud! i caught a cloud! *grin*} & \emph{textures courtesty of shadowhousecreations.blogspot.com/} \\
        \midrule
        \rowcolor{Apricot!95} \emph{it's the cloud, baby!} & \emph{my first cloud photo ever. :)} \\
        \midrule
        \rowcolor{Apricot!70} \emph{i liked this cloud.} & \emph{blowing cloud!} \\
        \midrule
        \rowcolor{Apricot!45} \emph{i took this one because of that cloud. seriously.} & $\downarrow$ \\
        \midrule
        \rowcolor{Apricot!20} \emph{i like this. i think it liked me.} & $\downarrow$ \\
        \midrule
        \texttt{\textbf{[ROOT]}} & \texttt{\textbf{[ROOT]}} \\
    \end{tabularx}
\end{figure*}

\clearpage

\begin{figure*}[h]
    \footnotesize
    \newcolumntype{Z}{>{\centering\arraybackslash}X}
    \setlength{\tabcolsep}{2pt}

    \vspace{5pt}
    \adjustbox{valign=t}{
        \begin{minipage}{0.18\linewidth}
            \centering
            \frame{\includegraphics[width=\linewidth]{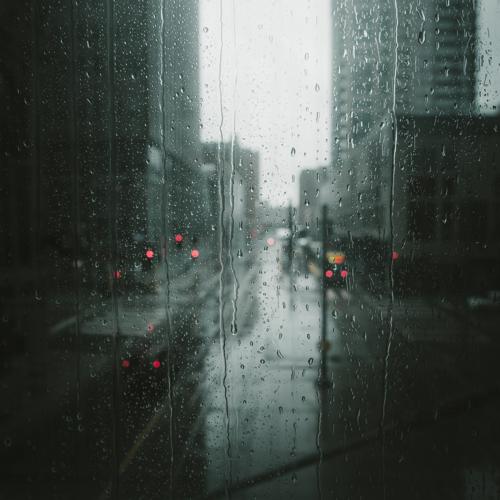}}
            (47)
        \end{minipage}
    }
    \hfill
    \begin{tabularx}{0.8\linewidth}[t]{ZZ}
        \rowcolor{Gray!40} \textbf{MERU} & \textbf{CLIP} \\
        \midrule
        \rowcolor{Apricot!120} \emph{a drizzly february day in vancouver.} & \emph{taken with a disposable camera on a rainy seattle day.} \\
        \midrule
        \rowcolor{Apricot!103} \emph{weather gloomy all the way to chicago} & \emph{a cold rainy day in chicago} \\
        \midrule
        \rowcolor{Apricot!86} \emph{summer rain in the city.} & \emph{junechicago} \\
        \midrule
        \rowcolor{Apricot!70} \emph{street$<$usr$>$} & $\downarrow$ \\
        \midrule
        \rowcolor{Apricot!53} \emph{$<$usr$>$ street} & $\downarrow$ \\
        \midrule
        \rowcolor{Apricot!36} \emph{urban as usual} & $\downarrow$ \\
        \midrule
        \rowcolor{Apricot!20} \emph{street pics} & $\downarrow$ \\
        \midrule
        \texttt{\textbf{[ROOT]}} & \texttt{\textbf{[ROOT]}} \\
    \end{tabularx}

    \vspace{5pt}
    \adjustbox{valign=t}{
        \begin{minipage}{0.18\linewidth}
            \centering
            \frame{\includegraphics[width=\linewidth]{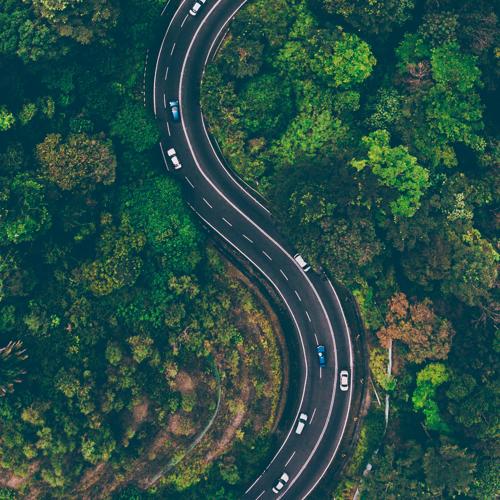}}
            (48)
        \end{minipage}
    }
    \hfill
    \begin{tabularx}{0.8\linewidth}[t]{ZZ}
        \rowcolor{Gray!40} \textbf{MERU} & \textbf{CLIP} \\
        \midrule
        \rowcolor{Apricot!120} \emph{this beautiful track goes through deep gorges of nilgiris towards mangalore.} & \emph{sepang international circuit malaysia} \\
        \midrule
        \rowcolor{Apricot!95} \emph{cipularang highway} & \emph{very well finished road for indonesia!! ... looking down the road} \\
        \midrule
        \rowcolor{Apricot!70} \emph{$<$usr$>$ road} & \emph{nature highway..} \\
        \midrule
        \rowcolor{Apricot!45} \emph{kinokuniya just down the road} & \emph{if you don't know where you are going, any road will take you there.} \\
        \midrule
        \rowcolor{Apricot!20} \emph{crookhaven} & \emph{look, there's a road and everything.} \\
        \midrule
        \texttt{\textbf{[ROOT]}} & \texttt{\textbf{[ROOT]}} \\
    \end{tabularx}

    \vspace{5pt}
    \adjustbox{valign=t}{
        \begin{minipage}{0.18\linewidth}
            \centering
            \frame{\includegraphics[width=\linewidth]{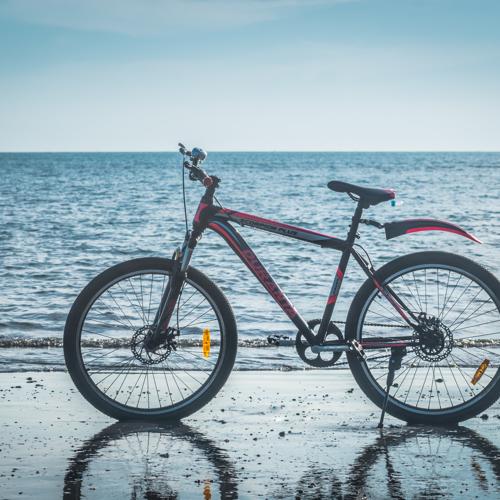}}
            (49)
        \end{minipage}
    }
    \hfill
    \begin{tabularx}{0.8\linewidth}[t]{ZZ}
        \rowcolor{Gray!40} \textbf{MERU} & \textbf{CLIP} \\
        \midrule
        \rowcolor{Apricot!120} \emph{lebei is driving the quadricycle we rented on the toronto islands.} & \emph{my cruiser on 80 mile beach. messing with my 15mm zeiss} \\
        \midrule
        \rowcolor{Apricot!86} \emph{this moto is still the best anything i've ever owned.} & \emph{biked to the nearest beach and took some pictures.} \\
        \midrule
        \rowcolor{Apricot!53} \emph{you have no idea how much i love this thing.} & \emph{i was traveling alone, so i instead of me i had to take pictures of my bike} \\
        \midrule
        \rowcolor{Apricot!20} $\downarrow$ & \emph{...and she rides like a dream. initial impressions ride report} \\
        \midrule
        \texttt{\textbf{[ROOT]}} & \texttt{\textbf{[ROOT]}} \\
    \end{tabularx}

    \vspace{5pt}
    \adjustbox{valign=t}{
        \begin{minipage}{0.18\linewidth}
            \centering
            \frame{\includegraphics[width=\linewidth]{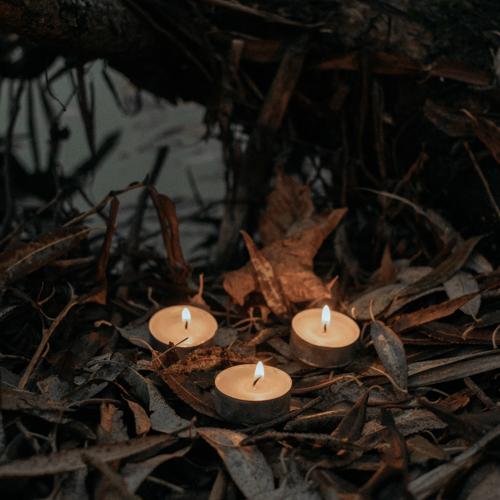}}
            (50)
        \end{minipage}
    }
    \hfill
    \begin{tabularx}{0.8\linewidth}[t]{ZZ}
        \rowcolor{Gray!40} \textbf{MERU} & \textbf{CLIP} \\
        \midrule
        \rowcolor{Apricot!120} \emph{in a few of the branches of some trees were little tea-light lanterns} & \emph{in a few of the branches of some trees were little tea-light lanterns} \\
        \midrule
        \rowcolor{Apricot!107} \emph{earth hour candle light} & \emph{they're only lanterns} \\
        \midrule
        \rowcolor{Apricot!95} \emph{my first try making this photo with a mini lantern} & \emph{2nd shoot coming out too} \\
        \midrule
        \rowcolor{Apricot!82} \emph{quant little light.} & \emph{not lit} \\
        \midrule
        \rowcolor{Apricot!70} \emph{little lights.} & $\downarrow$ \\
        \midrule
        \rowcolor{Apricot!57} \emph{i actually detest gold, but i liked the material contrast here.} & $\downarrow$ \\
        \midrule
        \rowcolor{Apricot!45} \emph{i just really liked the light, alright?} & $\downarrow$ \\
        \midrule
        \rowcolor{Apricot!32} \emph{a pain to get photos of these.} & $\downarrow$ \\
        \midrule
        \rowcolor{Apricot!20} \emph{if you use my photo please post a link and let me know.} & $\downarrow$ \\
        \midrule
        \texttt{\textbf{[ROOT]}} & \texttt{\textbf{[ROOT]}} \\
    \end{tabularx}

    \vspace{5pt}
    \adjustbox{valign=t}{
        \begin{minipage}{0.18\linewidth}
            \centering
            \frame{\includegraphics[width=\linewidth]{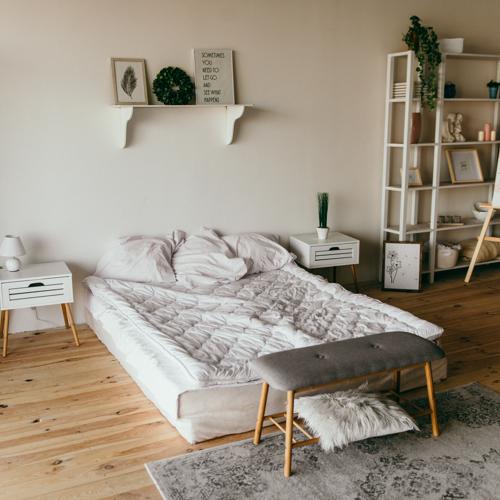}}
            (51)
        \end{minipage}
    }
    \hfill
    \begin{tabularx}{0.8\linewidth}[t]{ZZ}
        \rowcolor{Gray!40} \textbf{MERU} & \textbf{CLIP} \\
        \midrule
        \rowcolor{Apricot!120} \emph{simple scandinavian style decor. clifton, bristol} & \emph{ikea/helsinki design week party} \\
        \midrule
        \rowcolor{Apricot!95} \emph{rental apartment in berlin} & \emph{with jeff from simple plan in denmark} \\
        \midrule
        \rowcolor{Apricot!70} \emph{my flat getting more and more comfortable with more furniture coming in - and notice i now have fans!} & $\downarrow$ \\
        \midrule
        \rowcolor{Apricot!45} \emph{our first apartment} & $\downarrow$ \\
        \midrule
        \rowcolor{Apricot!20} \emph{ready made living space} & $\downarrow$ \\
        \midrule
        \texttt{\textbf{[ROOT]}} & \texttt{\textbf{[ROOT]}} \\
    \end{tabularx}
\end{figure*}

\clearpage

\begin{figure*}[h]
    \footnotesize
    \newcolumntype{Z}{>{\centering\arraybackslash}X}
    \setlength{\tabcolsep}{2pt}

    \vspace{5pt}
    \adjustbox{valign=t}{
        \begin{minipage}{0.18\linewidth}
            \centering
            \frame{\includegraphics[width=\linewidth]{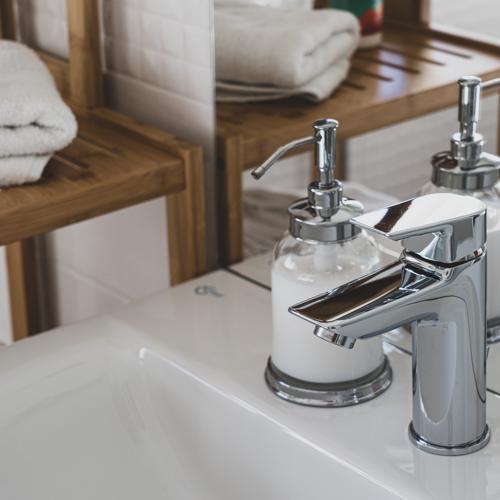}}
            (52)
        \end{minipage}
    }
    \hfill
    \begin{tabularx}{0.8\linewidth}[t]{ZZ}
        \rowcolor{Gray!40} \textbf{MERU} & \textbf{CLIP} \\
        \midrule
        \rowcolor{Apricot!120} \emph{studio di personaggio. character design.} & \emph{model: yasemin snoek stylist: melanie vink} \\
        \midrule
        \rowcolor{Apricot!107} \emph{decorated with silver and nickle plated} & \emph{we bought one of these for a friend's wedding. no, not you julee.} \\
        \midrule
        \rowcolor{Apricot!95} \emph{new faucet set.} & $\downarrow$ \\
        \midrule
        \rowcolor{Apricot!82} \emph{elegant bath items, though} & $\downarrow$ \\
        \midrule
        \rowcolor{Apricot!70} \emph{oh to have that kind of luxury} & $\downarrow$ \\
        \midrule
        \rowcolor{Apricot!57} \emph{a little luxury} & $\downarrow$ \\
        \midrule
        \rowcolor{Apricot!45} \emph{rich sigfrit kicks things off.} & $\downarrow$ \\
        \midrule
        \rowcolor{Apricot!32} \emph{the things i'll do for a shot} & $\downarrow$ \\
        \midrule
        \rowcolor{Apricot!20} \emph{can you guess what club this is for?!} & $\downarrow$ \\
        \midrule
        \texttt{\textbf{[ROOT]}} & \texttt{\textbf{[ROOT]}} \\
    \end{tabularx}

    \vspace{5pt}
    \adjustbox{valign=t}{
        \begin{minipage}{0.18\linewidth}
            \centering
            \frame{\includegraphics[width=\linewidth]{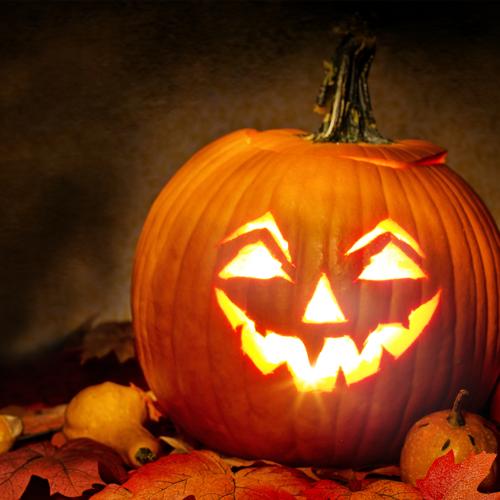}}
            (53)
        \end{minipage}
    }
    \hfill
    \begin{tabularx}{0.8\linewidth}[t]{ZZ}
        \rowcolor{Gray!40} \textbf{MERU} & \textbf{CLIP} \\
        \midrule
        \rowcolor{Apricot!120} \emph{pentax *ist ds/ iso 1600 || happy halloween!} & \emph{jack-o-lantern with other light up decorations. jack-o-lantern.} \\
        \midrule
        \rowcolor{Apricot!100} \emph{we wish you a happy all hallows night!} & \emph{i am so spoooooky} \\
        \midrule
        \rowcolor{Apricot!80} \emph{happy hallowe'en, everyone!} & $\downarrow$ \\
        \midrule
        \rowcolor{Apricot!60} \emph{happy halloween, yo.} & $\downarrow$ \\
        \midrule
        \rowcolor{Apricot!40} \emph{no be long now jack} & $\downarrow$ \\
        \midrule
        \rowcolor{Apricot!20} \emph{can you feel the spirit of e-xtrategy?} & $\downarrow$ \\
        \midrule
        \texttt{\textbf{[ROOT]}} & \texttt{\textbf{[ROOT]}} \\
    \end{tabularx}

    \vspace{5pt}
    \adjustbox{valign=t}{
        \begin{minipage}{0.18\linewidth}
            \centering
            \frame{\includegraphics[width=\linewidth]{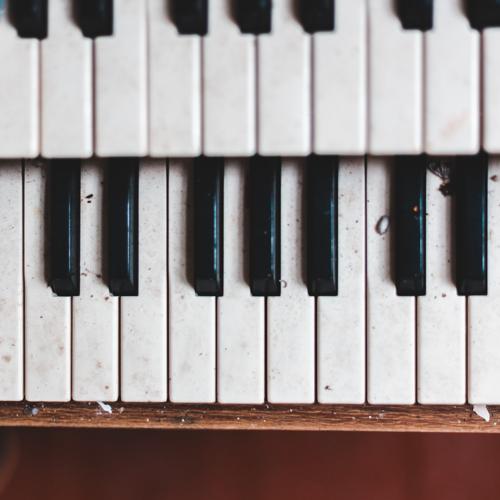}}
            (54)
        \end{minipage}
    }
    \hfill
    \begin{tabularx}{0.8\linewidth}[t]{ZZ}
        \rowcolor{Gray!40} \textbf{MERU} & \textbf{CLIP} \\
        \midrule
        \rowcolor{Apricot!120} \emph{piano keyboard} & \emph{vintage typewriter photo by rusty blazenhoff} \\
        \midrule
        \rowcolor{Apricot!95} \emph{musical ben} & \emph{the typewriter is the best dead thing i ever found} \\
        \midrule
        \rowcolor{Apricot!70} \emph{she was really good, great voice, excellent guitar playing, and really nice to chat to.} & \emph{musical harmony} \\
        \midrule
        \rowcolor{Apricot!45} \emph{the highly-touted prospect, not the guitar player} & \emph{where some parts of me came of age} \\
        \midrule
        \rowcolor{Apricot!20} \emph{l.a.r.g.e} & $\downarrow$ \\
        \midrule
        \texttt{\textbf{[ROOT]}} & \texttt{\textbf{[ROOT]}} \\
    \end{tabularx}

    \vspace{5pt}
    \adjustbox{valign=t}{
        \begin{minipage}{0.18\linewidth}
            \centering
            \frame{\includegraphics[width=\linewidth]{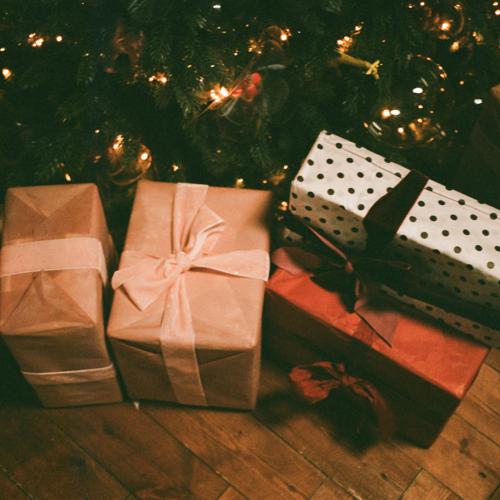}}
            (55)
        \end{minipage}
    }
    \hfill
    \begin{tabularx}{0.8\linewidth}[t]{ZZ}
        \rowcolor{Gray!40} \textbf{MERU} & \textbf{CLIP} \\
        \midrule
        \rowcolor{Apricot!120} \emph{my christmas shopping is done, and what's more, my presents are all wrapped!} & \emph{my homemade christmas book - dec. 5th and 6th.} \\
        \midrule
        \rowcolor{Apricot!100} \emph{this year's wrapping job.} & \emph{gift-wrapped for any occasion.} \\
        \midrule
        \rowcolor{Apricot!80} \emph{and gift wrapped!} & \emph{a wrapped xmas present} \\
        \midrule
        \rowcolor{Apricot!60} \emph{21 presents for my 21st for msh may} & \emph{our christmas present put to good use.} \\
        \midrule
        \rowcolor{Apricot!40} \emph{christmas is coming. won't someone think of my needs?} & $\downarrow$ \\
        \midrule
        \rowcolor{Apricot!20} \emph{the only good thing the guys did was dropped off the gifts.} & $\downarrow$ \\
        \midrule
        \texttt{\textbf{[ROOT]}} & \texttt{\textbf{[ROOT]}} \\
    \end{tabularx}

    \vspace{5pt}
    \adjustbox{valign=t}{
        \begin{minipage}{0.18\linewidth}
            \centering
            \frame{\includegraphics[width=\linewidth]{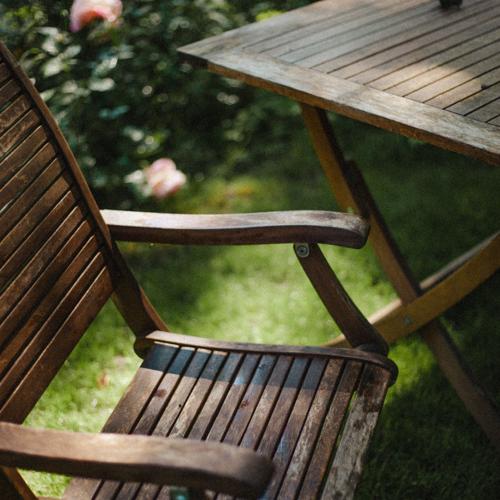}}
            (56)
        \end{minipage}
    }
    \hfill
    \begin{tabularx}{0.8\linewidth}[t]{ZZ}
        \rowcolor{Gray!40} \textbf{MERU} & \textbf{CLIP} \\
        \midrule
        \rowcolor{Apricot!120} \emph{table in the backyard of the summer house in melby, denmark.} & \emph{chair detail} \\
        \midrule
        \rowcolor{Apricot!105} \emph{kitch at airbnb at corte del correggio - note window behind chairs} & $\downarrow$ \\
        \midrule
        \rowcolor{Apricot!91} \emph{photo by laura nawrocik some patio furniture that needs a little cleaning.} & $\downarrow$ \\
        \midrule
        \rowcolor{Apricot!77} \emph{from a bench on the north side.} & $\downarrow$ \\
        \midrule
        \rowcolor{Apricot!62} \emph{the backyard of the b and b we stayed at} & $\downarrow$ \\
        \midrule
        \rowcolor{Apricot!48} \emph{another from this shoot in a more traditional style.} & $\downarrow$ \\
        \midrule
        \rowcolor{Apricot!34} \emph{traditional place to take a picture.} & $\downarrow$ \\
        \midrule
        \rowcolor{Apricot!20} \emph{a nice place to take pictures!} & $\downarrow$ \\
        \midrule
        \texttt{\textbf{[ROOT]}} & \texttt{\textbf{[ROOT]}} \\
    \end{tabularx}
\end{figure*}

\clearpage

\begin{figure*}[h]
    \footnotesize
    \newcolumntype{Z}{>{\centering\arraybackslash}X}
    \setlength{\tabcolsep}{2pt}

    \vspace{5pt}
    \adjustbox{valign=t}{
        \begin{minipage}{0.18\linewidth}
            \centering
            \frame{\includegraphics[width=\linewidth]{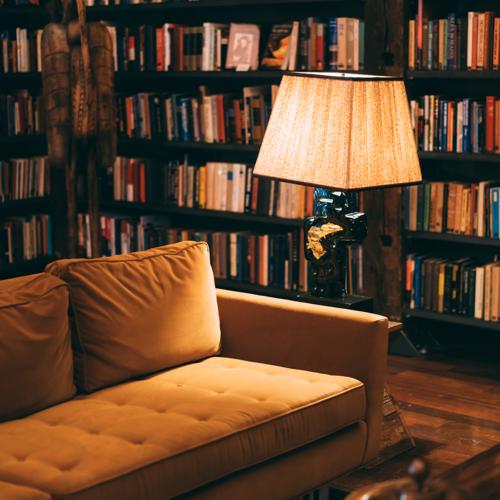}}
            (57)
        \end{minipage}
    }
    \hfill
    \begin{tabularx}{0.8\linewidth}[t]{ZZ}
        \rowcolor{Gray!40} \textbf{MERU} & \textbf{CLIP} \\
        \midrule
        \rowcolor{Apricot!120} \emph{found at city lights book store, sf www.citylights.com} & \emph{ikea catalog waiting for pickup, fairborn, ohio.} \\
        \midrule
        \rowcolor{Apricot!105} \emph{in powells rare book room} & \emph{special collections - amsterdam, netherlands} \\
        \midrule
        \rowcolor{Apricot!91} \emph{in the rare book reading room} & \emph{one needs to get there to read the it full.} \\
        \midrule
        \rowcolor{Apricot!77} \emph{book heaven.} & $\downarrow$ \\
        \midrule
        \rowcolor{Apricot!62} \emph{not a book in sight} & $\downarrow$ \\
        \midrule
        \rowcolor{Apricot!48} \emph{even more books} & $\downarrow$ \\
        \midrule
        \rowcolor{Apricot!34} \emph{sorry... i really like this space.} & $\downarrow$ \\
        \midrule
        \rowcolor{Apricot!20} \emph{with something like this, you would have to get a few} & $\downarrow$ \\
        \midrule
        \texttt{\textbf{[ROOT]}} & \texttt{\textbf{[ROOT]}} \\
    \end{tabularx}

    \vspace{5pt}
    \adjustbox{valign=t}{
        \begin{minipage}{0.18\linewidth}
            \centering
            \frame{\includegraphics[width=\linewidth]{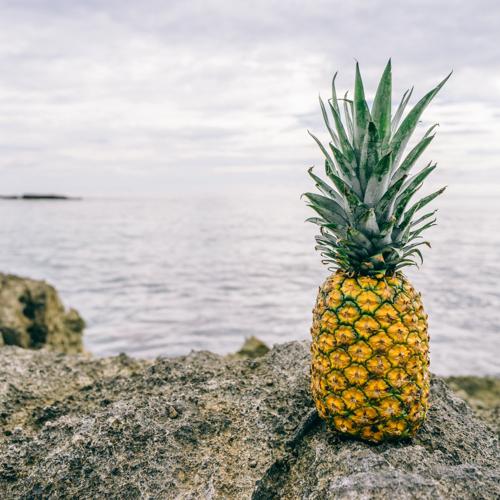}}
            (58)
        \end{minipage}
    }
    \hfill
    \begin{tabularx}{0.8\linewidth}[t]{ZZ}
        \rowcolor{Gray!40} \textbf{MERU} & \textbf{CLIP} \\
        \midrule
        \rowcolor{Apricot!120} \emph{a pineapple grows in the wild between goa gajah and yeh pulu, bali.} & \emph{so much organic burden on its way to the sea.} \\
        \midrule
        \rowcolor{Apricot!105} \emph{the pineapple, dunmore park. n} & \emph{everything is so organic on lamu island.} \\
        \midrule
        \rowcolor{Apricot!91} \emph{it started with more fruits but i didn't take this picture till late} & \emph{it's been that long since i took this that the next lot is already growing.} \\
        \midrule
        \rowcolor{Apricot!77} \emph{shot $<$usr$>$ beach} & \emph{the one that didn't get picked yet} \\
        \midrule
        \rowcolor{Apricot!62} \emph{a regular sight from our coast} & $\downarrow$ \\
        \midrule
        \rowcolor{Apricot!48} \emph{la nature au carre.} & $\downarrow$ \\
        \midrule
        \rowcolor{Apricot!34} \emph{i would be happy if all my photos turned out like this one.} & $\downarrow$ \\
        \midrule
        \rowcolor{Apricot!20} \emph{this is one is good.} & $\downarrow$ \\
        \midrule
        \texttt{\textbf{[ROOT]}} & \texttt{\textbf{[ROOT]}} \\
    \end{tabularx}

    \vspace{5pt}
    \adjustbox{valign=t}{
        \begin{minipage}{0.18\linewidth}
            \centering
            \frame{\includegraphics[width=\linewidth]{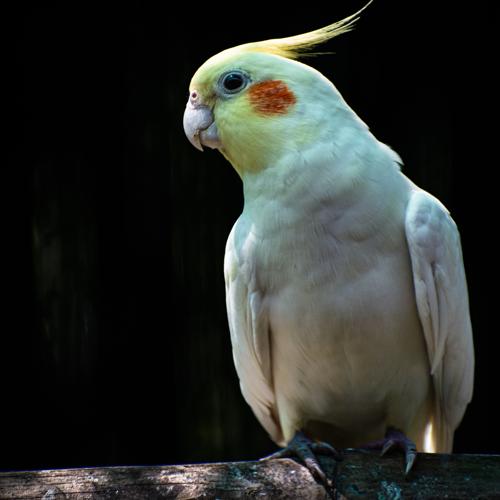}}
            (59)
        \end{minipage}
    }
    \hfill
    \begin{tabularx}{0.8\linewidth}[t]{ZZ}
        \rowcolor{Gray!40} \textbf{MERU} & \textbf{CLIP} \\
        \midrule
        \rowcolor{Apricot!120} \emph{sulphur crested cockatoos are great characters} & \emph{soleil when she was a baby with her green feathers} \\
        \midrule
        \rowcolor{Apricot!103} \emph{au pied du ciel} & \emph{lenny white )} \\
        \midrule
        \rowcolor{Apricot!86} \emph{pale male's mate \#5} & \emph{white tee's - photos for everyone} \\
        \midrule
        \rowcolor{Apricot!70} \emph{i think this is a nice photo of sean, though i doubt he will think so.} & $\downarrow$ \\
        \midrule
        \rowcolor{Apricot!53} \emph{photo: george struikelblok} & $\downarrow$ \\
        \midrule
        \rowcolor{Apricot!36} \emph{this is birdy.} & $\downarrow$ \\
        \midrule
        \rowcolor{Apricot!20} \emph{q's and a's} & $\downarrow$ \\
        \midrule
        \texttt{\textbf{[ROOT]}} & \texttt{\textbf{[ROOT]}} \\
    \end{tabularx}

    \vspace{5pt}
    \adjustbox{valign=t}{
        \begin{minipage}{0.18\linewidth}
            \centering
            \frame{\includegraphics[width=\linewidth]{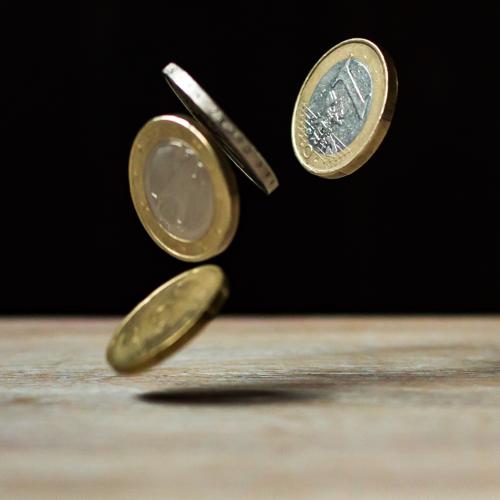}}
            (60)
        \end{minipage}
    }
    \hfill
    \begin{tabularx}{0.8\linewidth}[t]{ZZ}
        \rowcolor{Gray!40} \textbf{MERU} & \textbf{CLIP} \\
        \midrule
        \rowcolor{Apricot!120} \emph{uzi usb drive by dan helmick. brass, wood, riveted.} & \emph{i swear, this is how the coins landed. so i had to take a photo.} \\
        \midrule
        \rowcolor{Apricot!100} \emph{the coin toss} & \emph{money was a fun picture to take.} \\
        \midrule
        \rowcolor{Apricot!80} \emph{inset a coin...it moves...} & \emph{i really should have pictures of all the money, but after awhile one loses interest.} \\
        \midrule
        \rowcolor{Apricot!60} \emph{ver. 2} & $\downarrow$ \\
        \midrule
        \rowcolor{Apricot!40} \emph{made in a post secret kinda a way to tell something.} & $\downarrow$ \\
        \midrule
        \rowcolor{Apricot!20} \emph{no-one will ever guess} & $\downarrow$ \\
        \midrule
        \texttt{\textbf{[ROOT]}} & \texttt{\textbf{[ROOT]}} \\
    \end{tabularx}
\end{figure*}